\documentclass[10pt]{article}
\usepackage[preprint]{tmlr}

\usepackage{hyperref}
\usepackage{url}
\usepackage{xcolor}
\usepackage{enumitem}
\usepackage{amsmath}
\usepackage{amssymb}
\usepackage{amsthm}
\usepackage{graphicx}
\usepackage{rotating}
\usepackage{subcaption}
\usepackage{booktabs}

\providecommand{\manifold}{\mathcal{M}}
\providecommand{\R}{\mathbb{R}}

\newtheorem{theorem}{Theorem}
\newtheorem{corollary}{Corollary}

\title{Manifold-Matching Autoencoders}

\author{\name Laurent Cheret \email lcher021@uottawa.ca \\
      \addr Department of Computer Science, University of Ottawa, Ottawa, Canada
      \AND
      \name Vincent L\'etourneau \\
      \addr Mila, Montr\'eal, Canada
      \AND
      \name Isar Nejadgholi \\
      \addr National Research Council of Canada, Ottawa, Canada
      \AND
      \name Chris Drummond \\
      \addr National Research Council of Canada, Ottawa, Canada
      \AND
      \name Hussein Al Osman \\
      \addr Department of Computer Science, University of Ottawa, Ottawa, Canada
      \AND
      \name Maia Fraser \\
      \addr Department of Computer Science, University of Ottawa, Ottawa, Canada}

\def\month{MM}
\def\year{YYYY}
\def\openreview{\url{https://openreview.net/forum?id=XXXX}}

\begin{document}
\maketitle

\begin{abstract}
Manipulating the structure of data in the latent space of an autoencoder is a
long-standing goal of representation learning. We propose a simple
regularization, \emph{Manifold Matching}, that imposes an explicit geometric
structure on that latent space. The idea is to take a \emph{reference}
representation $R$ of the data, which need not be the raw input representation $X$ fed to
the network, and to minimize, batch-wise, the discrepancy between the
pairwise distances in $R$ and the pairwise distances in the latent space $Z$,
measured by the mean squared error between the two distance matrices. Although
the autoencoder still learns to reconstruct the input $X$, the shape of its latent space
comes to \emph{mimic} that of $R$. The matching objective requires no labels, only reference distances, and because those are single scalars, the dimensionality of $R$ is decoupled from that
of $Z$: any reference can guide any bottleneck, provided the reference
distances are known. Using this single instrument we study how different
geometric structures can be imposed on a latent space, comparing against
autoencoders built on geometric or topological regularizers. Across fifteen datasets under a shared
protocol with statistical testing, the resulting Manifold-Matching Autoencoder (MMAE) matches the global-structure quality of persistence-based
autoencoders, statistically tying RTD-AE on distance correlation and triplet accuracy, and stays
competitive on the local and topological metrics that other methods lead, at the computational cost
of a plain autoencoder: it uses only a batched distance loss, with no persistence computation. We frame MMAE
not as a universal winner but as an efficient and controllable regulariser whose reference selects which structure
the latent inherits, and which turns any fixed embedding into a reusable out-of-sample encoder. All
experiments use a two-dimensional bottleneck.
\end{abstract}

\section{Introduction}
\label{sec:intro}
Autoencoders are parametric models for dimensionality reduction: by minimising
reconstruction error, they learn an encoder that compresses data and can map new data points into the learned latent representation
\citep{hinton2006reducing}. That parametric generalisation is what many non-parametric dimensionality-reduction techniques lack. However, a reconstruction-only objective does not explicitly constrain neighbourhoods in the bottleneck, so it does not guarantee preservation of the input or any other chosen geometry. Two similar inputs
that ought to map to nearby regions of the latent space can be sent to opposite
sides of it. A reader visually inspecting that latent space would then wrongly conclude
that the two inputs are dissimilar, or that one is an anomaly, and would read the distance between them as
meaningful when it is an accident of optimization \citep{batson2021topological}. This is not a superficial
concern: the dimensionality-reduction literature commonly assesses the quality of an
embedding by how faithfully it preserves neighborhoods and
distances \citep{venna2001neighborhood, lee2007nonlinear}, used as
the standard axes when embeddings are compared \citep{espadoto2021survey}. But
these are exactly the quantities a plain reconstruction objective leaves free,
and distances in popular embeddings are routinely over-interpreted even when
they are not faithful \citep{wattenberg2016tsne}.

Examples of tasks that can benefit from controlling latent geometry include interpolation \citep{palma2025enforcing}, where a straight path in the latent should decode
to a plausible morph; sampling, where high-density regions of the latent should yield better generations \citep{chadebec2022geometric, xu2024Assessing}; anomaly detection, where ``far from the data'' must be
a trustworthy notion of distance; and clustering, where the relative distances between clusters and their size matter \citep{moor2020topological}. Local and global fidelity
are distinct, can trade off, and must be deliberately protected 
\citep{kobak2021initialization, amid2019trimap, moon2019phate}. Distance-distorting embeddings of single-cell data have been argued to be as
arbitrary as forcing the points into a chosen shape \citep{chari2023genomics}. In MMAE, by contrast, the target
shape is an explicit, inspectable modelling choice (Figure~\ref{fig:manual}) rather than a side effect of the algorithm.
The tension is long-standing: preserving global distances
\citep{tenenbaum2000global} versus local neighbourhoods
\citep{roweis2000nonlinear} is a central trade-off of manifold learning. The
reconstruction loss present in autoencoders, by contrast, is an average over individual points; it does not penalize geometric distortion and hence does not provide a guarantee that a latent geometry would be preserved.

\begin{figure}[t]
\centering
\includegraphics[width=\linewidth]{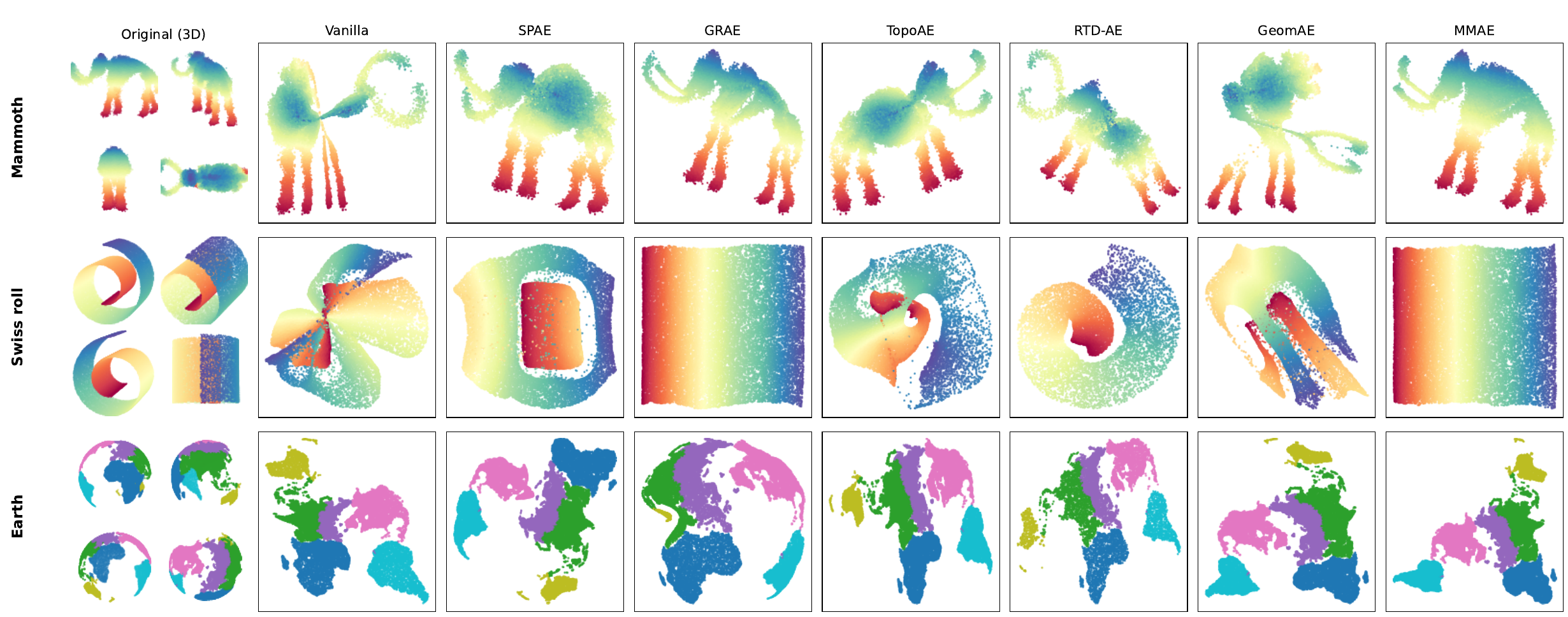}
\caption{\textbf{Distance preservation across three manifolds.} Rows: mammoth, swiss
roll, and a real Earth globe, each embedded from 3D to a 2D bottleneck. First column:
original 3D data from four viewpoints; other columns: seven autoencoders (${\sim}24$k
points), coloured by a manifold parameter (mammoth height / swiss-roll angle /
continent). MMAE (ours) preserves the geometry chosen by its reference, keeping the mammoth's
proportions, unrolling the swiss roll, and laying the globe out as a map, whereas
reference-free and topology-only variants tear or fold the manifold. Latents are shown up
to rotation and reflection, to which distance matching is invariant.}
\label{fig:manifold}
\end{figure}

Regularisation is the natural remedy: by adding a term to the objective one can
control the geometric structure that the bottleneck produces. But there is no single geometry that should always be preserved: ambient distances, intrinsic manifold distances, and neighborhood structure encode different relations and serve different purposes. The problem is therefore not only to preserve structure, but to specify which structure the latent should inherit. In an autoencoder this structure
must be imposed rather than assumed, and a growing family of geometry- and
topology-aware autoencoders attempts to do so through distinct mechanisms, 
persistent homology \citep{carlsson2009topology,moor2020topological, trofimov2023learning}, or isometric and geometry-preserving regularisation
\citep{lim2024ggae,nazari2023geomae}, alongside related graph- and
distance-based variants \citep{singh2021spae,GRAE,chen2022localdistae}. These approaches make different trade-offs in  computational cost, hyperparameter sensitivity, and the type of structure they constrain. Section 2.3 discusses these differences method by method. In light of this, we propose a middle ground
that leads to a simple form of structure manipulation, a regularizer we
call \emph{Manifold Matching}. See Figure~\ref{fig:manifold} for an illustration of how these distinct regularizations perform on 3D datasets when they are embedded into a 2D latent space.

\paragraph{Contributions.}
\textbf{(1)} We introduce Manifold Matching, a reference-based
regularizer whose reference dimension is decoupled from the bottleneck; the reference makes the target
geometry an explicit modelling choice rather than an implicit consequence of the training objective
(\S\ref{sec:method}). \textbf{(2)} We show, through a sequence of controlled demonstrations, that a
single distance-matching loss reproduces embeddings from diverse dimensionality-reduction algorithms
(t-SNE, UMAP, Isomap, PHATE) and extends them out-of-sample with one objective, a reusable-encoder capability that parametric
t-SNE/UMAP \citep{sainburg2021paraUMAP,maaten2009paramtSNE} and GRAE \citep{GRAE} each provide for a single
embedding method (we do not benchmark against these method-specific encoders here). \textbf{(3)} We show that with $X$ or
a near-isometric reference (PCA with sufficient $k$ components), MMAE recovers global structure,  such
as the nested-spheres nesting,  competitively with persistence-based autoencoders and without any
persistence computation. To support reproducibility, all code, configurations, and the evaluation
harness will be released upon publication.
\section{Background and related work}
\label{sec:background}
The introduction argued that when a latent space is intended to act as a readable map, its geometry should be meaningful rather than arbitrary, and that a plain reconstruction loss leaves that structure free. Here we make
``structure'' precise, map it to the quantities we later measure, review how prior methods
impose it, and isolate the single lever, matching distances to a chosen reference, that
Manifold Matching reduces to.
\subsection{What structure means, and who preserves it}

\paragraph{Two lenses: geometry and topology.}
Under the \emph{manifold hypothesis} \citep{Goodfellow-et-al-2016}, high-dimensional data
$\mathbf{X}=\{\mathbf{x}_i\}_{i=1}^N$, $\mathbf{x}_i\in\R^n$, lies on or near a $d$-dimensional
manifold $\manifold\subset\R^n$ with $d\ll n$,  a space locally homeomorphic to $\R^d$ whose
coordinates are the data's few factors of variation. Such a manifold can be read through two
complementary lenses, distinguished by the maps under which each is invariant. \emph{Geometry} is its
metric structure,  the distances, angles and local scales inherited from the ambient space,  and
is preserved exactly by \emph{isometries} (distance-preserving maps).
\emph{Topology} is the coarser structure preserved by \emph{homeomorphisms} (continuous deformations
with continuous inverse),  connected components, loops and voids, formally the homology groups
$H_0, H_1, H_2$. A circle and an ellipse are geometrically different but topologically identical; a
circle and a filled disc are topologically different.

Because the topology of the underlying manifold is not directly accessible from a finite sample, it
is measured through \emph{persistent homology} \citep{carlsson2009topology}: one grows a scale
parameter over a Vietoris--Rips filtration and records the scale at which each homological feature is
born and dies, summarising the result in a \emph{persistence diagram}. Controlling which lens an embedding preserves,  and it need not preserve both
,  is the subject of this work.

\paragraph{Classical dimensionality reduction.}
Linear methods preserve global geometry: PCA \citep{Hotelling1933pca,jolliffe2002principal}
projects onto the subspace of maximal variance, and classical MDS \citep{torgerson1952mds}
finds a configuration whose pairwise distances best match an input distance matrix; on
Euclidean data the two coincide \citep{borg2005modern}. Because their projections are linear
recombinations of the input coordinates, they behave \emph{more isometrically} as
components are added, each new dimension can only reduce the residual distance error.
An illustration of these results can be seen in \citet{trofimov2023learning}, where PCA outperforms the author's proposed method (RTD-AE) as well as all other competitors when sufficient components are added (i.e. MNIST 16D, FMNIST 64D, CIFAR10 128D,...), in terms of global structure preservation (distance correlation (DC), triplet accuracy (TA)). Nonlinear methods trade this off: Isomap \citep{tenenbaum2000global} substitutes
\emph{geodesic} distances estimated on a $k$-NN graph and can unroll curved manifolds such
as the swiss roll, while UMAP \citep{mcinnes2020umap} and t-SNE \citep{vandermaattsne}
preserve \emph{local} neighbourhoods for 2--3D visualisation but gain little from extra
output dimensions (the widely used Barnes--Hut implementation of t-SNE being limited to at most 3 components). Crucially, none of these extends to unseen points out of the box (the
parametric variants \citep{sainburg2021paraUMAP,maaten2009paramtSNE} being the exception),
whereas an autoencoder learns a reusable encoder. That out-of-sample generalization is the
reason we regularise an autoencoder rather than adopt a classical method, while still
recovering, through the reference dimension, the ``more components, less distortion'' behaviour of linear methods (\S\ref{sec:eval}).

\subsection{Distance preservation as a lever}
\label{sec:theory}
Distance preservation is the oldest lever in dimensionality reduction. For example, classical
multidimensional scaling framed embedding as finding a configuration whose pairwise
distances match a target matrix \citep{torgerson1952mds,DeLeeuw1977smacof,borg2005modern}.
Training a neural network to minimise a stress-type objective has a long history, from neural Sammon mapping \cite{maojain1995tnn} and RBF-based MDS \cite{webb1995pattern} onward; our matching-only variant (Table \ref{tab:ablation}) belongs to this family. What we add is its use as a regulariser alongside reconstruction, the decoupling of reference and bottleneck dimension, and a systematic comparison against geometric and topological autoencoders.
Its less obvious property, and what makes the
lever powerful, is that matching distances also controls \emph{topology}, through the
\emph{stability theorem} for Vietoris--Rips \citep{carlsson2009topology,edelsbrunner2010computational}
persistence diagrams of finite metric spaces.
\begin{theorem}[Stability \citep{cohen2007stability,chazal2009gromov,chazal2014structure}]
\label{thm:stability}
\begin{equation}
d_B(\mathrm{Dgm}_p(X),\mathrm{Dgm}_p(Y)) \;\leq\; 2\,d_{GH}(X,Y)
\end{equation}
for all homology dimensions $p\geq 0$, where $d_B$ is the bottleneck distance and $d_{GH}$
the Gromov--Hausdorff distance.
\end{theorem}
Since the Gromov--Hausdorff distance is at most half the largest pairwise distance
discrepancy, $d_{GH}(X,Z)\le \tfrac12\max_{i,j}\lvert d_X(x_i,x_j)-d_Z(z_i,z_j)\rvert$,
uniform distance preservation bounds $d_{GH}$, and we obtain:
\begin{corollary}[Distance preservation bounds topological change]
\label{cor:main}
If $|d_X(x_i,x_j)-d_Z(f_\theta(x_i),f_\theta(x_j))|\leq\epsilon$ for all pairs, then
$d_B(\mathrm{Dgm}_p(X),\mathrm{Dgm}_p(f_\theta(X)))\leq \epsilon$ for all $p\geq 0$.
\end{corollary}
Three caveats govern how we use this. First, MMAE minimises the \emph{average} (mean-squared)
distortion, whereas Corollary~\ref{cor:main} assumes a uniform ($L_\infty$) bound on \emph{every} pair:
a small MSE can still leave a few badly distorted pairs, so our $L_2$ objective does not enforce the
corollary's premise and hence does not, in the worst case, bound the bottleneck distance. We therefore
treat Theorem~\ref{thm:stability} and Corollary~\ref{cor:main} as \emph{motivation} for why distance
preservation should control topology, not as a guarantee that MMAE does so; the empirical
$W_0,W_1$ results of \S\ref{sec:eval} are the actual evidence. Second, the bound is relative to the
\emph{reference} geometry: aligning the latent to a reference $R$ controls the topology with respect to $R$, which equals
the ambient topology only when $R$ is the ambient distance matrix; with a geodesic (Isomap)
reference the target can be instead the unrolled, intrinsic structure. Third, the implication
is one-way, a manifold can be unrolled, stretched, bent, or shrunk (topology kept) with its ambient geometry destroyed but its neighborhoods maintained, so distance preservation is sufficient but not necessary for topology
preservation. The reference is thus the design choice that selects \emph{which} structure is
kept, and the connection between the two lenses is exactly what we probe experimentally.
Because training is batch-wise, we rely on the observation that batch topology converges to
global topology as the batch grows \citep{moor2020topological}: the probability that a mini-batch
$X^{(b)}$ deviates topologically from $X$ is bounded by a Hausdorff sampling term,
$P(d_B(\mathrm{Dgm}(X),\mathrm{Dgm}(X^{(b)}))>\epsilon)\leq P(d_H(X,X^{(b)})>\tfrac{\epsilon}{2})$,
which vanishes as $b\to N$. \citet{moor2020topological, moor2020challenging} use this trick to extract the most persistent features from the distance matrices of $X^{(b)}$ and $Z^{(b)}$, and by minimizing the discrepancies, to align the topological structures of $X$ and $Z$. \citet{trofimov2023learning} later propose a more rigorous topological alignment, which, nonetheless, relies on the same batch-wise approximation. 
MMAE does not rely on this approximation: its batch loss is an unbiased estimate of the full-dataset loss (\S\ref{sec:method}), and it requires no persistence computation.

\subsection{Structure-aware autoencoders}
An autoencoder trained for reconstruction alone leaves the geometry of its latent space
unconstrained: the bottleneck fixes the dimension, but nothing pins down how points are arranged
within it. A growing line of work adds structure-preserving regularisers to close this gap,
reshaping the latent space to match, or surpass, non-parametric dimensionality reduction methods
while keeping the autoencoder's parametric, out-of-sample encoder.

\emph{Topology-derived} methods act on persistence: TopoAE \citep{moor2020topological} matches the
distances of topologically significant point pairs, in practice $H_0$, the zeroth homology that
tracks connected components, whose death scales coincide with the edges of the batch's minimum
spanning tree. This limits the loss to MST-edge distances, cheaply computed in $O(b^2)$ for a
batch of size $b$, but the signal is discontinuous under perturbation, a small change can swap
which edge enters the tree, and blind to higher-order features such as loops ($H_1$) or voids
($H_2$). RTD-AE \citep{trofimov2023learning} instead compares the two filtrations directly, through a
relative \emph{cross-barcode} whose bars measure the scale-by-scale disagreement between the
persistence of $X$ and of $Z$,  including loops and voids, not just the components TopoAE sees.
Summing these bar lengths is a stable, Lipschitz functional, so the loss is \emph{continuous} under
perturbation (the property TopoAE lacks), but at a high per-batch cost: persistence in degree $p$
requires the $(p{+}1)$-skeleton of the Rips complex, whose $\binom{b}{p+2}=\mathcal{O}(b^{\,p+2})$
simplices make the boundary reduction super-quadratic in $b$ and often force two-stage training. Both approaches try to preserve the
original data structure, meaning their regularizers ultimately operate on the pairwise Euclidean
distances in $Z$ and in $X$. A famous example of their improvement with respect to other
autoencoders and also other non-parametric dimensionality reduction mechanisms is the synthetic
dataset \textit{``nested spheres''}. Although this dataset is straightforward to construct, it has
consistently proven a challenging benchmark for many dimensionality reduction methods, which
struggle to preserve its topology \citep{moor2020topological,moor2020challenging}. To date, TopoAE and
RTD-AE are the only methods reported to preserve the nesting relationship between spheres, even among
non-parametric methods, a claim we revisit in \S\ref{sec:eval}, where a distance-faithful reference
reaches it without any persistence computation.

\emph{Distance- and graph-based} methods act on
geometry: SPAE \citep{singh2021spae} minimizes the variance of log distance-ratios
$\mathrm{Var}[\log(d_Z/d_X)]$, where for a pair of points $d_X=\lVert x_i-x_j\rVert$ and
$d_Z=\lVert z_i-z_j\rVert$ are their input and latent-space distances and the variance runs over all
pairs. The ratio $d_Z/d_X$ is the local factor by which the encoder stretches that pair, so driving
the variance to zero forces a single common stretch across every pair,  a globally consistent
rescaling of the geometry rather than an arbitrary one. The idea is also to preserve the original
data structure: the criterion is scale-invariant (a global change of scale leaves the variance
untouched) but unstable for small distances (as $d_X\to0$ the log-ratio blows up) and reliant on raw
high-dimensional distances. 
Unlike the topological variants, which use persistence to keep only the most salient features, SPAE weighs all pairs.
As dimensionality grows, pairwise distances concentrate around their mean \citep{aggarwal2001surprising}, so $\operatorname{log}(d_X)$ becomes nearly constant across pairs, and the regulariser mostly asks for equal latent distances, carrying little information about the input structure.
SPAE are closely related to MMAE with $R = X$: both are minimised by maps that preserve input distances (or up to a global scaling for SPAE), but they differ in form. SPAE penalises the spread of multiplicative distortions $\operatorname{log}(d_Z/d_X)$, whereas MMAE penalises additive distortions $(d_Z - d_X)^2$, which weights large distances more heavily. As we show in \S\ref{sec:eval}, this difference matters in practice.
Instead of using
straight-line Euclidean distances, GGAE \citep{lim2024ggae} builds a similarity graph that mimics the
data's true curved surface through the Laplace--Beltrami operator \citep{shi2016convergencelaplacian}
,  while actively preventing geometric distortion. Its weakness is the \textit{bandwidth} $t$ of the Gaussian heat kernel
$w_{ij}=\exp(-\lVert x_i-x_j\rVert^{2}/2t)$, whose entries form the affinity matrix $W$ and the graph
Laplacian $L=D-W$ that the regulariser acts on: too small isolates every point, too large connects
them all, so only a narrow, data-dependent $t$ recovers the Laplace--Beltrami operator. Building $W$
is also an $\mathcal{O}(N^2)$ bottleneck over much of the data, which limits GGAE to smaller datasets.
In a similar spirit, CAE, a contrastive autoencoder built on the supervised-contrastive objective of \citet{khosla2020supcon} with input-space $k$-nearest neighbours as positives, preserves local structure not through distances or
a graph but contrastively, treating each point's input-space $k$-nearest neighbours as positives and
pushing all others apart, so that neighbourhoods in $X$ are reproduced in $Z$.

\emph{Decoder-geometry} methods take the opposite route from the regularisers above: rather than
shaping where the encoder places points, they constrain how the decoder $g:Z\to X$ \emph{generates}
data from latent coordinates. The motivation is that $g$ turns the latent space into a
parameterisation of the reconstructed manifold $g(Z)$, whose local geometry is set by the decoder
Jacobian $J_g(z)=\partial g/\partial z$ and the pullback metric $G(z)=J_g(z)^\top J_g(z)$; making
$G(z)$ well behaved is what lets straight lines and uniform sampling in $Z$ correspond to sensible
motions on the manifold. 
GeomAE \citep{nazari2023geomae} penalises the variance of the log generalised Jacobian determinant
$\operatorname{Var}_z\left[\log\det\left(J_g(z)^\top J_g(z)\right)\right] = \operatorname{Var}_z\left[\log\det G(z)\right]$
 is the local factor by which the decoder scales volume, forcing it to be constant across $Z$ makes the map \emph{equiareal}, removing the arbitrary expansions and contractions that let a vanilla autoencoder manufacture fake dense or sparse regions. 
IsoAE \citep{gropp2020isometric} asks for more, a local
\emph{isometry} $\lVert J_g(z)v\rVert\approx\lVert v\rVert$, i.e. $G(z)\approx I$, enforced by
sampling random unit vectors $v$ and penalising $(\lVert J_g(z)v\rVert-1)^2$, so that latent
Euclidean distance approximates geodesic distance on the manifold.
Both constrain \emph{local, first-order} structure, the differential behavior of the decoder at
each $z$, which fixes stretching and volume distortion but says nothing about global arrangement.

\textit{Reference-based} methods are the closest in spirit to our work. More precisely GRAE \citep{GRAE}, which precomputes an
embedding $E$ and aligns latent \emph{coordinates} to it,
$\mathcal{L}_g=\sum_i\|E_i-f_\theta(x_i)\|^2$; this forces the reference and bottleneck
dimensions to coincide. The encoder becomes a parametrized version of the reference embedding, allowing new inputs to be mapped and compared to others on the same $E$. It limits the autoencoder to the structure uncovered by the chosen embedding mechanism (t-SNE, PCA, UMAP, PHATE, ...etc), being exclusively a parametrization and generalization mechanism of known embeddings. 
\begin{figure}[t]
\centering
\includegraphics[width=0.8\linewidth]{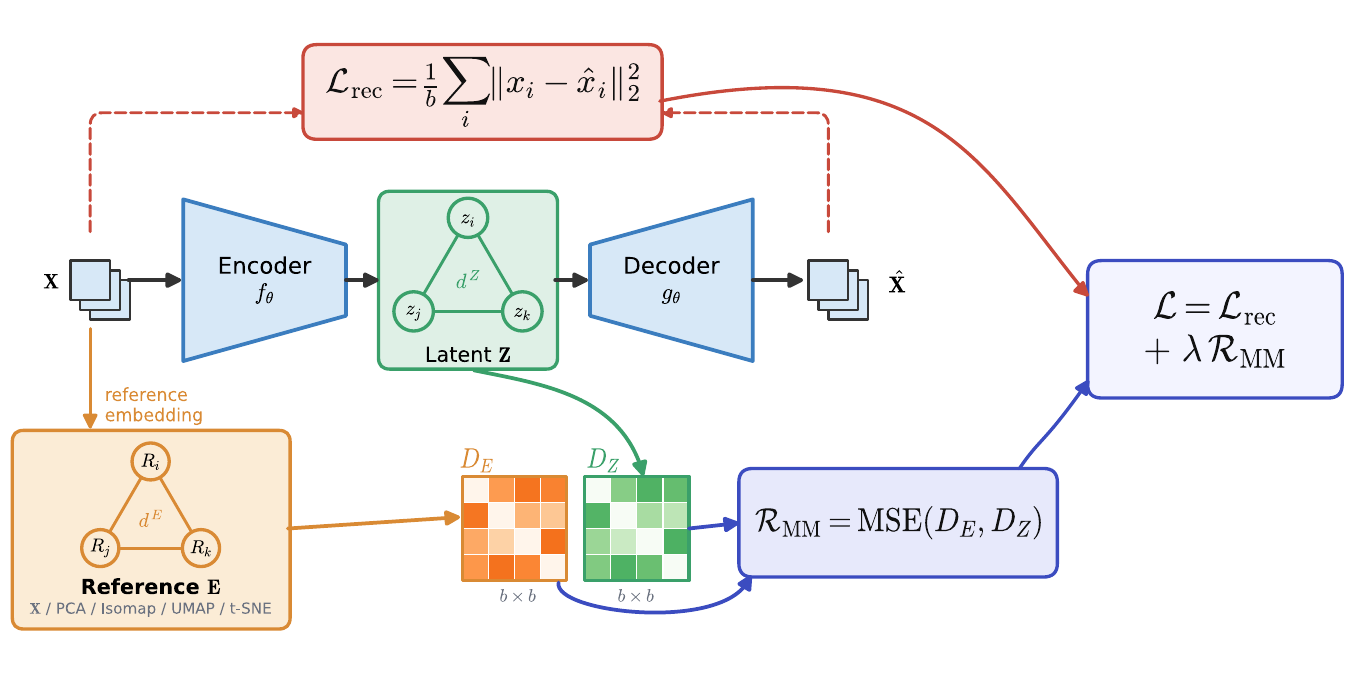}
\caption{\textbf{The Manifold-Matching framework.} On each minibatch, the encoder
$f_\theta$ maps inputs $\mathbf{X}$ to latent codes $\mathbf{Z}$ and the decoder
$g_\phi$ reconstructs $\hat{\mathbf{X}}$. Alongside the reconstruction loss, the
manifold-matching regulariser $\mathcal{R}_{\mathrm{MM}}$ compares the pairwise
distance matrix of the codes, $D_Z$, with that of a reference embedding $\mathbf{E}$,
$D_E$. The reference is a modelling choice (the ambient data $\mathbf{X}$, PCA,
Isomap, UMAP, t-SNE, \dots) and may live in any dimension, independent of the
bottleneck.}
\label{fig:method}
\end{figure}
\section{Manifold-Matching Autoencoders}
\label{sec:method}

An autoencoder \citep{hinton2006reducing} pairs an encoder $f_\theta:\R^n\!\to\!\R^d$ ($d\ll n$) with a decoder
$g_\phi:\R^d\!\to\!\R^n$ trained to reconstruct the input; the encoder is a learned
coordinate chart for the data manifold. Reconstruction constrains only
$g_\phi\!\circ\!f_\theta$ to be near-identity on the data,  it fixes the chart's
image but leaves its geometry free, so two codes can be placed arbitrarily without
changing the loss. Manifold Matching removes that freedom with a single term that
asks the latent to reproduce a reference distance structure (Fig.~\ref{fig:method}).

Concretely, let $\mathbf{E}=\{R_i\}$ be a reference embedding of the training set,
$R_i\in\R^m$. On a minibatch of $b$ points we form the two $b\times b$
pairwise-distance matrices $D_Z=[\,\|z_i-z_j\|\,]$ and $D_E=[\,\|R_i-R_j\|\,]$ and add
their mean-squared discrepancy to the reconstruction loss:
\begin{equation}
\mathcal{L}\;=\;
\underbrace{\tfrac{1}{b}\textstyle\sum_i \|x_i-\hat{x}_i\|_2^2}_{\mathcal{L}_{\mathrm{rec}}}
\;+\;\lambda\,
\underbrace{\tfrac{1}{b^2}\textstyle\sum_{i,j}\big(\|z_i-z_j\|-\|R_i-R_j\|\big)^2}_{\mathcal{R}_{\mathrm{MM}}=\mathrm{MSE}(D_Z,\,D_E)},
\label{eq:mm}
\end{equation}
where $\hat{x}_i=g_\phi(f_\theta(x_i))$ and $\lambda$ balances geometry against
reconstruction. The reference $\mathbf{E}$ is precomputed once and indexed per batch,
so each step costs only the $\mathcal{O}(b^2)$ batch distances. Because batches are drawn from a uniform distribution, the batch term is an unbiased estimate, up to the factor $(b - 1)/b$ from the zero diagonal, of the mean squared distortion over all pairs of the training set. Stochastic gradient descent on $\mathcal{R}_{MM}$ therefore minimises the full-dataset objective.

Two properties distinguish this construction. First, because
$\mathcal{R}_{\mathrm{MM}}$ matches a $b\times b$ matrix of \emph{distances} rather
than coordinates, the reference dimension $m$ is decoupled from the bottleneck
dimension $d$: a low-dimensional reference can shape a high-dimensional latent, and
vice versa. This is the key difference from coordinate-aligning methods such as GRAE
\citep{GRAE}, which require $m=d$. Second, distances may be matched raw,  pinning the
latent scale to the reference,  or after a per-batch $z$-score normalisation, $D\mapsto(D-\bar D)/\sigma_D$ applied to each batch distance
matrix $D_Z$ and $D_E$, that makes the term scale-invariant; we treat this as a hyperparameter, and variants
using it carry the suffix ``-z'' (e.g.\ MMAE-raw-z). 
In the limit $\mathcal{R}_{MM} \to 0$, the encoder maps the training sample isometrically onto its image with respect to the reference metric, and Corollary~\ref{cor:main} with $\varepsilon = 0$ gives identical persistence diagrams. Away from that limit, a small $\mathcal{R}_{MM}$ does not bound the bottleneck distance (\S\ref{sec:theory}), so we treat this as motivation and assess topology empirically.

\begin{figure}[!h]
\centering
\includegraphics[width=0.8\linewidth]{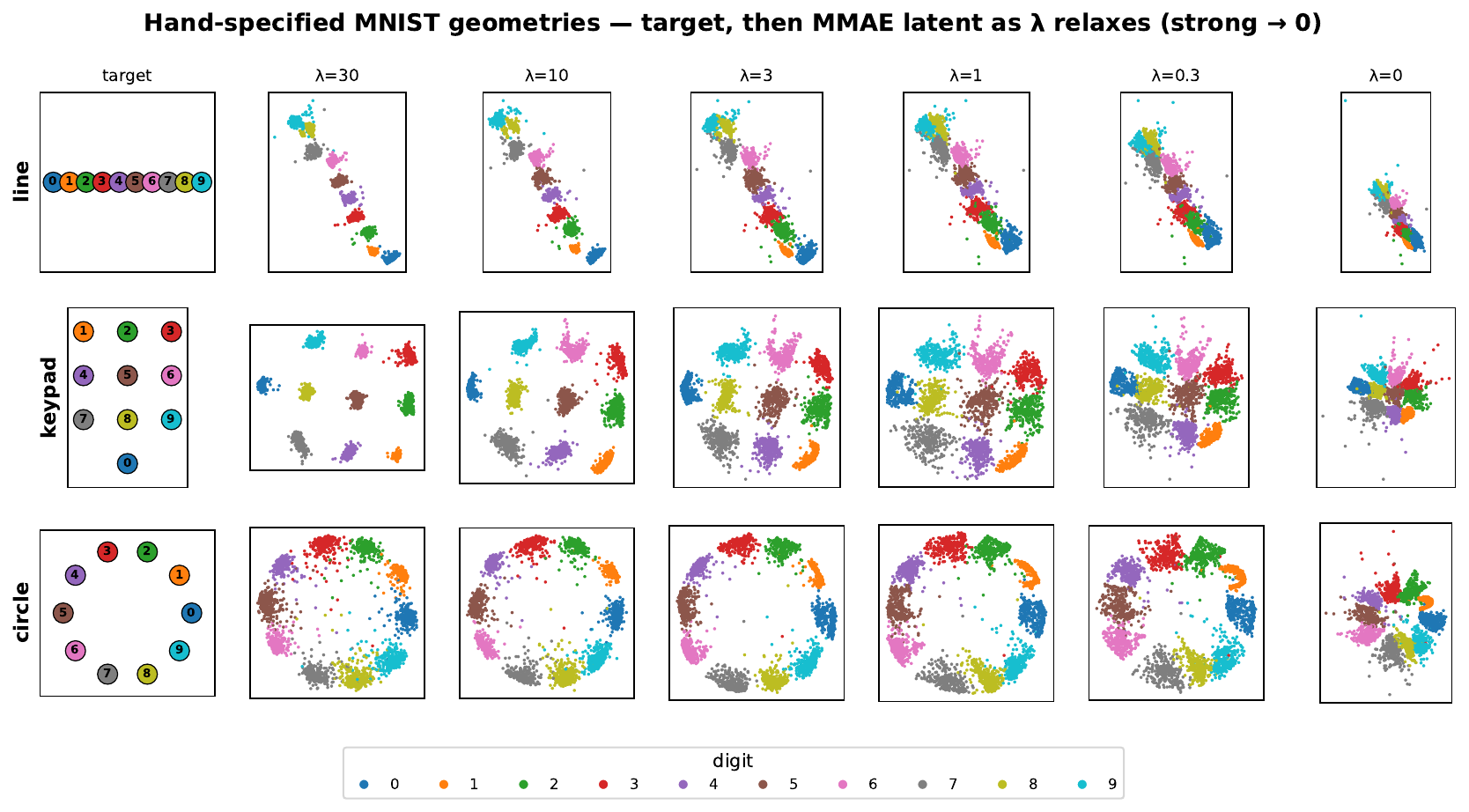}
\caption{\textbf{Geometries can be specified by hand.} Each row is a hand-drawn target
layout,  a line, a calculator keypad, a circle (leftmost schematic),  used as the
Manifold-Matching reference. The remaining columns show the realised MMAE latent as the
strength $\lambda$ is relaxed from strong (left) to $0$ (right), each model initialised from the previous one; the $\lambda=0$ latent retains elements of the earlier structure. Clusters
are coloured by digit $0$--$9$; orientation is arbitrary, since distance matching is
invariant to rotation and reflection. In this specific illustration the geometry is constructed using the known labels, however, our work focuses on unsupervised construction of references.}
\label{fig:manual}
\end{figure}

\paragraph{The reference selects the geometry.}
Equation~\eqref{eq:mm} does not say \emph{which} distances to preserve; the reference
is a modelling choice,  its type and its dimension,  and it decides which geometry
the chart inherits. We consider three families:
\begin{itemize}[leftmargin=1.4em,itemsep=1pt,topsep=2pt]
  \item \textbf{raw},  the ambient distance $\|x_i-x_j\|_2$, trustworthy for
  low-dimensional geometric data but unreliable in high dimensions, where distances
  concentrate \citep{aggarwal2001surprising};
  \item \textbf{PCA},  the distance after projecting onto the leading principal
  components; a denoised distance whose fidelity increases with the number of
  components;
  \item \textbf{geodesic},  distance \emph{along} the manifold, estimated by Isomap
  as shortest paths in a $k$-nearest-neighbour graph \citep{tenenbaum2000global},
  which respects curvature the ambient chord ignores;
\end{itemize}
On a curved manifold the chord
short-circuits through ambient space while the geodesic respects the surface, so the
same data and the same architecture yield \emph{different} latent geometries. The
rest of the paper is, in effect, a study of what each reference buys. Additionally, it is worth noting that the geometry imposed can be designed manually. Although this is not the main focus of this work, as we are interested in representations requiring no labels to be generated, we leave the illustration of this idea in Figure~\ref{fig:manual}. 
\section{Reproducing external representations}
\label{sec:reproduce}
The reference in Equation~\eqref{eq:mm} need not be a classical geometry: it can be \emph{any}
embedding, including one produced by a non-parametric method. This exposes a distinctive use of
Manifold Matching. Methods such as t-SNE \citep{vandermaattsne}, UMAP \citep{mcinnes2020umap},
Isomap \citep{tenenbaum2000global} and PHATE \citep{moon2019phate} 
embed a fixed dataset; out-of-sample extensions exist for some, such as the Nyström extension for Isomap \cite{bengio-oosExtensionIsomap2003neurips} and UMAP's transform, but each is method-specific.
If we take
such an embedding as the reference $E$, MMAE learns an \emph{encoder} $f_\theta$ whose 2D output
reproduces $E$,  turning a non-parametric layout into a parametric one that also embeds unseen
data. We include PCA \citep{jolliffe2002principal} as a linear control that already admits an
out-of-sample projection. We stress that this is not a new capability: parametric t-SNE
\citep{maaten2009paramtSNE}, parametric UMAP \citep{sainburg2021paraUMAP} and GRAE \citep{GRAE}
already provide reusable encoders,  each for its own embedding method. Our contribution here is
\emph{unification}: one distance-matching loss reproduces all of them and extends them out-of-sample,
and it is that out-of-sample fidelity (not the in-sample fit, which is function approximation) that is
the non-trivial part.

\begin{table}[!h]
\centering
\small
\caption{\textbf{MMAE reproduces external embeddings.} Training-set reproduction (distance
correlation DC, Procrustes disparity) and out-of-sample fidelity (DC of held-out points' distances to
all points), for five reference methods; means over five seeds with rotated splits. Reproduction is
near-perfect everywhere and out-of-sample DC stays $\ge 0.96$.}
\label{tab:reproduce}
\begin{tabular}{@{}ll cc c@{}}
\toprule
& & \multicolumn{2}{c}{Reproduction (train)} & OOS (test) \\
\cmidrule(lr){3-4}\cmidrule(lr){5-5}
Dataset & Ref & DC$\uparrow$ & Proc$\downarrow$ & DC$\uparrow$ \\
\midrule
MNIST & PCA & 0.999 & 0.001 & 0.999\,{\scriptsize$\pm$0.000} \\
 & t-SNE & 0.999 & 0.001 & 0.975\,{\scriptsize$\pm$0.001} \\
 & UMAP & 0.999 & 0.001 & 0.978\,{\scriptsize$\pm$0.001} \\
 & Isomap & 0.998 & 0.003 & 0.985\,{\scriptsize$\pm$0.002} \\
 & PHATE & 0.999 & 0.001 & 0.989\,{\scriptsize$\pm$0.001} \\
\midrule
Fashion-MNIST & PCA & 0.999 & 0.001 & 0.999\,{\scriptsize$\pm$0.000} \\
 & t-SNE & 0.999 & 0.001 & 0.989\,{\scriptsize$\pm$0.001} \\
 & UMAP & 0.999 & 0.001 & 0.995\,{\scriptsize$\pm$0.001} \\
 & Isomap & 0.998 & 0.002 & 0.993\,{\scriptsize$\pm$0.001} \\
 & PHATE & 0.999 & 0.002 & 0.997\,{\scriptsize$\pm$0.000} \\
\midrule
PBMC3k & PCA & 0.999 & 0.006 & 0.993\,{\scriptsize$\pm$0.004} \\
 & t-SNE & 0.999 & 0.001 & 0.961\,{\scriptsize$\pm$0.003} \\
 & UMAP & 1.000 & 0.001 & 0.974\,{\scriptsize$\pm$0.003} \\
 & Isomap & 0.999 & 0.004 & 0.975\,{\scriptsize$\pm$0.002} \\
 & PHATE & 1.000 & 0.000 & 0.983\,{\scriptsize$\pm$0.001} \\
\midrule
Paul15 & PCA & 0.997 & 0.028 & 0.995\,{\scriptsize$\pm$0.004} \\
 & t-SNE & 0.998 & 0.017 & 0.984\,{\scriptsize$\pm$0.002} \\
 & UMAP & 0.999 & 0.021 & 0.989\,{\scriptsize$\pm$0.003} \\
 & Isomap & 0.998 & 0.036 & 0.985\,{\scriptsize$\pm$0.002} \\
 & PHATE & 0.998 & 0.028 & 0.989\,{\scriptsize$\pm$0.001} \\
\bottomrule
\end{tabular}
\end{table}
\paragraph{Setup.}
For each dataset and reference method we compute a 2D embedding $E$ on the full dataset\footnote{$E$ is computed once on the full dataset, so the held-out points help shape the target layout; MMAE itself is trained on the training split only, and the out-of-sample score measures whether its encoder places unseen points where that full-data embedding puts them.}, standardise
it to unit scale, and train MMAE (2D bottleneck) on an $85\%/15\%$ split to match its pairwise distances.
Because MMAE matches \emph{raw} distances, the reference must first be placed on a comparable scale:
PHATE's coordinates are $\sim\!10^{3}$ times smaller than t-SNE's, and without this rescaling the
reconstruction term overwhelms the (tiny) matching signal. We report the average reproduction
fidelity on the training points,  distance correlation (DC) and Procrustes disparity, after optimal
rotation/scale/reflection,  and out-of-sample fidelity on the held-out $15\%$: the distance
correlation between a held-out point's distances to \emph{all} points under the reference and under
the encoder, i.e.\ whether a new point is placed at the right distances from everything. All numbers
are means over five seeds, each with its own random split.

\begin{figure}[!h]
\centering
\includegraphics[width=0.8\linewidth]{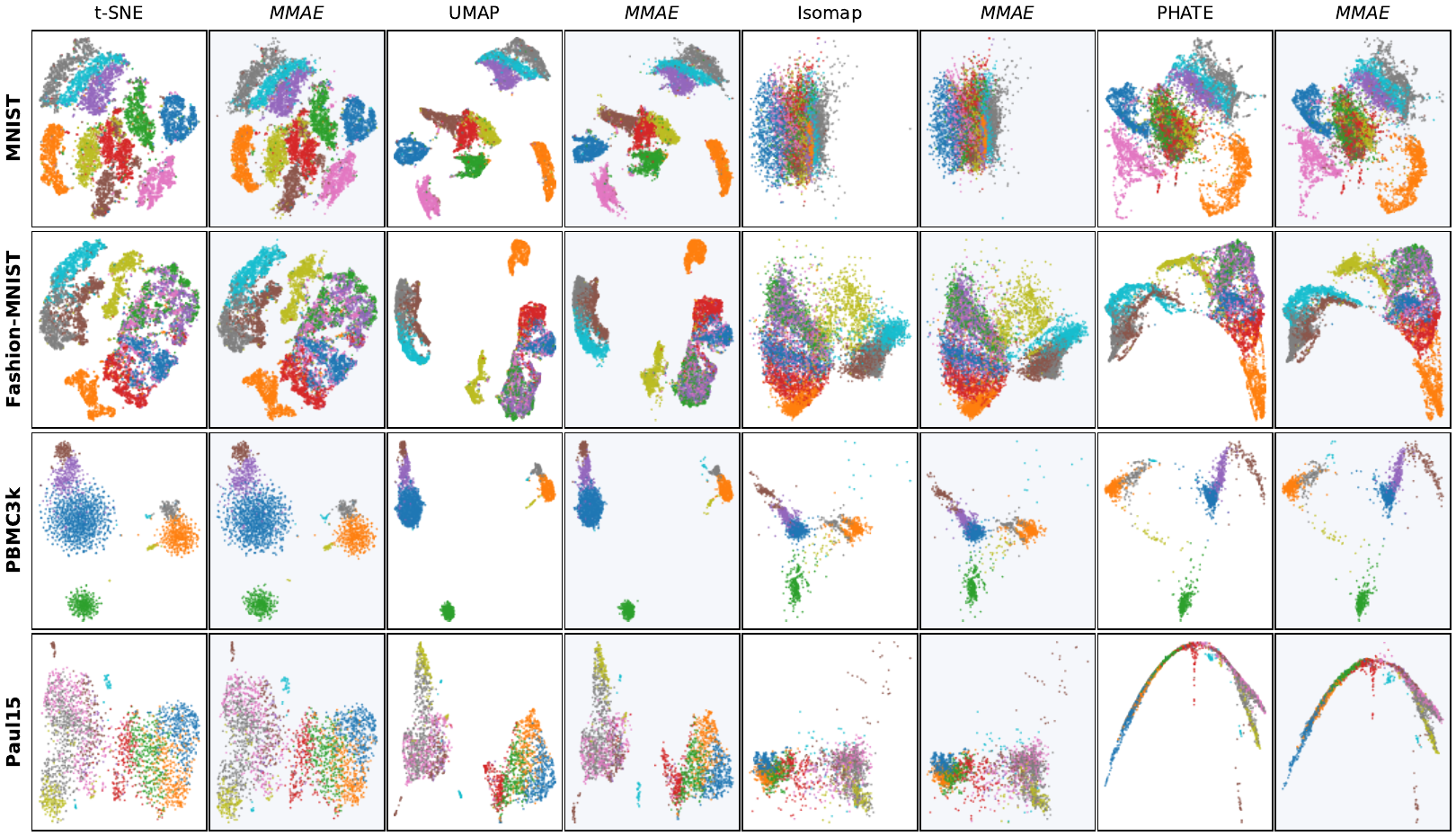}
\caption{\textbf{MMAE as a parametric embedding.} Rows: four datasets (two image, two single-cell).
Columns: the four non-parametric references, each shown as a pair,  the reference embedding beside
the MMAE latent trained to match it (tinted),  coloured by class and Procrustes-aligned per pair.
In every pair the MMAE panel reproduces its reference.}
\label{fig:reproduce}
\end{figure}

\paragraph{MMAE copies the reference almost exactly.}
Across all five references and all four datasets, the training reproduction is near-perfect: distance
correlation $\ge 0.997$ and Procrustes disparity $\le 0.04$ (Table~\ref{tab:reproduce}). The
embeddings are visually indistinguishable from their references (Fig.~\ref{fig:reproduce}): the MMAE
panel copies the clusters of t-SNE and UMAP, the branch structure of PHATE, and the differentiation
trajectory of Paul-2015. The encoder is, in effect, a parametric surrogate for the chosen embedding.

\paragraph{And it generalises to new points.}
The learned encoder does not merely fit the training set: it extends the reference to unseen points.
On held-out data, a new point's distances to the rest of the embedding match the reference with
DC $\ge 0.96$ for every method and dataset (Table~\ref{tab:reproduce}), with negligible variance over
splits and seeds. Fidelity tracks how smoothly the reference maps the input, as expected: PCA,  a
linear map with a native out-of-sample projection,  is reproduced almost perfectly out-of-sample
(DC $\approx 0.99$), the smooth manifold methods (PHATE, Isomap, UMAP) follow closely, and t-SNE is
the marginally hardest ($0.96$--$0.99$). MMAE thus turns each non-parametric embedding into a
reusable encoder that places new data consistently with the original.

\section{Preserving the original data geometry}
\label{sec:eval}

Section~\ref{sec:reproduce} showed that MMAE can \emph{reshape} the latent to match an external
target: it manipulates the geometry to order. Here we show the complementary ability. When the
reference is the data's \emph{own} high-dimensional distances, MMAE \emph{preserves} that geometry in
two dimensions,  recovering global organisation that PCA, t-SNE, UMAP and even other autoencoders
discard. We begin with a case where the structure to preserve is known exactly and is famously hard to
keep in two dimensions.

\subsection{A simple yet difficult case: nested spheres.}
\begin{figure}[!h]
\centering
\includegraphics[width=0.8\linewidth]{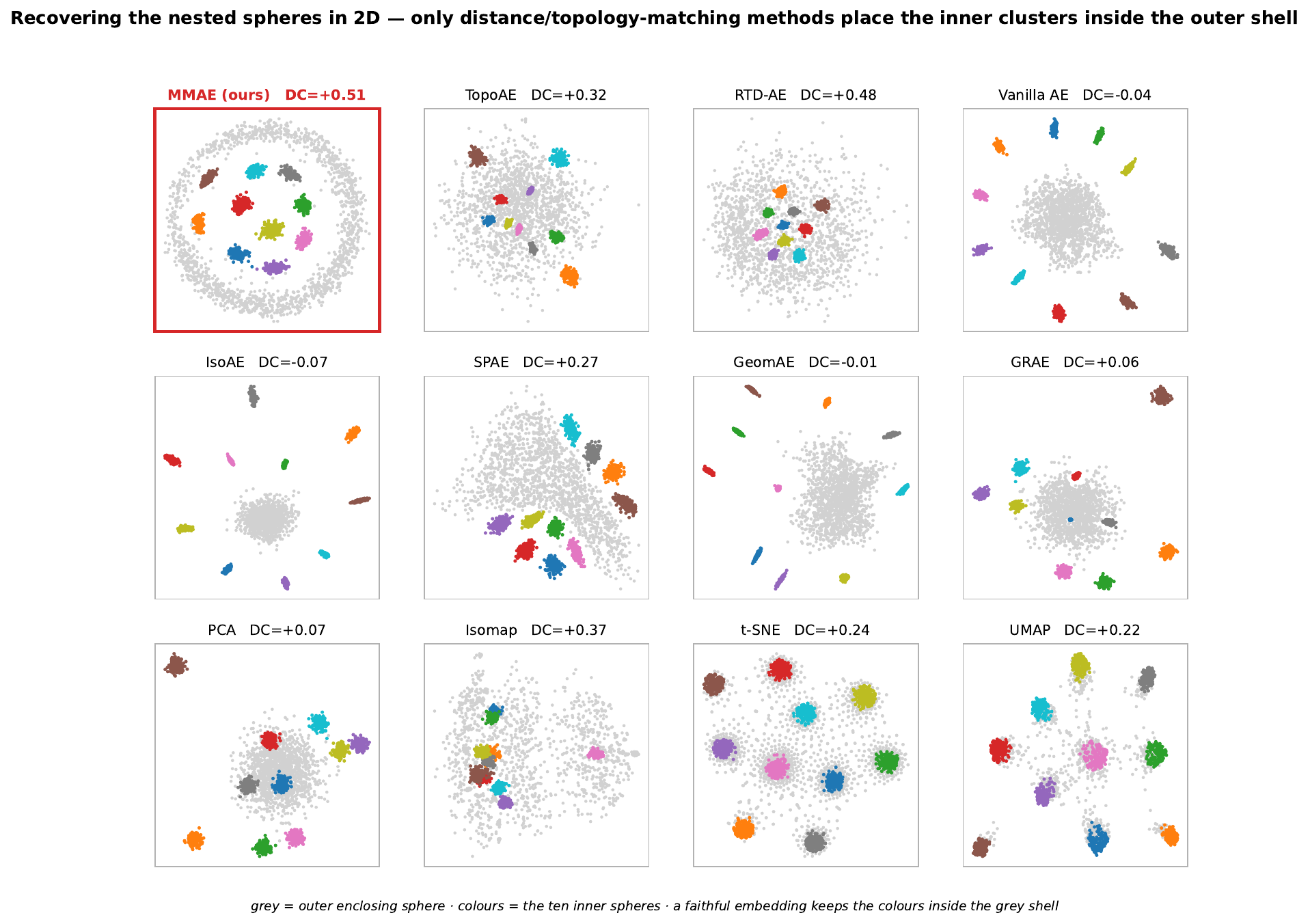}
\caption{\textbf{A distance-matching autoencoder recovers the nesting.} Two-dimensional embeddings of
the nested spheres by twelve methods (grey~=~outer enclosing sphere; colours~=~the ten inner spheres;
distance correlation to the input printed per panel). A faithful embedding keeps the coloured clusters
\emph{inside} the grey shell. MMAE,  matching the full-dimensional input distances,  recovers the
nesting; the only other reference-based method, GRAE, cannot, being tied
to a two-dimensional reference; 
TopoAE and RTD-AE also place the inner spheres in the shell, but without the gap between them;
the remaining autoencoders and
all four non-parametric reducers (PCA, Isomap, t-SNE, UMAP) scatter the inner spheres outside the
enclosure. This is a single illustrative run (seed $0$, $3{,}000$ points, a fixed $\lambda$ per method rather than the tuned configurations; MMAE uses the raw-distance reference); Table \ref{tab:app-DC} reports means over five seeds.}
\label{fig:nested}
\end{figure}

\begin{figure}[!h]
  \centering
  \includegraphics[width=\linewidth]{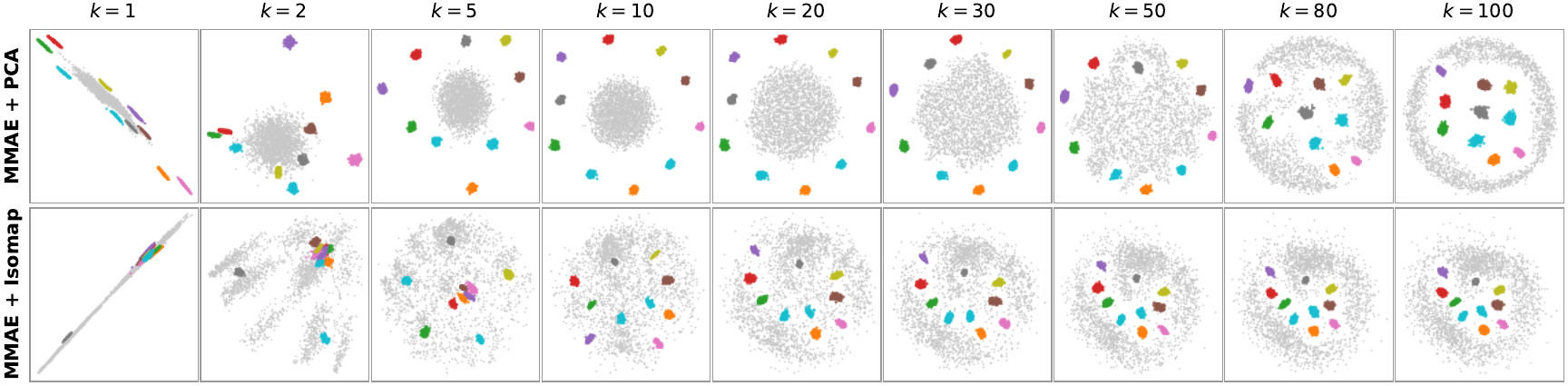}
  \caption{\textbf{More reference components, more nesting.} Two-dimensional MMAE latents on the
  nested spheres (grey = outer enclosing sphere; colours = the ten inner spheres), trained to match
  the pairwise distances of a $k$-dimensional PCA (top) or Isomap (bottom) reference as $k$ grows.}
  \label{fig:ref_ladder}
\end{figure}

The nested-spheres \citep{moor2020topological} is an interesting benchmark due to its simplicity but also difficulty to represent using available algorithms. It places ten small spheres inside a larger enclosing shell (a larger hollow sphere)
in a high-dimensional ambient space; a faithful two-dimensional embedding must keep the coloured
clusters \emph{inside} the outer shell. Figure~\ref{fig:nested} embeds it with twelve methods. It was shown by \citet{moor2020topological, trofimov2023learning} that only autoencoders with a topology-based regularization could retrieve the nesting relationship that exists in the data, and that no other dimensionality reduction methods such as t-SNE, UMAP, PCA, and even vanilla autoencoders would manage to reproduce. While we arrived at the same conclusion as these authors regarding this dataset and other methods, we also challenge the exclusivity of topologically regularized autoencoders in obtaining the correct representation. 
MMAE, matching the \emph{full-dimensional} input distances, recovers the nesting structure while also leaving a clear gap (known to exist) between the outer and inner spheres. We stress that this is a qualitative claim: absolute distance correlations here are low for \emph{every} method (from $0.51$ for MMAE down to near zero, and negative for some autoencoders), so the metric sits close to its floor and the per-panel numbers separate methods only weakly. The quantitative backbone of the claim is instead Figure~\ref{fig:ref_ladder}, where the nesting reappears monotonically as the reference dimension grows. TopoAE and RTD-AE also achieve the nesting relationship but show no gap despite all efforts at hyperparameter tuning. Because MMAE matches the original input distances, which stay fixed across runs, it shows significant robustness, reproducing the same representation repeatedly. RTD-AE also shows robust results with a nesting structure that requires very little hyperparameter tuning to emerge, while TopoAE can sometimes fail due to their approach for topology preservation based on minimum spanning tree, a finding already explored by \citet{trofimov2023learning}. This example also illustrates the differences in coordinate alignment versus distance alignment, as GRAE doesn't manage to reproduce the nesting structure unless it uses an external embedding mechanism which is capable of doing so. Indeed, using PCA with two components as reference or any other mechanism won't generate the correct structure as stated before, limiting this variant. In that sense MMAE combines GRAE's \textit{copying} effect (Section~\ref{sec:reproduce}) with the ability to recover structure directly from the data's own distances, without an external embedding.

Figure~\ref{fig:ref_ladder} makes this concrete by sweeping the reference dimension $k$ for a PCA
and an Isomap reference. With a one-dimensional reference the latent collapses to a line, and with
$k{=}2$ the encoder reproduces the behavior of GRAE: the inner clusters are laid
out \emph{around} the outer sphere rather than within it, and the nesting is lost, a result already shown in \citet{moor2020topological, trofimov2023learning}. 
As $k$ increases, the PCA reference distances approach the ambient distances, and the inner clusters are dragged inside the enclosing shell, recovering the nesting that only TopoAE and RTD-AE reached in  Figure~\ref{fig:nested}. The Isomap reference instead approaches graph-geodesic distances on the $k$-nearest-neighbour graph ($k{=}15$); because the inner spheres and the shell form disconnected components of that graph, the scikit-learn Isomap implementation joins them by adding the shortest Euclidean edge between components, so geodesics between components pass through these bridging edges.
Two points follow. First, the full ambient dimension is never required: the nesting is already faithful well
below it. Second, how many components are needed is governed by how isometric the reference already
is at low dimension. 
Isomap, whose leading coordinates already capture most of its geodesic distance structure, recovers the nesting with only $k\!\approx\!10$--$20$ components;
PCA orders its axes by variance and spends
its leading components on the high-variance outer sphere, encoding the inner-sphere arrangement only
in later, low-variance directions, so it needs $k\!\approx\!80$ to reach the same result. In both
cases the effect is monotone in $k$: MMAE preserves the nesting not through any dedicated persistence
machinery but by inheriting it from a reference that is sufficiently distance-faithful, 
consistent with the intuition behind the stability theorem of \S\ref{sec:theory}, although our $L^2$ objective does not itself guarantee it.

\subsection{Structure preservation in different datasets}
One striking case invites a systematic question: across many datasets, which reference gives the most
faithful latent, and how do we measure ``faithful''? In the previous scenario the topology of the nested spheres is known but for real-world scenarios this is not the case. There may be cases, such as the \textit{swiss roll} one, in which an unrolled representation may be desired, destroying the geometry but preserving the topology, however, we will focus on assessing the preservation of the original input data structure, following previous literature \citep{moor2020topological, trofimov2023learning, nazari2023geomae, singh2021spae}. For that, no single number suffices, so we report four
complementary views which rely all on the comparison of Euclidean distances between the input and the latent. A \emph{global} score, distance correlation (DC)\footnote{Throughout, DC denotes the Pearson correlation between the pairwise distances in the input and in the latent, not the distance correlation of \citet{szekely2007measuring}.} and triplet accuracy (TA), 
captures whether large-scale relations survive; a \emph{local} score,  trustworthiness~$\times$~
continuity (T$\times$C) \citep{venna2001neighborhood}, each averaged over a range of neighbourhood
sizes rather than fixed to one $k$,  captures whether neighbourhoods are preserved; a
\emph{topological} check,  the persistence-Wasserstein distances $W_0,W_1$ between the input and
latent diagrams,  catches loops and voids the correlations miss; and a \emph{density} score,  the $KL_{0.1}$
divergence between kernel-density estimates of input and latent, \citep{moor2020topological},  measures how
evenly the latent is populated. These trade off rather than co-optimise: a latent that preserves
ambient distances (high DC) can place manifold-distant points close together,  the swiss roll again
,  hurting the local and topological scores. Plotting a global against a local axis makes the trade-off
visible directly.

\begin{figure}[t]
\centering
\includegraphics[width=\linewidth]{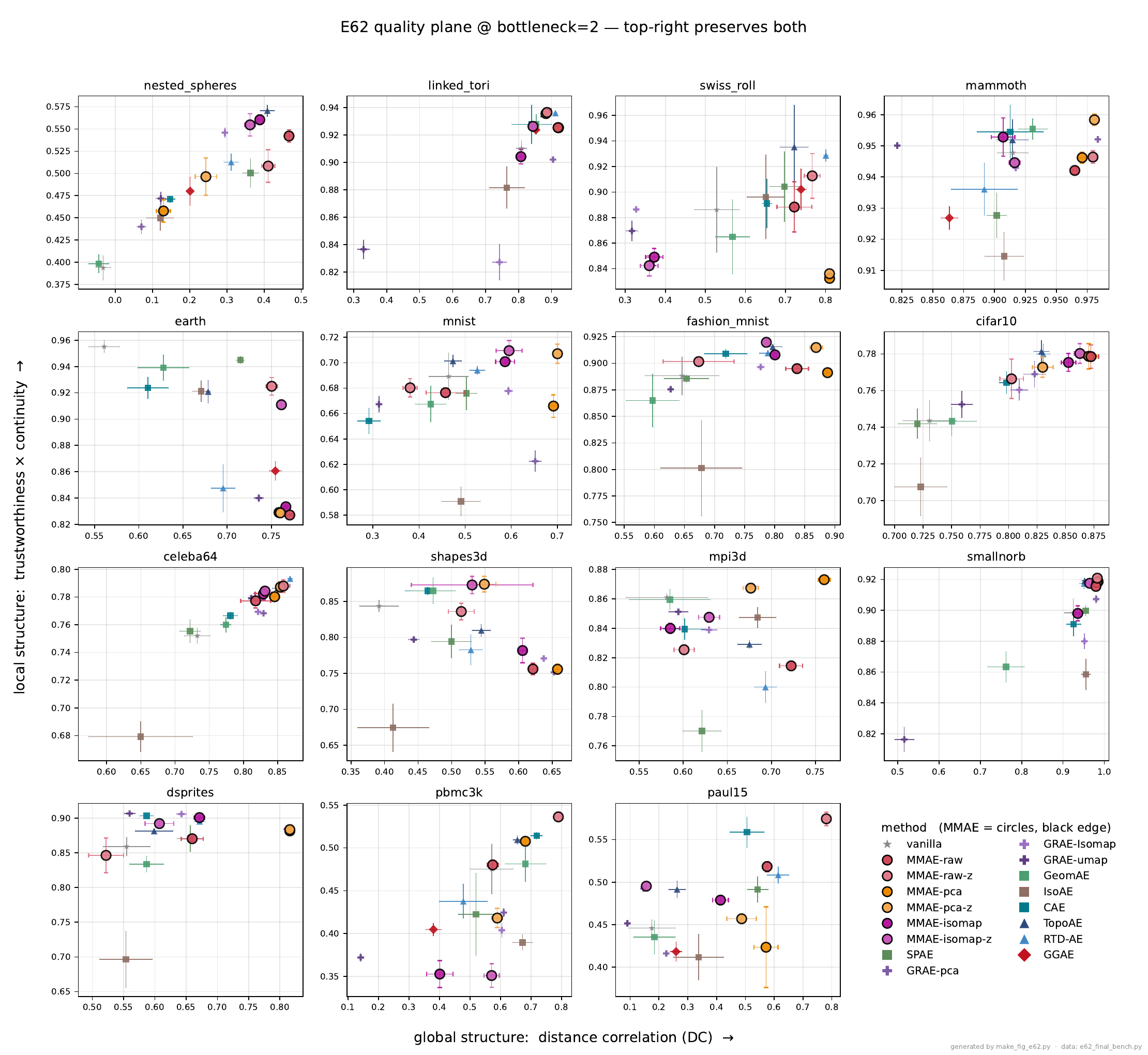}
\caption{\textbf{Quality lives in two dimensions.} Global (distance correlation, DC) versus local
(trustworthiness~$\times$~continuity) preservation at the 2D bottleneck, one panel per dataset (mean
over five seeds, each method under its own tuned configuration). MMAE appears as six variants
(circles) spanning three reference geometries,  raw, PCA, Isomap,  each with and without
normalisation. The global edge is owned by the Euclidean-reference MMAE variants; the local axis is
contested by TopoAE and RTD-AE; the top-right corner preserves both.}
\label{fig:plane}
\end{figure}

\paragraph{Setup.}
We evaluate on fifteen datasets in three families: five synthetic manifolds with known geometry and
topology (\textit{nested spheres} \citep{moor2020topological}, \textit{linked tori}, \textit{swiss roll}, \textit{mammoth} \citep{mnoichl2025mammoth},
\textit{earth} \citep{nazari2023geomae}), eight image collections (MNIST \citep{lecun1998mnist}, Fashion-MNIST \cite{xiao2017fashionmnistnovelimagedataset}, CIFAR-10 \citep{krizhevsky2009learning}, CelebA \citep{liu2015celeba}, 3D-Shapes \citep{3dshapes18}, MPI3D \citep{gondal2019transfer},
smallNORB \citep{lecun-cvpr-2004}, dSprites \citep{dsprites17}), and two single-cell transcriptomics datasets (PBMC3k \citep{10xGenomics2016, Wolf2018scanpy}, Paul15 \citep{Paul2015}). Full specifications and sources are listed in Appendix~\ref{app:datasets}. To keep the
comparison fair every method shares one backbone,  a $256$--$128$ MLP encoder with a mirrored
decoder,  together with per-feature standardisation of the inputs and frozen $80/10/10$
train/validation/test splits; the image datasets are subsampled to a frozen set of $40{,}000$ points shared across methods. All metrics are computed on the held-out test split.

Each (dataset, method) pair is tuned independently with Optuna's TPE sampler
\citep{akiba2019optuna}: $25$ trials at the two-dimensional bottleneck, selecting the configuration
that minimises the validation $KL_{0.1}$ density divergence \citep{moor2020topological}. The winning
configuration is retrained on five seeds, and every reported number is a mean over those seeds on the
blind test split, each metric estimated from five $500$-point subsamples (triplet accuracy over
$5{,}000$ triplets). MMAE enters as six variants,  the factorial of three reference geometries
(raw ambient distances, PCA, and a landmark geodesic Isomap) with and
without latent normalisation; for the PCA and Isomap references the number of components is itself an
Optuna-searched knob (grid $\{2,4,\dots,256\}$, capped below the ambient dimension), letting the data
choose the reference dimension. GRAE, the only other reference-based method, receives a matching
three-reference study (PCA, Isomap, UMAP); because GRAE aligns \emph{coordinates} rather than
distances, each of its references must equal the $2$D bottleneck, whereas MMAE matches distances and
can use a reference of any dimension,  an asymmetry that is itself part of the comparison. The
remaining competitors (vanilla AE, SPAE, TopoAE, RTD-AE, GeomAE, GGAE, IsoAE, CAE) run unchanged;
the persistence-based TopoAE and RTD-AE have their batch capped at $64$ on the large image sets (a tractability constraint: batch size is otherwise a tuned hyperparameter, and Figure~\ref{fig:batch_quality} shows larger batches can help, so this cap may slightly disadvantage them there), and
GGAE, whose $\mathcal{O}(N^2)$ kernel exhausts memory on the larger datasets (due to the bandwidth-$t$ kernel it must build before training), is reported only where
it fits (Appendix~\ref{app:full-metrics}).

\begin{table}[t]
\centering
\small
\setlength{\tabcolsep}{4.5pt}
\caption{\textbf{Where each method's strengths lie.} Mean rank across the $15$ datasets at the
2D bottleneck (lower is better; \textbf{\underline{best}}, \underline{second}). No method wins
everywhere: MMAE leads the global distance metrics (DC, TA) and the overall average, the topological
methods lead locally (T$\times$C), the $W_1$ and density columns go to decoder- and graph-based methods, and the $W_0$ column shows no significant differences across methods (\S\ref{sec:eval}).
Because DC and TA are near-proxies for MMAE's training objective, its lead there is partly by
construction (see text). Ranks are over the 16 method configurations (GGAE excluded for missing
datasets). Per-metric results are reported in the appendix.}
\label{tab:rank}
% mean-rank table @ bottleneck=2; GGAE excluded (OOM subset); 15 datasets
\begin{tabular}{@{}lccccccc@{}}
\toprule
Method & DC & TA & T$\times$C & W$_0$ & W$_1$ & KL$_{0.1}$ & Avg \\
\midrule
MMAE-raw & \underline{4.80} & \underline{3.87} & 9.13 & 9.07 & 10.27 & 7.53 & 7.44 \\
MMAE-raw-z & 6.93 & 4.27 & \underline{5.73} & 6.47 & 7.13 & 8.20 & \textbf{\underline{6.46}} \\
MMAE-pca & \textbf{\underline{3.07}} & \textbf{\underline{3.60}} & 9.33 & 10.20 & 10.27 & \underline{6.73} & 7.20 \\
MMAE-pca-z & \underline{4.80} & 5.40 & 6.33 & 8.07 & 7.27 & 7.53 & \underline{6.57} \\
MMAE-Isomap & 8.53 & 8.60 & 8.20 & 10.80 & 11.27 & 6.87 & 9.04 \\
MMAE-Isomap-z & 8.07 & 8.33 & 6.20 & 8.20 & 7.20 & 8.07 & 7.68 \\
\midrule
Vanilla AE & 13.40 & 13.00 & 9.47 & \textbf{\underline{5.60}} & 5.87 & 13.00 & 10.06 \\
SPAE & 9.87 & 8.53 & 9.67 & 9.00 & 13.07 & 9.60 & 9.96 \\
GRAE-pca & 5.20 & 6.47 & 10.80 & 9.20 & 6.93 & 8.13 & 7.79 \\
GRAE-Isomap & 8.80 & 9.53 & 9.60 & 10.07 & 6.73 & 7.20 & 8.66 \\
GRAE-umap & 14.00 & 14.07 & 10.27 & 11.27 & 7.67 & \textbf{\underline{5.40}} & 10.44 \\
TopoAE & 7.73 & 7.00 & \textbf{\underline{4.73}} & 9.73 & 11.87 & 9.07 & 8.36 \\
RTD-AE & 6.87 & 7.13 & 6.33 & 7.67 & 12.73 & 9.33 & 8.34 \\
GeomAE & 12.00 & 12.67 & 9.67 & \underline{5.73} & \underline{5.00} & 7.67 & 8.79 \\
IsoAE & 11.20 & 12.53 & 13.47 & 6.80 & \textbf{\underline{4.93}} & 10.80 & 9.96 \\
CAE & 10.73 & 11.00 & 7.07 & 8.13 & 7.80 & 10.87 & 9.27 \\
\bottomrule
\end{tabular}
\end{table}

\paragraph{Results}
Table~\ref{tab:rank} ranks every method by its mean rank across the fifteen datasets. Before reading
this as a leaderboard, one alignment must be stated: distance correlation and, more loosely, triplet
accuracy are near-proxies for MMAE's own training objective (matching input--latent distances), so
MMAE leading \emph{those} columns is partly by construction. The informative signal is therefore not
that MMAE tops DC/TA, but how it fares on the axes its objective does \emph{not} target,  local
neighbourhoods (T$\times$C), topology ($W_0,W_1$) and density (KL),  and how the reference choice
trades these off. In terms of DC
and TA which measure the preservation of the global structure of the input, MMAE variants (with
exception of Isomap) outperform all other approaches,  led by the non-normalized MMAE-pca and
MMAE-raw, though on DC the normalized raw variant is edged out by GRAE-pca and RTD-AE. GRAE-pca follows, then the topological variants: PCA's leading directions already capture much of the
global structure. For local neighbourhood preservation, TopoAE takes first place and MMAE second, 
notable since MMAE does not target neighbourhoods,  underlining that the reference choice shapes the
resulting latent. In terms of the Wasserstein distances the picture is
mixed: a Friedman test finds no significant differences across methods on $W_0$ ($p=0.21$), so the
$H_0$ column should not be read as a ranking, whereas $W_1$ does differ significantly and is led by the
decoder-geometry methods. 
Additionally we note that using a normalized version of MMAE tends to lead to better results in terms
of $W_0, W_1$, and $T\times C$, while a global structure preservation is better maintained when not
using it. On density, GRAE-UMAP leads,  likely due to the UMAP reference rather than the regulariser, since
the other GRAE variants do not. Overall, the two normalized MMAE variants (PCA and raw input
distances) achieve the best average rank: raw distances capture all the geometry, and PCA preserves
the global structure while dropping noise.
By inspecting Figure~\ref{fig:plane} this is confirmed as both pbmc3k and paul15 which are
single-cell datasets are known to have dimensions of little variance that are of high importance in
data analysis, hence, cannot be ignored, and we see the MMAE-raw-z variant taking the lead. In other
scenarios, more precisely images, we see that in general the MMAE-pca variants take the lead, also expected as several dimensions (pixels) can be considered as noise. A case
that is interesting to observe is the swiss roll one as, even though the unrolled representation of
the dataset given by both GRAE and MMAE isomap variants is the most desirable, it performs worst on
these axes. The local neighborhood isn't far behind the other variants, but the global structure is
significantly affected. This shows the right metric depends on the task: unfolding the swiss roll helps interpolation but not
visualising its folded nature. Taken together, these results characterise MMAE less as a universal winner than as a controllable one: with a distance-faithful reference it holds the global axis by construction while remaining competitive on the local and topological axes its objective does not directly optimise, and the reference is the knob that moves it along the trade-off.

\begin{figure}[t]
\centering
\includegraphics[width=0.82\linewidth]{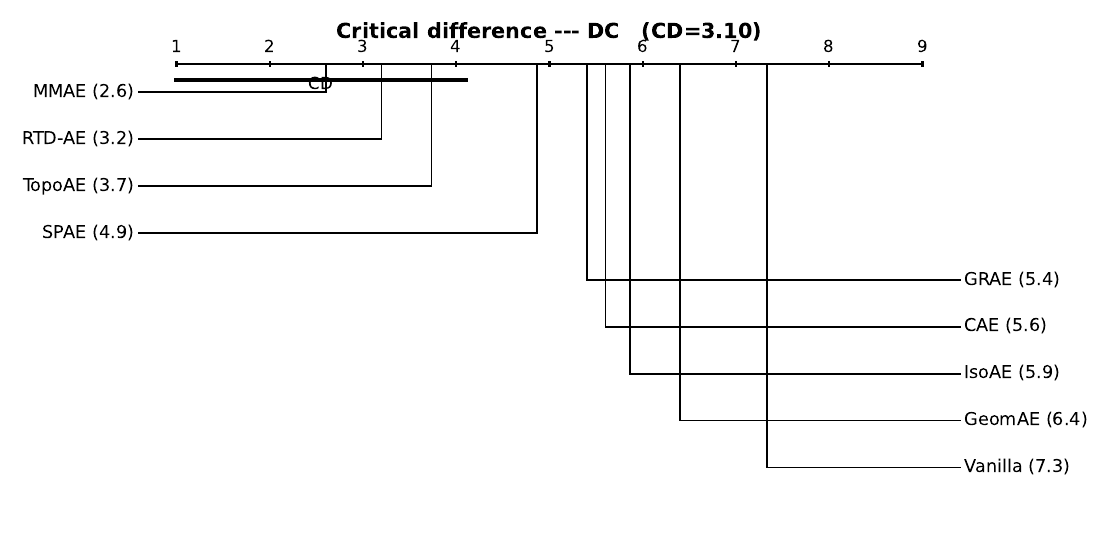}
\caption{\textbf{Critical-difference diagram (distance correlation).} Methods placed by mean rank
across the $15$ datasets (one configuration each); the bar is the Nemenyi critical difference
($\alpha{=}0.05$, CD$\,{=}\,3.10$). MMAE, RTD-AE and TopoAE have the three best mean ranks; MMAE is
significantly ahead of IsoAE, GeomAE and Vanilla, while its differences from SPAE, GRAE and CAE fall
within the CD.}
\label{fig:cd}
\end{figure}

\paragraph{Is the ranking fair, and is it significant?}
Two concerns about Table~\ref{tab:rank} deserve a direct answer: MMAE enters as six tuned variants
while each competitor enters as one, and a mean rank is not a significance test. We therefore re-rank
with every family collapsed to a \emph{single} configuration per dataset,  the variant minimising
the same validation $KL_{0.1}$ used throughout, so MMAE and GRAE each contribute exactly one entry
(Table~\ref{tab:rank_fair}),  and run a Friedman omnibus test with a Nemenyi critical-difference
(CD) post-hoc across the fifteen datasets, plus pairwise Wilcoxon signed-rank tests. Three findings.
First, the variant count does not explain MMAE's position: even as a single $KL_{0.1}$-selected config
it keeps the best average rank and leads DC and TA. Because that entry is still the best of six
variants under validation KL$_{0.1}$ (GRAE: best of three), we also apply a stricter check that fixes a
single variant (MMAE-raw-z) on every dataset: MMAE then keeps the best average rank ($3.63$, vs.\
$4.57$ for TopoAE and $4.58$ for RTD-AE) and the best TA rank ($1.87$), while RTD-AE edges it on DC
($3.00$ vs.\ $3.27$) and its KL$_{0.1}$ lead disappears ($4.33$ vs.\ $2.00$ for GRAE), so that
advantage in Table~\ref{tab:rank_fair} reflects variant selection. Second, the differences are real where it matters
,  the Friedman test is highly significant for DC, TA, T$\times$C, $W_1$ and KL ($p<10^{-4}$) but
\emph{not} for $W_0$ ($p=0.21$), so no method is separable on the $H_0$ persistence-Wasserstein and
its apparent ``winner'' is not meaningful. Third, pairwise Wilcoxon signed-rank tests, Holm-corrected over the three comparisons per metric,
place MMAE at the top without over-claiming: it significantly outperforms GRAE (DC $p{=}0.020$, TA
$p{=}0.006$) and TopoAE (TA $p{=}0.021$, $W_1$ $p{=}0.043$, KL $p{=}0.001$; not DC, $p{=}0.071$), and
is statistically \emph{indistinguishable} from RTD-AE on DC/TA/T$\times$C while beating it on
$W_1$ ($p{=}0.037$) and density ($p{<}0.001$).
By the conservative Nemenyi CD (Figure~\ref{fig:cd}), MMAE, RTD-AE and TopoAE have the best mean ranks on DC, and MMAE is significantly ahead of IsoAE, GeomAE and the vanilla autoencoder; its differences from SPAE, GRAE and CAE fall within the CD.

\begin{table}[t]
\centering\small
\caption{\textbf{Fair re-ranking: one configuration per method.} Mean rank across the $15$ datasets at
the 2D bottleneck (lower is better; \textbf{\underline{best}}, \underline{second}) with MMAE and GRAE
each collapsed to a single $KL_{0.1}$-selected configuration (GGAE excluded for missing datasets).
MMAE keeps the best average even as a single entry.}
\label{tab:rank_fair}
\begin{tabular}{@{}lccccccc@{}}
\toprule
Method & DC & TA & T$\times$C & $W_0$ & $W_1$ & KL$_{0.1}$ & Avg \\
\midrule
MMAE & \textbf{\underline{2.60}} & \textbf{\underline{1.87}} & 4.53 & 5.27 & 4.73 & \textbf{\underline{2.13}} & \textbf{\underline{3.52}} \\
TopoAE & 3.73 & \underline{3.33} & \textbf{\underline{2.80}} & 5.80 & 6.73 & 5.20 & \underline{4.60} \\
RTD-AE & \underline{3.20} & 3.47 & \underline{3.87} & 5.00 & 7.13 & 5.27 & 4.66 \\
GRAE & 5.40 & 5.40 & 5.60 & 6.07 & 4.20 & \underline{2.20} & 4.81 \\
GeomAE & 6.40 & 6.80 & 5.53 & \textbf{\underline{3.73}} & \textbf{\underline{3.13}} & 4.67 & 5.04 \\
CAE & 5.60 & 5.80 & 4.20 & 5.20 & 4.67 & 6.27 & 5.29 \\
SPAE & 4.87 & 4.40 & 5.33 & 5.60 & 7.60 & 5.33 & 5.52 \\
IsoAE & 5.87 & 6.87 & 7.93 & 4.47 & \textbf{\underline{3.13}} & 6.33 & 5.77 \\
Vanilla & 7.33 & 7.07 & 5.20 & \underline{3.87} & \underline{3.67} & 7.60 & 5.79 \\
\bottomrule
\end{tabular}
\end{table}

\paragraph{Does the reconstruction term matter?}
To isolate what the autoencoder buys over a bare distance-matching encoder, we compare three
configurations under one fixed operating point (raw reference, $\lambda{=}1$, identical fair frame),
each retrained on three seeds and averaged over all fifteen datasets (Table~\ref{tab:ablation}): the
full MMAE ($\mathcal{L}_{\mathrm{rec}}+\lambda\mathcal{R}_{\mathrm{MM}}$), a \emph{matching-only}
encoder ($\mathcal{R}_{\mathrm{MM}}$ alone,  a parametric, batched MDS), and a vanilla autoencoder
($\mathcal{L}_{\mathrm{rec}}$ alone). Two things stand out. First, the reconstruction term is nearly
neutral for the latent's geometry: matching-only reproduces full MMAE within noise on every metric
(DC $0.743$ vs.\ $0.743$, T$\times$C $0.775$ vs.\ $0.779$, $W_1$ $3.80$ vs.\ $3.80$, KL $0.111$ vs.\
$0.110$). Second, the matching term is what does the geometric work: removing it (vanilla) drops DC
from $0.743$ to $0.568$, TA from $0.79$ to $0.71$ and density KL from $0.11$ to $0.22$ (only the
topological $W_0$ and $W_1$ marginally favour vanilla). The latent geometry is thus produced by the encoder's
distance matching, not by reconstruction; we retain the decoder not for latent quality but because it
leaves the geometry unchanged. The converse does not hold: matching does cost some reconstruction. The
final training reconstruction error on standardised inputs rises from $0.43$ (vanilla) to $0.51$ with
MMAE, a trade-off governed by $\lambda$ (fixed at $1$ here), while matching-only trains no decoder and
stays at $0.98$, close to the input variance. On the metrics reported here the decoder
therefore does no geometric work, and we do not claim otherwise: MMAE is best read as a
\emph{regulariser} that adds geometric control to a standard autoencoder at negligible computational cost, and when only the embedding is needed the same loss defines a parametric
distance-matching encoder (the matching-only column, i.e.\ a batched MDS). We keep the autoencoder
framing because MMAE is a drop-in term for the reconstruction pipelines practitioners already use, not
because the decoder shapes the latent; the generative payoff that would exploit the decoder is left to
future work.

\begin{table}[t]
\centering\small
\caption{\textbf{The reconstruction term is near-neutral for latent geometry.} Mean over three seeds
and all fifteen datasets. Matching-only (a parametric distance-matching encoder) equals full MMAE on
every geometric metric; removing the matching term (vanilla) collapses global and density fidelity.
Rec.: final-epoch training reconstruction MSE on standardised inputs (matching-only trains no decoder).}
\label{tab:ablation}
\begin{tabular}{@{}lccccccc@{}}
\toprule
Configuration & DC$\uparrow$ & TA$\uparrow$ & T$\times$C$\uparrow$ & $W_0\downarrow$ & $W_1\downarrow$ & KL$_{0.1}\downarrow$ & Rec.$\downarrow$ \\
\midrule
Vanilla (recon only) & 0.568 & 0.711 & 0.770 & 113.059 & 3.544 & 0.216 & 0.426 \\
Matching-only (parametric MDS) & 0.743 & 0.793 & 0.775 & 114.467 & 3.797 & 0.111 & 0.980 \\
MMAE (recon + match) & 0.743 & 0.792 & 0.779 & 114.504 & 3.797 & 0.110 & 0.512 \\
\bottomrule
\end{tabular}
\end{table}

\section{Discussion \& Conclusion}
\label{sec:discussion}
\subsection{Discussion}
\paragraph{Computational scaling \& Stability}
\label{sec:scaling}
One of the problems in imposing geometric structure on autoencoders is that it frequently comes with
a large computational overhead. Two factors drive this cost: the batch size $b$ and the ambient data
dimensionality $n$. Topology-based methods such as TopoAE and RTD-AE build a persistence filtration
from the $b\times b$ batch distance matrix (the same input as MM-reg; Section~\ref{sec:theory}), so
their cost is dominated by $b$. Decoder-geometry methods such as GeomAE and IsoAE act instead on the
decoder, and are therefore sensitive to the ambient dimension $n$, since the decoder's output,  and
hence its Jacobian,  lives in $\R^n$. GRAE, SPAE, GGAE and MMAE scale essentially like a plain
autoencoder during training: their regularisers touch the ambient dimension only through the network
itself, so all four can be applied to larger datasets and larger batches without appreciable overhead
,  with the caveat that GGAE must first assemble its $\mathcal{O}(N^2)$ bandwidth kernel over (much
of) the data \emph{before} training begins, a one-time bottleneck that in practice limits it to
smaller datasets.
\begin{figure}[!h]
\centering
\includegraphics[width=0.9\linewidth]{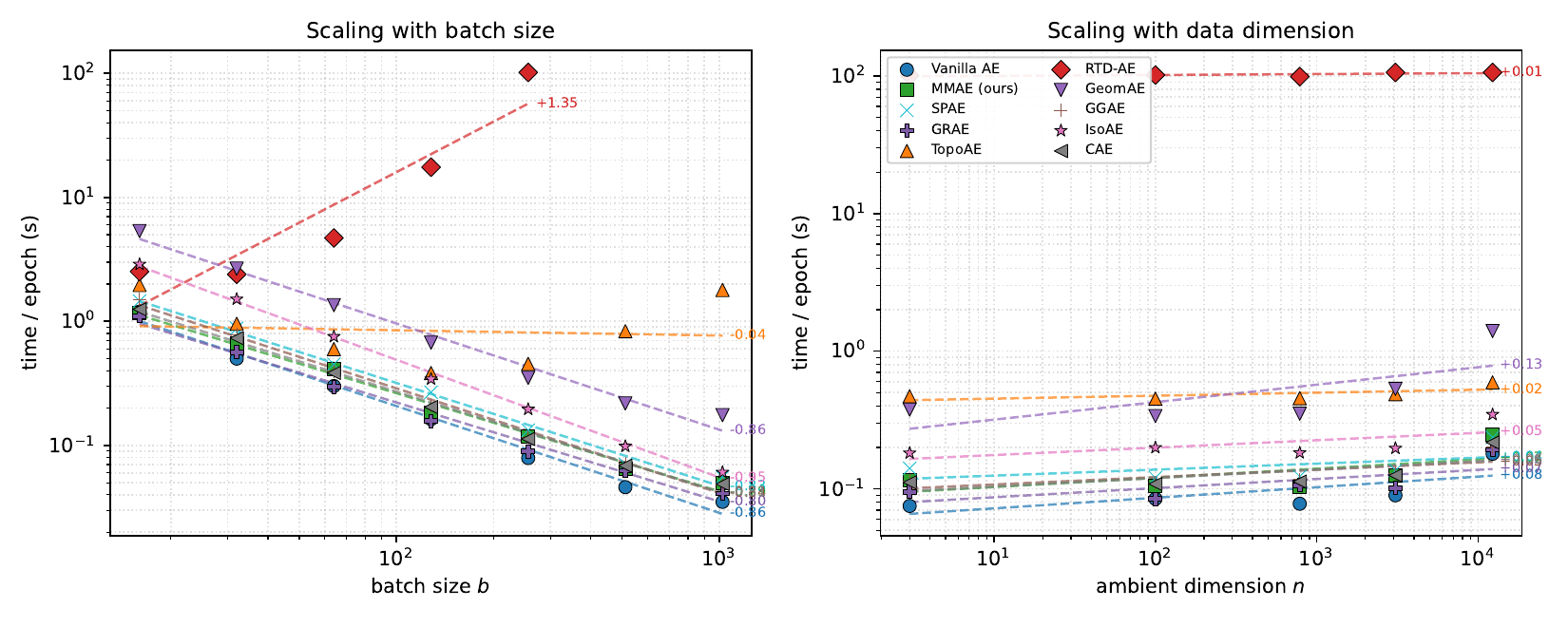}
\caption{\textbf{Computational scaling.} Mean training time per epoch (one warmup epoch dropped,
three averaged; reference extraction and kernel/precompute excluded), log--log, with the fitted
scaling exponent annotated per method. \emph{Left:} versus batch size on MNIST. \emph{Right:} versus
ambient dimension ($3$ to $12{,}288$).}
\label{fig:scaling}
\end{figure}

Figure~\ref{fig:scaling} reports the training time of a single epoch for each autoencoder as a
function of batch size and of ambient dimension, timing only the training epochs (reference
extraction and any kernel/precompute are excluded). The manifold-matching term is a batched
pairwise-distance MSE computed on the latent codes, which costs no more than the reconstruction pass
itself and is independent of $n$; MMAE therefore inherits the cost profile of a plain autoencoder, 
its time per epoch falls with batch size at essentially the vanilla rate and grows only marginally
with ambient dimension, from the $3$-dimensional swiss roll to the $12{,}288$-dimensional CelebA
images. The persistence-based methods behave very differently. RTD-AE's per-epoch time climbs steeply
with batch size,  a persistence diagram must be recomputed each mini-batch, and that computation is
super-quadratic in $b$,  making it roughly two orders of magnitude slower in absolute terms and the
reason we cap it at $b{=}256$ in this timing study. TopoAE, by contrast, gains nothing from larger batches: its
minimum-spanning-tree signal keeps its cost roughly flat in $b$, but it remains the next most
expensive method overall. Along the dimension axis the ordering flips. Most methods are stressed
little by $n$, which enters only through the encoder's first layer; the exception is GeomAE, which
grows fastest with the output dimension. Its regulariser penalises the variance of
$\log(\det (J_g^\top J_g))$, so each step must form the decoder Jacobian $J_g(z)\in\R^{n\times d}$ and the
pullback metric $J_g^\top J_g$ at every latent point, and then back-propagate through that Jacobian
,  a second-order computation whose cost carries $n$ several times over. IsoAE avoids this: it
enforces a local isometry with a single random Jacobian--vector product rather than the full Jacobian
and determinant, so despite acting on the decoder it scales like a plain autoencoder. In short, MMAE
provides explicit geometric control at essentially the price of reconstruction, whereas the
topological alternatives pay a large, batch-scaling overhead and the decoder-determinant regulariser
a dimension-scaling one,  which is what makes the reference-selection studies above practical at
scale.

One caveat on ``cheap.'' The timings above cover only the training epochs and deliberately exclude
\emph{reference construction}, which is a genuine one-time cost paid before training. For raw and PCA
references it is negligible, but the geodesic and neighbour-graph references (Isomap, UMAP, t-SNE,
PHATE) require building a $k$-NN graph over the full dataset ($O(N\log N)$ to $O(N^2)$),  comparable
to running those methods once. Because the reference is computed a single time and reused across every
bottleneck, component count and value of $\lambda$, this cost amortises over the many configurations we
sweep; but it is real, and the cheapness claim applies to the per-epoch \emph{training} term, not to a
t-SNE/UMAP reference pipeline taken as a whole.

\begin{figure}[!h]
\centering
\includegraphics[width=0.9\linewidth]{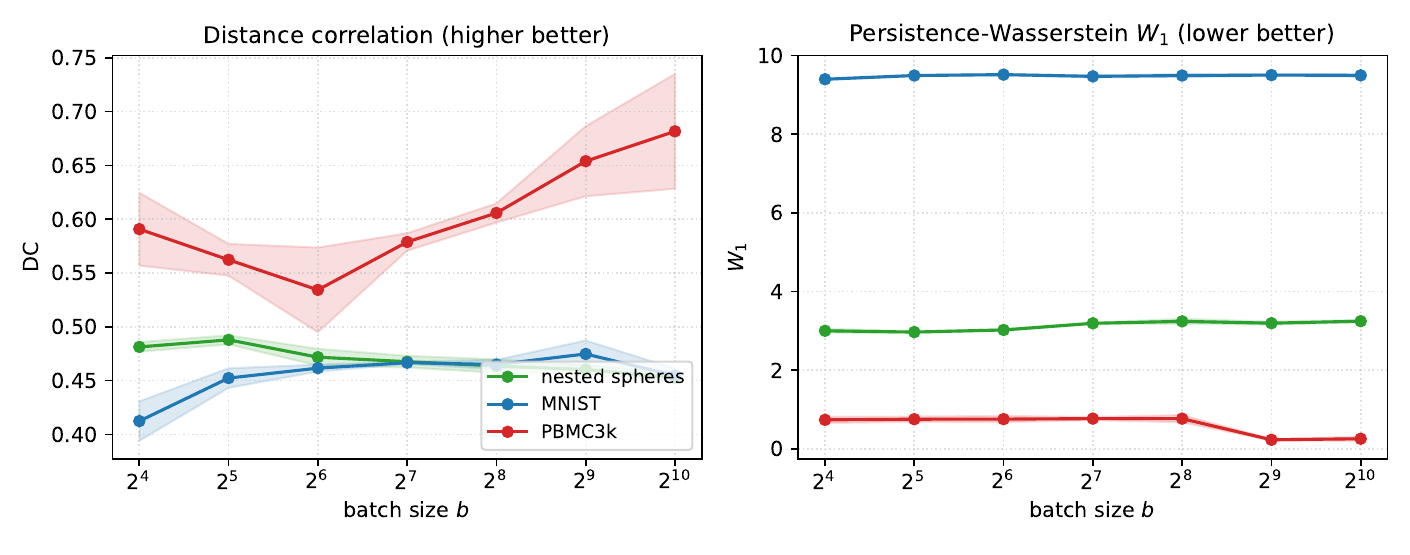}
\caption{\textbf{Fidelity is robust to batch size.} MMAE (raw reference) retrained to convergence
across batch sizes $b\in\{16,\dots,1024\}$ on three datasets, mean over three seeds (bands $\pm$1 s.d.).
\emph{Left:} distance correlation (higher better). \emph{Right:} persistence-Wasserstein $W_1$ (lower
better). Both are broadly flat or improving with $b$ (PBMC3k dips at $b{=}64$); the smallest batch costs
nothing on nested spheres, about $0.06$ DC on MNIST and about $0.09$ on PBMC3k.}
\label{fig:batch_quality}
\end{figure}

\paragraph{Fidelity is robust to batch size.}
The cheapness argument assumes the $b\times b$ batch distance matrix is an adequate proxy for global
structure (\S\ref{sec:theory}). Figure~\ref{fig:batch_quality} tests this directly: we retrain MMAE
(raw reference) to convergence across batch sizes from $16$ to $1024$ on a synthetic, an image, and a
single-cell dataset, and measure distance correlation and the persistence-Wasserstein $W_1$. Quality
does not require large batches. Distance correlation is flat or \emph{improving} with $b$,  nested
spheres varies by $0.03$ over the whole range, MNIST by $0.06$, and PBMC3k rises from $0.59$ to
$0.68$,  while the topological $W_1$ is flat throughout, if anything falling at large $b$. On nested
spheres $b{=}16$ is the best setting, on MNIST it is $0.06$ below the best, and PBMC3k shows the
largest gap ($0.59$ at $b{=}16$ vs.\ $0.68$ at $b{=}1024$).
Single-cell data benefits most from larger batches, its many low-variance directions being better
sampled, but never collapses. The batch approximation is therefore not a liability,  MMAE reaches its
fidelity with the same cheap small batches that make it fast, rather than needing large ones that would
erode the scaling advantage.

\paragraph{When preserving geometry does not help}
\label{sec:whennot}
Structure preservation is not universally desirable. Some tasks are actively hurt
by keeping the input geometry,  classification is the clearest example. In the
nested-spheres case, preserving the geometry keeps the inner and outer classes
concentric, so the decision boundary a classifier must learn is a complex, closed curve;
a representation free to tear the manifold apart could separate the classes with a
straight line. This reflects a general fact: a well-trained classifier deliberately
\emph{destroys} topology, driving each class's Betti numbers toward trivial values as
depth increases, because separating classes requires pulling an entangled manifold apart
\citep{naitzat2020topology}. We therefore scope MMAE to settings where the latent is meant
to be a faithful map of the data,  visualisation and dimensionality reduction,  and make no claim
for supervised, task-transforming representations. Interpolation, generation and higher-dimensional
latents motivate the approach but are not evaluated here; we treat them strictly as future work.
\paragraph{Future Work}
A recent wave of work has shed light on the importance of the structure of the latent space for
generative tasks. For example, \citet{xu2024Assessing} show that generative models such as GANs
\citep{goodfellow2014generativeadversarialnetworks}, VAEs \citep{kingma2022VAE}, and LDMs
\citep{rombach2022highresolutionimagesynthesislatent} produce better output when the latent code comes
from a high-density region (a region where several training examples lie in proximity),  a result
that may appear obvious but is confirmed and formalised in their work. \citet{chadebec2022geometric}
adopts a geometric interpretation of the latent space of VAEs and shows that sampling improves
significantly when the local curvature of the latent space is taken into account,  in other words,
sampling by following the shape of the latent space. LDMs have achieved state-of-the-art results in
high-resolution image generation, and recent work has tackled their slow convergence and
controllability by restructuring the latent space of the autoencoders \citep{xu2023geometric,
chen2025dc}. That merely \emph{acknowledging} the shape of the latent space already improves
generative capabilities motivates a natural question: what happens if we not only acknowledge but
actively \emph{manipulate} that shape? And how would a regulariser such as Manifold Matching behave in
the higher-dimensional bottlenecks these generative tasks require, beyond the 2D setting studied here?
We see this as a promising direction for future work.

\subsection{Conclusion}
In this work we presented Manifold Matching as a single, cheap, unsupervised regulariser that turns an
autoencoder's latent space into a faithful map of a \emph{chosen} geometry: the reference is the
lever, and its type and dimension are the design choice. We showed that this one instrument does two
complementary things,  first, it reproduces external embeddings (e.g.\ t-SNE, UMAP, Isomap, PHATE)
and extends them to new points (\S\ref{sec:reproduce}); and second, taking the data's own distances as
the reference, or a near-isometric projection, it qualitatively recovers global structure such as the
nested-spheres nesting that previously required dedicated topological machinery, matching the
global-structure frontier of the persistence-based methods across synthetic, image and single-cell
data,  at essentially the cost of reconstruction (\S\ref{sec:scaling}). The
payoff is a \emph{characterisation} rather than a leaderboard win: no reference preserves every kind of
structure at once, and the value of MMAE is that a single knob selects which structure the latent
inherits. Thanks to its small computational overhead and the decoupling of bottleneck size from
reference dimensionality, we expect the regulariser to extend naturally to larger bottlenecks (beyond
the 2D setting studied here), where it could shape the higher-dimensional latent spaces that
generative models rely on,  a direction we leave to future work.

\subsubsection*{Broader Impact Statement}
Manifold Matching is a dimensionality-reduction regulariser, and its societal risks are those common
to visualisation and representation learning. Faithful low-dimensional maps aid interpretation but can
also mislead when over-interpreted,  distances in any two-dimensional embedding remain a lossy
summary,  and we caution against treating a preserved geometry as ground truth. When used to extend
embeddings of sensitive data (e.g.\ single-cell or biometric datasets), the privacy and dataset-bias
considerations that apply to the underlying embedding apply equally here. We foresee no direct harmful
application specific to the method.

\subsubsection*{Code Availability}
Code, configurations and the per-seed results needed to rebuild every table and figure are available at
\url{https://github.com/laurent-cheret/manifold-matching-autoencoders}.

\subsubsection*{Acknowledgments}
This work was supported by the National Research Council Canada (NRC) through the Artificial Intelligence
for Design (AI4D) Challenge program, project AI4D-161, \emph{Topology and Transformers in tandem: developing
methodology to combine these cutting-edge tools}. This research was enabled in part by support provided by
Calcul Qu\'ebec (\url{https://www.calculquebec.ca}) and the Digital Research Alliance of Canada
(\url{https://alliancecan.ca}).

\bibliography{tmlr}

@inproceedings{bengio-oosExtensionIsomap2003neurips,
 author = {Bengio, Yoshua and Paiement, Jean-fran\c{c}cois and Vincent, Pascal and Delalleau, Olivier and Roux, Nicolas and Ouimet, Marie},
 booktitle = {Advances in Neural Information Processing Systems},
 editor = {S. Thrun and L. Saul and B. Sch\"{o}lkopf},
 pages = {},
 publisher = {MIT Press},
 title = {Out-of-Sample Extensions for LLE, Isomap, MDS, Eigenmaps, and Spectral Clustering},
 url = {https://proceedings.neurips.cc/paper_files/paper/2003/file/cf05968255451bdefe3c5bc64d550517-Paper.pdf},
 volume = {16},
 year = {2003}
}

@article{webb1995pattern,
title = {Multidimensional scaling by iterative majorization using radial basis functions},
journal = {Pattern Recognition},
volume = {28},
number = {5},
pages = {753-759},
year = {1995},
issn = {0031-3203},
doi = {https://doi.org/10.1016/0031-3203(94)00135-9},
url = {https://www.sciencedirect.com/science/article/pii/0031320394001359},
author = {Andrew R. Webb},
}

@ARTICLE{maojain1995tnn,
  author={Kraaijveld, M.A. and Mao, J. and Jain, A.K.},
  journal={IEEE Transactions on Neural Networks}, 
  title={A nonlinear projection method based on Kohonen's topology preserving maps}, 
  year={1995},
  volume={6},
  number={3},
  pages={548-559},
  doi={10.1109/72.377962}}

@article{adamaszek2027pacific,
  title   = {The Vietoris–Rips complexes of a circle},
  author  = {Adamaszek, Michał and Adams, Henry},
  journal = {Pacific Journal of Mathematics},
  volume  = {290},
  number  = {1},
  pages   = {1--40},
  year    = {2017},
  url     = {https://msp.org/pjm/2017/290-1/p01.xhtml},
  doi     = {https://doi.org/10.2140/pjm.2017.290.1}
}

@article{naitzat2020topology,
  title   = {Topology of Deep Neural Networks},
  author  = {Naitzat, Gregory and Zhitnikov, Andrey and Lim, Lek-Heng},
  journal = {Journal of Machine Learning Research},
  volume  = {21},
  number  = {184},
  pages   = {1--40},
  year    = {2020},
  url     = {https://jmlr.org/papers/v21/20-345.html}
}

@inproceedings{moor2020topological,
  title     = {Topological Autoencoders},
  author    = {Moor, Michael and Horn, Max and Rieck, Bastian and Borgwardt, Karsten},
  booktitle = {Proceedings of the 37th International Conference on Machine Learning (ICML)},
  series    = {PMLR},
  volume    = {119},
  pages     = {7045--7054},
  year      = {2020},
  url       = {https://proceedings.mlr.press/v119/moor20a.html}
}

@inproceedings{trofimov2023learning,
  title     = {Learning Topology-Preserving Data Representations},
  author    = {Trofimov, Ilya and Cherniavskii, Daniil and Tulchinskii, Eduard
               and Balabin, Nikita and Burnaev, Evgeny and Barannikov, Serguei},
  booktitle = {International Conference on Learning Representations (ICLR)},
  year      = {2023},
  url       = {https://openreview.net/forum?id=lIu-ixf-Tzf}
}

@article{carlsson2009topology,
  title   = {Topology and Data},
  author  = {Carlsson, Gunnar},
  journal = {Bulletin of the American Mathematical Society},
  volume  = {46},
  number  = {2},
  pages   = {255--308},
  year    = {2009}
}

@article{moon2019phate,
  title   = {Visualizing structure and transitions in high-dimensional biological data},
  author  = {Moon, Kevin R. and van Dijk, David and Wang, Zheng and Gigante, Scott
             and Burkhardt, Daniel B. and Chen, William S. and Yim, Kristina
             and van den Elzen, Antonia and Hirn, Matthew J. and Coifman, Ronald R.
             and Ivanova, Natalia B. and Wolf, Guy and Krishnaswamy, Smita},
  journal = {Nature Biotechnology},
  volume  = {37},
  number  = {12},
  pages   = {1482--1492},
  year    = {2019},
  note    = {PHATE}
}

@article{chadebec2022geometric,
  title   = {A Geometric Perspective on Variational Autoencoders},
  author  = {Chadebec, Cl{\'e}ment and Allassonni{\`e}re, St{\'e}phanie},
  journal = {arXiv preprint arXiv:2209.07370},
  year    = {2022}
}

@misc{shi2016convergencelaplacian,
      title={Convergence of the Laplace-Beltrami Operator from Point Cloud}, 
      author={Zuoqiang Shi and Jian Sun},
      year={2016},
      eprint={1403.2141},
      archivePrefix={arXiv},
      primaryClass={math.NA},
      url={https://arxiv.org/abs/1403.2141}, 
}

@article{cohen2007stability,
  title={Stability of persistence diagrams},
  author={Cohen-Steiner, David and Edelsbrunner, Herbert and Harer, John},
  journal={Discrete \& Computational Geometry},
  volume={37},
  number={1},
  pages={103--120},
  year={2007},
  publisher={Springer}
}

@book{chazal2014structure,
  title={The structure and stability of persistence modules},
  author={Chazal, Fr{\'e}d{\'e}ric and De Silva, Vin and Glisse, Marc and Oudot, Steve},
  year={2016},
  publisher={Springer}
}

@article{torgerson1952mds,
  title={Multidimensional scaling: {I}. {T}heory and method},
  author={Torgerson, Warren S},
  journal={Psychometrika},
  volume={17},
  number={4},
  pages={401--419},
  year={1952},
  publisher={Springer}
}

@article{batson2021topological,
   title={Topological obstructions to autoencoding},
   volume={2021},
   ISSN={1029-8479},
   url={http://dx.doi.org/10.1007/JHEP04(2021)280},
   DOI={10.1007/jhep04(2021)280},
   number={4},
   journal={Journal of High Energy Physics},
   publisher={Springer Science and Business Media LLC},
   author={Batson, Joshua and Haaf, C. Grace and Kahn, Yonatan and Roberts, Daniel A.},
   year={2021}}

@inproceedings{kingma2022VAE,
  title     = {Auto-Encoding Variational {B}ayes},
  author    = {Kingma, Diederik P. and Welling, Max},
  booktitle = {International Conference on Learning Representations (ICLR)},
  year      = {2014},
  url       = {https://arxiv.org/abs/1312.6114}
}

@InProceedings{chen2022localdistae,
  title = 	 {Local Distance Preserving Auto-encoders using Continuous kNN Graphs},
  author =       {Chen, Nutan and van der Smagt, Patrick and Cseke, Botond},
  booktitle = 	 {Proceedings of Topological, Algebraic, and Geometric Learning Workshops 2022},
  pages = 	 {55--66},
  year = 	 {2022},
  editor = 	 {Cloninger, Alexander and Doster, Timothy and Emerson, Tegan and Kaul, Manohar and Ktena, Ira and Kvinge, Henry and Miolane, Nina and Rieck, Bastian and Tymochko, Sarah and Wolf, Guy},
  volume = 	 {196},
  series = 	 {Proceedings of Machine Learning Research},
  month = 	 {25 Feb--22 Jul},
  publisher =    {PMLR},
  url = 	 {https://proceedings.mlr.press/v196/chen22b.html}
}

@article{mcinnes2020umap, doi = {10.21105/joss.00861}, url = {https://doi.org/10.21105/joss.00861}, year = {2018}, publisher = {The Open Journal}, volume = {3}, number = {29}, pages = {861}, author = {Leland McInnes and John Healy and Nathaniel Saul and Lukas Großberger}, title = {UMAP: Uniform Manifold Approximation and Projection}, journal = {Journal of Open Source Software} }

@article{vandermaattsne,
  author  = {Laurens van der Maaten and Geoffrey Hinton},
  title   = {Visualizing Data using t-SNE},
  journal = {Journal of Machine Learning Research},
  year    = {2008},
  volume  = {9},
  number  = {86},
  pages   = {2579--2605},
  url     = {http://jmlr.org/papers/v9/vandermaaten08a.html}
}

@ARTICLE{GRAE,
  author={Duque, Andres F. and Morin, Sacha and Wolf, Guy and Moon, Kevin R.},
  journal={IEEE Transactions on Pattern Analysis and Machine Intelligence}, 
  title={Geometry Regularized Autoencoders}, 
  year={2023},
  volume={45},
  number={6},
  pages={7381-7394},
  doi={10.1109/TPAMI.2022.3222104}}

@InProceedings{moor2020challenging,
    title       = {Challenging Euclidean Topological Autoencoders},
    author      = {Moor, Michael and Horn, Max and Borgwardt, Karsten and Rieck, Bastian},
    booktitle   = {NeurIPS 2020 Workshop on Topological Data Analysis and Beyond},
    year        = {2020},
    url         = {https://openreview.net/forum?id=P3dZuOUnyEY},
}

@misc{xiao2017fashionmnistnovelimagedataset,
      title={Fashion-MNIST: a Novel Image Dataset for Benchmarking Machine Learning Algorithms}, 
      author={Han Xiao and Kashif Rasul and Roland Vollgraf},
      year={2017},
      eprint={1708.07747},
      archivePrefix={arXiv},
      primaryClass={cs.LG},
      url={https://arxiv.org/abs/1708.07747}, 
}

@ARTICLE{lecun1998mnist,
  author={Lecun, Y. and Bottou, L. and Bengio, Y. and Haffner, P.},
  journal={Proceedings of the IEEE}, 
  title={Gradient-based learning applied to document recognition}, 
  year={1998},
  volume={86},
  number={11},
  pages={2278-2324},
  doi={10.1109/5.726791}}

@techreport{krizhevsky2009learning,
  title={Learning multiple layers of features from tiny images},
  author={Krizhevsky, Alex},
  year={2009},
  institution={University of Toronto}
}

@article{tenenbaum2000global,
  title={A Global Geometric Framework for Nonlinear Dimensionality Reduction},
  author={Tenenbaum, Joshua B and De Silva, Vin and Langford, John C},
  journal={Science},
  volume={290},
  number={5500},
  pages={2319--2323},
  year={2000},
  publisher={American Association for the Advancement of Science}
}

@article{roweis2000nonlinear,
  title={Nonlinear Dimensionality Reduction by Locally Linear Embedding},
  author={Roweis, Sam T and Saul, Lawrence K},
  journal={Science},
  volume={290},
  number={5500},
  pages={2323--2326},
  year={2000},
  publisher={American Association for the Advancement of Science}
}

@book{jolliffe2002principal,
  title={Principal Component Analysis},
  author={Jolliffe, I.T.},
  isbn={9780387954424},
  lccn={2002019560},
  series={Springer Series in Statistics},
  url={https://books.google.ca/books?id=_olByCrhjwIC},
  year={2002},
  publisher={Springer}
}

@article{
hinton2006reducing,
author = {G. E. Hinton  and R. R. Salakhutdinov },
title = {Reducing the Dimensionality of Data with Neural Networks},
journal = {Science},
volume = {313},
number = {5786},
pages = {504-507},
year = {2006},
doi = {10.1126/science.1127647},
URL = {https://www.science.org/doi/abs/10.1126/science.1127647},
eprint = {https://www.science.org/doi/pdf/10.1126/science.1127647}}

@inproceedings{venna2001neighborhood,
  title={Neighborhood preservation in nonlinear projection methods: An experimental study},
  author={Venna, Jarkko and Kaski, Samuel},
  booktitle={International conference on artificial neural networks},
  pages={485--491},
  year={2001},
  organization={Springer}
}

@book{lee2007nonlinear,
  title={Nonlinear dimensionality reduction},
  author={Lee, John A and Verleysen, Michel and others},
  volume={1},
  year={2007},
  publisher={Springer}
}

@misc{kingma2017adam,
      title={Adam: A Method for Stochastic Optimization}, 
      author={Diederik P. Kingma and Jimmy Ba},
      year={2017},
      eprint={1412.6980},
      archivePrefix={arXiv},
      primaryClass={cs.LG},
      url={https://arxiv.org/abs/1412.6980}, 
}

@article{vietoris1927hoheren,
  title={{\"U}ber den h{\"o}heren Zusammenhang kompakter R{\"a}ume und eine Klasse von zusammenhangstreuen Abbildungen},
  author={Vietoris, Leopold},
  journal={Mathematische Annalen},
  volume={97},
  number={1},
  pages={454--472},
  year={1927},
  publisher={Springer}
}

@article{Hotelling1933pca,
  title = {Analysis of a complex of statistical variables into principal components},
  author = {Hotelling, Harold},
  journal = {Journal of Educational Psychology},
  volume = {24},
  number = {6},
  pages = {417-441},
  year = {1933},
  publisher = {American Psychological Association}
}

@misc{sainburg2021paraUMAP,
      title={Parametric UMAP embeddings for representation and semi-supervised learning}, 
      author={Tim Sainburg and Leland McInnes and Timothy Q Gentner},
      year={2021},
      eprint={2009.12981},
      archivePrefix={arXiv},
      primaryClass={cs.LG},
      url={https://arxiv.org/abs/2009.12981}, 
}

@InProceedings{maaten2009paramtSNE,
  title = 	 {Learning a Parametric Embedding by Preserving Local Structure},
  author = 	 {van der Maaten, Laurens},
  booktitle = 	 {Proceedings of the Twelfth International Conference on Artificial Intelligence and Statistics},
  pages = 	 {384--391},
  year = 	 {2009},
  editor = 	 {van Dyk, David and Welling, Max},
  volume = 	 {5},
  series = 	 {Proceedings of Machine Learning Research},
  address = 	 {Hilton Clearwater Beach Resort, Clearwater Beach, Florida USA},
  month = 	 {16--18 Apr},
  publisher =    {PMLR},
  url = 	 {https://proceedings.mlr.press/v5/maaten09a.html}
}

@inproceedings{liu2015celeba,
  title = {Deep Learning Face Attributes in the Wild},
  author = {Liu, Ziwei and Luo, Ping and Wang, Xiaogang and Tang, Xiaoou},
  booktitle = {Proceedings of International Conference on Computer Vision (ICCV)},
  month = {December},
  year = {2015} 
}

@misc{dsprites17,
author = {Loic Matthey and Irina Higgins and Demis Hassabis and Alexander Lerchner},
title = {dSprites: Disentanglement testing Sprites dataset},
howpublished= {https://github.com/deepmind/dsprites-dataset/},
year = "2017",
}

@misc{3dshapes18,
  title={3D Shapes Dataset},
  author={Burgess, Chris and Kim, Hyunjik},
  howpublished={https://github.com/deepmind/3dshapes-dataset/},
  year={2018}
}

@inproceedings{xu2024Assessing,
author = {Xu, Jingyi and Le, Hieu and Samaras, Dimitris},
title = {Assessing Sample Quality via the Latent Space of Generative Models},
year = {2024},
isbn = {978-3-031-73201-0},
publisher = {Springer-Verlag},
address = {Berlin, Heidelberg},
url = {https://doi.org/10.1007/978-3-031-73202-7_26},
doi = {10.1007/978-3-031-73202-7_26},
booktitle = {Computer Vision – ECCV 2024: 18th European Conference, Milan, Italy, September 29–October 4, 2024, Proceedings, Part LIX},
pages = {449–464},
numpages = {16},
location = {Milan, Italy}
}

@book{borg2005modern,
  address = {New York},
  author = {Borg, Ingwer and Groenen, Patrick J. F.},
  doi = {10.1007/0-387-28981-X},
  isbn = {038728981X},
  publisher = {Springer},
  title = {Modern Multidimensional Scaling Theory and Applications},
  year = 2005
}

@inproceedings{aggarwal2001surprising,
author = {Aggarwal, Charu C. and Hinneburg, Alexander and Keim, Daniel A.},
title = {On the Surprising Behavior of Distance Metrics in High Dimensional Spaces},
year = {2001},
isbn = {3540414568},
publisher = {Springer-Verlag},
address = {Berlin, Heidelberg},
booktitle = {Proceedings of the 8th International Conference on Database Theory},
pages = {420–434},
numpages = {15},
series = {ICDT '01}
}

@InProceedings{lim2024ggae,
  title = 	 {Graph Geometry-Preserving Autoencoders},
  author =       {Lim, Jungbin and Kim, Jihwan and Lee, Yonghyeon and Jang, Cheongjae and Park, Frank C.},
  booktitle = 	 {Proceedings of the 41st International Conference on Machine Learning},
  pages = 	 {29795--29815},
  year = 	 {2024},
  editor = 	 {Salakhutdinov, Ruslan and Kolter, Zico and Heller, Katherine and Weller, Adrian and Oliver, Nuria and Scarlett, Jonathan and Berkenkamp, Felix},
  volume = 	 {235},
  series = 	 {Proceedings of Machine Learning Research},
  month = 	 {21--27 Jul},
  publisher =    {PMLR},
  url = 	 {https://proceedings.mlr.press/v235/lim24a.html}
}

@InProceedings{nazari2023geomae,
  title = 	 {Geometric Autoencoders - What You See is What You Decode},
  author =       {Nazari, Philipp and Damrich, Sebastian and Hamprecht, Fred A},
  booktitle = 	 {Proceedings of the 40th International Conference on Machine Learning},
  pages = 	 {25834--25857},
  year = 	 {2023},
  editor = 	 {Krause, Andreas and Brunskill, Emma and Cho, Kyunghyun and Engelhardt, Barbara and Sabato, Sivan and Scarlett, Jonathan},
  volume = 	 {202},
  series = 	 {Proceedings of Machine Learning Research},
  month = 	 {23--29 Jul},
  publisher =    {PMLR},
  url = 	 {https://proceedings.mlr.press/v202/nazari23a.html}
}

@article{chari2023genomics,
    doi = {10.1371/journal.pcbi.1011288},
    author = {Chari, Tara AND Pachter, Lior},
    journal = {PLOS Computational Biology},
    publisher = {Public Library of Science},
    title = {The specious art of single-cell genomics},
    year = {2023},
    month = {08},
    volume = {19},
    url = {https://doi.org/10.1371/journal.pcbi.1011288},
    pages = {1-20},
    number = {8},

}

@misc{mnoichl2025mammoth,
  author       = {MNoichl},
  title        = {MNoichl/UMAP-examples-mammoth: 0.0.1},
  month        = oct,
  year         = 2025,
  publisher    = {Zenodo},
  version      = {0.0.1},
  doi          = {10.5281/zenodo.17290165},
  url          = {https://doi.org/10.5281/zenodo.17290165},
  swhid        = {swh:1:dir:4a671d691b3eb804363af8678823ca6a42de45de
                   ;origin=https://doi.org/10.5281/zenodo.17290164;vi
                   sit=swh:1:snp:5d9397181f896a85bf5ce885903ca24b088e
                   4e0b;anchor=swh:1:rel:e2556c0773fa6c20af20d1c515d9
                   3fb2dab90ca0;path=MNoichl-UMAP-examples-mammoth-
                   ff828d9
                  },
}

@article{Paul2015,
  author = {Paul, Franziska and Arkin, Ya'ara and Giladi, Amir and Jaitin, Diego Adhemar and Kenigsberg, Ephraim and Keren-Shaul, Hadas and Winter, Deborah and Lara-Astiaso, David and Gury, Meital and Weiner, Assaf and David, Eyal and Cohen, Nadav and Lauridsen, Felicia Kathrine Bratt and Haas, Simon and Schlitzer, Andreas and Mildner, Alexander and Ginhoux, Florent and Jung, Steffen and Trumpp, Andreas and Porse, Bo Torben and Tanay, Amos and Amit, Ido},
  title = {Transcriptional Heterogeneity and Lineage Commitment in Myeloid Progenitors},
  journal = {Cell},
  year = {2015},
  volume = {163},
  number = {7},
  pages = {1663--1677},
  month = {Dec},
  doi = {10.1016/j.cell.2015.11.013},
  note = {PMID: 26627738}
}

@misc{10xGenomics2016,
  author = {{10x Genomics}},
  title = {3k {PBMCs} from a {H}ealthy {D}onor},
  year = {2016},
  howpublished = {Single Cell Gene Expression Dataset, Version 1.1.0},
  url = {https://www.10xgenomics.com/datasets/3-k-pbm-cs-from-a-healthy-donor-1-standard-1-1-0},
  note = {Accessed via Scanpy datasets}
}

@article{Wolf2018scanpy,
  author = {Wolf, F. Alexander and Angerer, Philipp and Theis, Fabian J.},
  title = {SCANPY: large-scale single-cell gene expression data analysis},
  journal = {Genome Biology},
  year = {2018},
  volume = {19},
  number = {1},
  pages = {15},
  doi = {10.1186/s13059-017-1382-0},
  url = {https://doi.org/10.1186/s13059-017-1382-0},
  note = {PMID: 29409532}
}

@inproceedings{DeLeeuw1977smacof,
  author = {De Leeuw, Jan},
  title = {Applications of convex analysis to multidimensional scaling},
  booktitle = {Recent Developments in Statistics},
  editor = {Barra, J. R. and Brodeau, F. and Romier, G. and van Cutsem, B.},
  publisher = {North-Holland},
  address = {Amsterdam},
  year = {1977},
  pages = {133--145}
}

@INPROCEEDINGS{singh2021spae,
  author={Singh, Ayushman and Nag, Kaustuv},
  booktitle={2021 IEEE/ACIS 22nd International Conference on Software Engineering, Artificial Intelligence, Networking and Parallel/Distributed Computing (SNPD)}, 
  title={Structure-Preserving Deep Autoencoder-based Dimensionality Reduction for Data Visualization}, 
  year={2021},
  volume={},
  number={},
  pages={43-48},
  doi={10.1109/SNPD51163.2021.9705000}}

@book{Goodfellow-et-al-2016,
    title={Deep Learning},
    author={Ian Goodfellow and Yoshua Bengio and Aaron Courville},
    publisher={MIT Press},
    note={\url{http://www.deeplearningbook.org}},
    year={2016}
}

@article{espadoto2021survey,
  author={Espadoto, Mateus and Martins, Rafael M. and Kerren, Andreas and Hirata, Nina S. T. and Telea, Alexandru C.},
  title={Toward a Quantitative Survey of Dimension Reduction Techniques},
  journal={IEEE Transactions on Visualization and Computer Graphics},
  volume={27}, number={3}, pages={2153--2173}, year={2021},
  doi={10.1109/TVCG.2019.2944182}}

@article{wattenberg2016tsne,
  author={Wattenberg, Martin and Vi{\'e}gas, Fernanda and Johnson, Ian},
  title={How to Use t-SNE Effectively}, journal={Distill}, year={2016},
  doi={10.23915/distill.00002}, url={https://distill.pub/2016/misread-tsne/}}

@article{kobak2021initialization,
  author={Kobak, Dmitry and Linderman, George C.},
  title={Initialization is critical for preserving global data structure in both t-SNE and UMAP},
  journal={Nature Biotechnology}, volume={39}, number={2}, pages={156--157}, year={2021},
  doi={10.1038/s41587-020-00809-z}}

@misc{amid2019trimap,
  author={Amid, Ehsan and Warmuth, Manfred K.},
  title={{TriMap}: Large-scale Dimensionality Reduction Using Triplets},
  year={2019}, eprint={1910.00204}, archivePrefix={arXiv}, primaryClass={cs.LG}}

@misc{gropp2020isometric,
  title={Isometric Autoencoders},
  author={Gropp, Amos and Atzmon, Matan and Lipman, Yaron},
  year={2020}, eprint={2006.09289}, archivePrefix={arXiv}, primaryClass={cs.LG}}

@inproceedings{khosla2020supcon,
  title={Supervised Contrastive Learning},
  author={Khosla, Prannay and Teterwak, Piotr and Wang, Chen and Sarna, Aaron and Tian, Yonglong and Isola, Phillip and Maschinot, Aaron and Liu, Ce and Krishnan, Dilip},
  booktitle={Advances in Neural Information Processing Systems (NeurIPS)}, year={2020}}

@article{palma2025enforcing,
  title={Enforcing latent euclidean geometry in single-cell VAEs for manifold interpolation},
  author={Palma, Alessandro and Rybakov, Sergei and Hetzel, Leon and G{\"u}nnemann, Stephan and Theis, Fabian J},
  journal={arXiv preprint arXiv:2507.11789},
  year={2025}
}

@inproceedings{akiba2019optuna,
  title={Optuna: A next-generation hyperparameter optimization framework},
  author={Akiba, Takuya and Sano, Shotaro and Yanase, Toshihiko and Ohta, Takeru and Koyama, Masanori},
  booktitle={Proceedings of the 25th ACM SIGKDD international conference on knowledge discovery \& data mining},
  pages={2623--2631},
  year={2019}
}

@inproceedings{lecun-cvpr-2004,
    author = {LeCun, Yann and Huang, Fu Jie and Bottou, Leon},
    title = {Learning methods for generic object recognition with invariance to pose and lighting},
    booktitle = {Proceedings of the 2004 IEEE Computer Society Conference on Computer Vision and Pattern Recognition (CVPR)},
    year = {2004},
    volume = {2},
    pages = {II-104--II-111},
    doi = {10.1109/CVPR.2004.1315150}
}

@article{gondal2019transfer,
  title={On the transfer of inductive bias from simulation to the real world: a new disentanglement dataset},
  author={Gondal, Muhammad Waleed and Wuthrich, Manuel and Miladinovic, Djordje and Locatello, Francesco and Breidt, Martin and Volchkov, Valentin and Akpo, Joel and Bachem, Olivier and Sch{\"o}lkopf, Bernhard and Bauer, Stefan},
  journal={Advances in Neural Information Processing Systems},
  volume={32},
  year={2019}
}

@misc{goodfellow2014generativeadversarialnetworks,
      title={Generative Adversarial Networks}, 
      author={Ian J. Goodfellow and Jean Pouget-Abadie and Mehdi Mirza and Bing Xu and David Warde-Farley and Sherjil Ozair and Aaron Courville and Yoshua Bengio},
      year={2014},
      eprint={1406.2661},
      archivePrefix={arXiv},
      primaryClass={stat.ML},
      url={https://arxiv.org/abs/1406.2661}, 
}

@misc{rombach2022highresolutionimagesynthesislatent,
      title={High-Resolution Image Synthesis with Latent Diffusion Models}, 
      author={Robin Rombach and Andreas Blattmann and Dominik Lorenz and Patrick Esser and Björn Ommer},
      year={2022},
      eprint={2112.10752},
      archivePrefix={arXiv},
      primaryClass={cs.CV},
      url={https://arxiv.org/abs/2112.10752}, 
}

@inproceedings{xu2023geometric,
  title={Geometric latent diffusion models for 3d molecule generation},
  author={Xu, Minkai and Powers, Alexander S and Dror, Ron O and Ermon, Stefano and Leskovec, Jure},
  booktitle={International Conference on Machine Learning},
  pages={38592--38610},
  year={2023},
  organization={PMLR}
}

@inproceedings{chen2025dc,
  title={Dc-ae 1.5: Accelerating diffusion model convergence with structured latent space},
  author={Chen, Junyu and Zou, Dongyun and He, Wenkun and Chen, Junsong and Xie, Enze and Han, Song and Cai, Han},
  booktitle={2025 IEEE/CVF International Conference on Computer Vision (ICCV)},
  pages={1--10},
  year={2025},
  organization={IEEE}
}

@book{edelsbrunner2010computational,
  title     = {Computational Topology: An Introduction},
  author    = {Edelsbrunner, Herbert and Harer, John},
  publisher = {American Mathematical Society},
  address   = {Providence, RI},
  year      = {2010}
}

@article{chazal2009gromov,
  title   = {{G}romov--{H}ausdorff Stable Signatures for Shapes using Persistence},
  author  = {Chazal, Fr{\'e}d{\'e}ric and Cohen-Steiner, David and Guibas, Leonidas J. and M{\'e}moli, Facundo and Oudot, Steve Y.},
  journal = {Computer Graphics Forum},
  volume  = {28},
  number  = {5},
  pages   = {1393--1403},
  year    = {2009}
}

@article{cohen2010lipschitz,
  title   = {Lipschitz Functions Have ${L}_p$-Stable Persistence},
  author  = {Cohen-Steiner, David and Edelsbrunner, Herbert and Harer, John and Mileyko, Yuriy},
  journal = {Foundations of Computational Mathematics},
  volume  = {10},
  number  = {2},
  pages   = {127--139},
  year    = {2010}
}

@article{szekely2007measuring,
  title   = {Measuring and Testing Dependence by Correlation of Distances},
  author  = {Sz{\'e}kely, G{\'a}bor J. and Rizzo, Maria L. and Bakirov, Nail K.},
  journal = {The Annals of Statistics},
  volume  = {35},
  number  = {6},
  pages   = {2769--2794},
  year    = {2007}
}
\bibliographystyle{tmlr}

\appendix

\section{Persistent homology}
\label{app:persistent-homology}
This appendix gives the formal construction underlying the persistence diagrams and the stability
theorem used in \S\ref{sec:background} and \S\ref{sec:theory}, and on which the topological
competitors (TopoAE, RTD-AE) and our $W_0,W_1$ metrics rely. Throughout, homology is taken with
coefficients in a field $\mathbb{F}$ (e.g.\ $\mathbb{F}=\mathbb{Z}/2$), so each homology group is a
vector space and its dimension (the Betti number) is well defined. See
\citet{carlsson2009topology,edelsbrunner2010computational} for full treatments.

\paragraph{Vietoris--Rips complex.}
Let $X=\{x_1,\dots,x_N\}\subset\R^n$ and fix a \emph{scale} $\varepsilon>0$. The
\emph{Vietoris--Rips complex} \citep{vietoris1927hoheren} $\mathrm{VR}_\varepsilon(X)$ is the
simplicial complex whose $p$-simplices are the subsets $\sigma=\{x_{i_0},\dots,x_{i_p}\}\subseteq X$
of diameter at most $\varepsilon$, i.e.\ with $\lVert x_{i_a}-x_{i_b}\rVert\le\varepsilon$ for every
pair. Concretely: each point is a $0$-simplex, each $\varepsilon$-close pair an edge ($1$-simplex),
each triple that is pairwise $\varepsilon$-close a triangle ($2$-simplex), and so on. As in
\S\ref{sec:background}, computing homology in degree $p$ requires the $(p{+}1)$-skeleton.

\paragraph{Chain complex and homology.}
Let $C_p$ be the $\mathbb{F}$-vector space with basis the (oriented) $p$-simplices of
$\mathrm{VR}_\varepsilon(X)$. Writing the $j$-th face of $\sigma=(x_{i_0},\dots,x_{i_p})$ as
$\hat\sigma_j=(x_{i_0},\dots,x_{i_{j-1}},x_{i_{j+1}},\dots,x_{i_p})$, the boundary operator
$\partial_p:C_p\to C_{p-1}$ is
\[
  \partial_p(\sigma)=\sum_{j=0}^{p}(-1)^j\,\hat\sigma_j,
\]
(over $\mathbb{F}=\mathbb{Z}/2$ the signs may be dropped). The \emph{cycles} are
$Z_p=\ker\partial_p$ and the \emph{boundaries} are $B_p=\operatorname{im}\partial_{p+1}$; since
$\partial_p\partial_{p+1}=0$ we have $B_p\subseteq Z_p$, and the $p$-th homology group is the quotient
\[
  H_p\big(\mathrm{VR}_\varepsilon(X)\big)=Z_p/B_p=\ker\partial_p/\operatorname{im}\partial_{p+1}.
\]
Its dimension $\beta_p=\dim H_p$ counts the $p$-dimensional ``holes'': $\beta_0$ the connected
components, $\beta_1$ the loops, $\beta_2$ the enclosed voids.

\paragraph{Filtration and persistence.}
For $\varepsilon\le\varepsilon'$ the inclusion
$\mathrm{VR}_\varepsilon(X)\subseteq\mathrm{VR}_{\varepsilon'}(X)$ induces a linear map
$f_p^{\varepsilon,\varepsilon'}:H_p(\mathrm{VR}_\varepsilon(X))\to H_p(\mathrm{VR}_{\varepsilon'}(X))$.
Tracking these maps as $\varepsilon$ grows, a homology class is \emph{born} at the smallest scale $b$
at which it appears (it is not in the image of $f_p^{\varepsilon,b}$ from any smaller $\varepsilon<b$)
and \emph{dies} at the smallest scale $d\ge b$ at which it becomes trivial (it enters the kernel of
$f_p^{b,d}$, i.e.\ merges into an older class or is filled in). The \emph{persistence diagram}
$\mathrm{dgm}_p(X)$ records each class as a point $(b,d)$ in the plane; classes that never die are
assigned $d=\infty$. These are precisely the diagrams whose bottleneck and $p$-Wasserstein distances
we use as $W_0,W_1$ (\S\ref{app:metrics}), and whose stability under perturbation of $X$ underlies the
theorem of \S\ref{sec:theory}.

\paragraph{Worked example: points on a circle.}
Let $X$ be the four vertices of a square inscribed in the unit circle, with side $s = \sqrt{2}$ and diagonal 2.
\begin{itemize}
    \item For $\varepsilon < \sqrt{2}$ the points are isolated: $\beta_0 = 4$ and $\beta_1 = 0$.
    \item At $\varepsilon = \sqrt{2}$ the four sides appear together: the components merge $(\beta_0=1)$ and one loop is born $(\beta_1 = 1)$.
    \item At $\varepsilon = 2$ both diagonals appear, all four points are pairwise $\varepsilon$-close, and the 3-simplex they span fills the loop $(\beta_1 = 0)$.
\end{itemize}
Hence $\operatorname{dgm}_0(X)$ has the point $(0, \sqrt{2})$ with multiplicity 3 plus the essential class $(0, \infty)$, and  $\operatorname{dgm}_1(X)$ is the single point $(\sqrt{2}, 2)$. For m evenly spaced points the death scale, and even the homotopy type of the complex, depends on $m$ \cite{adamaszek2027pacific}

\section{Datasets}
\label{app:datasets}
Table~\ref{tab:datasets} lists the fifteen datasets, their ambient dimension $n$ (the flattened input
seen by every method), the number of points used, and the source. The five geometric sets are
synthetic manifolds with a known intended structure; the eight image sets are flattened to $n$ and
capped at $40{,}000$ points (a frozen subsample, shared across methods); the two single-cell sets are
used in full. Nested spheres places ten small spheres inside a larger enclosing shell; linked tori are
two interlocked tori; the swiss roll is a rolled 2D sheet; mammoth is a 3D scan; earth samples a
sphere with procedural continent labels (the illustration in Figure~\ref{fig:manifold} instead uses a
real globe whose land points are labelled by continent); the geometric sets other than the swiss roll and the
nested spheres (native dimension $101$) are lifted to $100$ ambient dimensions by a random orthogonal
embedding (distances preserved).

\begin{table}[t]
\centering\small
\caption{Datasets: ambient dimension $n$, number of points, and source. Geometric sets are synthetic manifolds or low-dimensional point clouds;
image sets are flattened and subsampled to $40{,}000$ points; single-cell sets are used in full.}
\label{tab:datasets}
\begin{tabular}{@{}llrrl@{}}
\toprule
Dataset & Structure & $n$ & points & Source \\
\midrule
\multicolumn{5}{@{}l}{\emph{Geometric (synthetic)}}\\
Nested spheres & 10 spheres in a shell   & 101    & 10{,}000 & \citep{moor2020topological} \\
Linked tori    & two interlocked tori    & 100    & 4{,}000  & this work \\
Swiss roll     & rolled 2D sheet         & 3      & 4{,}000  & \citep{tenenbaum2000global} \\
Mammoth        & 3D scan (lifted)        & 100    & 6{,}000  & \citep{mnoichl2025mammoth} \\
Earth          & 3D sphere with continents (lifted)  & 100    & 6{,}000  & \citep{nazari2023geomae} \\
\midrule
\multicolumn{5}{@{}l}{\emph{Image}}\\
MNIST          & handwritten digits      & 784    & 40{,}000 & \citep{lecun1998mnist} \\
Fashion-MNIST  & clothing                & 784    & 40{,}000 & \citep{xiao2017fashionmnistnovelimagedataset} \\
smallNORB      & toys ($32{\times}32$)   & 1{,}024  & 40{,}000 & \citep{lecun-cvpr-2004} \\
CIFAR-10       & natural images          & 3{,}072  & 40{,}000 & \citep{krizhevsky2009learning} \\
dSprites       & 2D shapes (factors)     & 4{,}096  & 40{,}000 & \citep{dsprites17} \\
3D-Shapes      & rendered scenes         & 12{,}288 & 40{,}000 & \citep{3dshapes18} \\
MPI3D          & robotic-arm images      & 12{,}288 & 40{,}000 & \citep{gondal2019transfer} \\
CelebA         & faces ($64{\times}64$)  & 12{,}288 & 40{,}000 & \citep{liu2015celeba} \\
\midrule
\multicolumn{5}{@{}l}{\emph{Single-cell}}\\
PBMC3k         & PBMC (highly-var.\ genes) & 1{,}838 & 2{,}638 & \citep{10xGenomics2016,Wolf2018scanpy} \\
Paul15         & myeloid progenitors      & 1{,}000 & 2{,}730 & \citep{Paul2015} \\
\bottomrule
\end{tabular}
\end{table}

\section{Evaluation metrics}
\label{app:metrics}
All metrics compare the input $X=\{x_i\}$ with the latent $Z=\{z_i\}$ and are computed on five random
$500$-point subsamples, of which we report the mean. Let $D_X,D_Z$ be the Euclidean pairwise-distance
matrices.
\begin{itemize}[leftmargin=1.4em,itemsep=2pt,topsep=2pt]
\item \textbf{Distance correlation (DC)},  Pearson correlation of the flattened upper-triangular
distances, $\mathrm{DC}=\mathrm{corr}\big(\{D_X^{ij}\}_{i<j},\{D_Z^{ij}\}_{i<j}\big)$; global metric
fidelity, higher is better \citep{moor2020topological}.
\item \textbf{Triplet accuracy (TA)},  over $5{,}000$ random triplets $(a,p,q)$, the fraction with
$\operatorname{sign}(D_X^{ap}-D_X^{aq})=\operatorname{sign}(D_Z^{ap}-D_Z^{aq})$; ordinal/global fidelity.
\item \textbf{Trustworthiness $\times$ continuity (T$\times$C)},  trustworthiness $T(k)$ penalises
points that intrude into the latent $k$-neighbourhood but were not input neighbours, and continuity
$C(k)$ penalises the reverse \citep{venna2001neighborhood}:
$T(k)=1-\tfrac{2}{Nk(2N-3k-1)}\sum_i\sum_{j\in U_k(i)}\big(r(i,j)-k\big)$, with $U_k(i)$ the latent
$k$-NN of $i$ that are not input $k$-NN and $r(i,j)$ the input-distance rank of $j$; $C(k)$ swaps input
and latent. We report $\tfrac{1}{|K|}\sum_{k\in K}T(k)\,C(k)$ over a valid range $K$ of neighbourhood
sizes (local fidelity, higher is better).
\item \textbf{Persistence-Wasserstein ($W_0,W_1$)},  the $p{=}1$ Wasserstein distance between the
Vietoris--Rips persistence diagrams of $X$ and $Z$ in homology degrees $0$ and $1$; topological
fidelity, lower is better \citep{cohen2010lipschitz}. Before computing the diagrams, each point cloud is
rescaled so that its 90th-percentile pairwise distance equals one, so $W_0,W_1$ compare shape rather
than overall scale.
\item \textbf{Density agreement (KL$_\sigma$).} Following \citet{moor2020topological}, distances are first
rescaled to $[0,1]$ (giving $\tilde D_X,\tilde D_Z$) and a kernel density is formed at bandwidth
$\sigma$,
\[
  \rho_X(i)=\frac{\sum_{j}\exp\!\big(-(\tilde D_X^{ij})^2/\sigma\big)}
                 {\sum_{k}\sum_{j}\exp\!\big(-(\tilde D_X^{kj})^2/\sigma\big)},
\]
and likewise $\rho_Z$; their divergence is
$\mathrm{KL}_\sigma=\sum_i \rho_X(i)\,\log\!\big(\rho_X(i)/\rho_Z(i)\big)$. Lower is better. We select
hyperparameters on $\mathrm{KL}_{0.1}$; the reported density-agreement score averages $\mathrm{KL}_\sigma$
over a range of bandwidths $\sigma$.
\end{itemize}

\section{Training and hyperparameter selection}
\label{app:proto}
\paragraph{Architecture and optimisation.} Every method uses the same MLP encoder/decoder with hidden
widths $(256,128)$, batch normalisation and ReLU, and a two-dimensional bottleneck (the ``fair
frame''), trained with Adam \citep{kingma2017adam}. Data are split $80/10/10$ into train/validation/test
with a frozen seed shared across all methods.

\paragraph{Hyperparameter search.} For each (dataset, method) pair we run Optuna \citep{akiba2019optuna}
with a TPE sampler for $25$ trials at the 2D bottleneck, minimising the validation density-agreement
KL; the search covers the learning rate ($10^{-4}$--$3\times10^{-3}$), batch size, and each method's
regulariser weight. Batch-size grids are method-specific: $\{8,\dots,256\}$ for most methods,
$\{32,\dots,256\}$ for TopoAE and $\{32,64,80,100\}$ for RTD-AE, whose per-batch persistence cost grows
quickly with $b$. To keep the $25$ trials tractable, each trial trains for $100$ epochs (TopoAE: $100$ or $200$, searched) on a
fixed random subsample of $800$ training points and is scored on the validation split. The best configuration is then
retrained with the same number of epochs on the full training set (capped at $40{,}000$ points) with five test seeds and evaluated on the
blind test split; reported numbers are means over the five seeds.

\section{Full per-dataset metrics}
\label{app:full-metrics}
Table~\ref{tab:rank} summarizes performance as a mean rank across datasets. Here we give the
underlying per-dataset values for all six metrics: distance correlation (DC) and triplet
accuracy (TA) for global/ordinal structure, trustworthiness~$\times$~continuity (T$\times$C)
for local neighbourhoods, the persistence-Wasserstein distances (W$_0$, W$_1$) for topology,
and the density agreement KL$_{0.1}$. Every entry is the mean over $5$ seeds under each method's
tuned configuration; ``--'' marks GGAE where its $O(n^2)$ kernel exceeds memory.

% DC @ bottleneck=2  (higher better: True)
\begin{sidewaystable}[p]
\centering
\footnotesize
\caption{\textbf{Distance correlation (DC, $\uparrow$)} per dataset at the 2D bottleneck; mean$\pm$std over 5 seeds, \textbf{best} per column in bold; ``--'' marks GGAE where its $O(n^2)$ kernel exceeds memory. Datasets are split across two blocks for legibility.}
\label{tab:app-DC}
\begin{tabular}{lrrrrrrrr}
\toprule
method & nested\_spheres & linked\_tori & swiss\_roll & mammoth & earth & mnist & fashion\_mnist & cifar10 \\
\midrule
Vanilla AE & -0.033$\pm$0.021 & 0.809$\pm$0.017 & 0.528$\pm$0.056 & 0.915$\pm$0.013 & 0.561$\pm$0.018 & 0.464$\pm$0.044 & 0.646$\pm$0.062 & 0.731$\pm$0.024 \\
MMAE-raw & \textbf{0.467$\pm$0.010} & 0.920$\pm$0.009 & 0.722$\pm$0.044 & 0.965$\pm$0.002 & \textbf{0.771$\pm$0.003} & 0.457$\pm$0.042 & 0.837$\pm$0.019 & \textbf{0.872$\pm$0.007} \\
MMAE-raw-z & 0.411$\pm$0.018 & 0.886$\pm$0.009 & 0.767$\pm$0.020 & 0.980$\pm$0.002 & 0.750$\pm$0.006 & 0.381$\pm$0.015 & 0.674$\pm$0.059 & 0.803$\pm$0.010 \\
MMAE-pca & 0.129$\pm$0.019 & \textbf{0.922$\pm$0.009} & \textbf{0.810$\pm$0.001} & 0.971$\pm$0.003 & 0.759$\pm$0.003 & 0.692$\pm$0.010 & \textbf{0.888$\pm$0.004} & 0.870$\pm$0.007 \\
MMAE-pca-z & 0.243$\pm$0.029 & 0.883$\pm$0.006 & 0.810$\pm$0.001 & 0.981$\pm$0.002 & 0.760$\pm$0.003 & \textbf{0.700$\pm$0.009} & 0.869$\pm$0.010 & 0.830$\pm$0.009 \\
MMAE-Isomap & 0.388$\pm$0.006 & 0.807$\pm$0.014 & 0.372$\pm$0.022 & 0.907$\pm$0.010 & 0.766$\pm$0.002 & 0.587$\pm$0.021 & 0.801$\pm$0.003 & 0.853$\pm$0.006 \\
MMAE-Isomap-z & 0.362$\pm$0.014 & 0.844$\pm$0.003 & 0.360$\pm$0.022 & 0.916$\pm$0.005 & 0.761$\pm$0.005 & 0.595$\pm$0.028 & 0.786$\pm$0.004 & 0.863$\pm$0.005 \\
SPAE & 0.363$\pm$0.022 & 0.892$\pm$0.007 & 0.697$\pm$0.037 & 0.902$\pm$0.008 & 0.715$\pm$0.005 & 0.503$\pm$0.023 & 0.653$\pm$0.037 & 0.720$\pm$0.017 \\
GRAE-pca & 0.070$\pm$0.013 & 0.905$\pm$0.003 & 0.808$\pm$0.002 & \textbf{0.984$\pm$0.001} & 0.762$\pm$0.003 & 0.652$\pm$0.014 & 0.887$\pm$0.004 & 0.823$\pm$0.010 \\
GRAE-Isomap & 0.294$\pm$0.008 & 0.742$\pm$0.021 & 0.327$\pm$0.005 & 0.917$\pm$0.002 & 0.767$\pm$0.002 & 0.594$\pm$0.009 & 0.777$\pm$0.007 & 0.809$\pm$0.008 \\
GRAE-umap & 0.122$\pm$0.014 & 0.329$\pm$0.019 & 0.317$\pm$0.015 & 0.821$\pm$0.002 & 0.736$\pm$0.005 & 0.313$\pm$0.007 & 0.627$\pm$0.005 & 0.759$\pm$0.010 \\
TopoAE & 0.409$\pm$0.020 & 0.875$\pm$0.008 & 0.722$\pm$0.035 & 0.915$\pm$0.013 & 0.678$\pm$0.003 & 0.474$\pm$0.019 & 0.797$\pm$0.016 & 0.829$\pm$0.007 \\
RTD-AE & 0.311$\pm$0.020 & 0.912$\pm$0.007 & 0.801$\pm$0.006 & 0.892$\pm$0.027 & 0.695$\pm$0.014 & 0.526$\pm$0.017 & 0.789$\pm$0.014 & 0.830$\pm$0.006 \\
GeomAE & -0.045$\pm$0.028 & 0.854$\pm$0.018 & 0.568$\pm$0.043 & 0.931$\pm$0.012 & 0.628$\pm$0.029 & 0.425$\pm$0.034 & 0.597$\pm$0.045 & 0.750$\pm$0.022 \\
GGAE & 0.201$\pm$0.009 & 0.853$\pm$0.011 & 0.739$\pm$0.014 & 0.863$\pm$0.007 & 0.755$\pm$0.007 & -- & -- & -- \\
IsoAE & 0.121$\pm$0.037 & 0.765$\pm$0.054 & 0.651$\pm$0.048 & 0.908$\pm$0.016 & 0.671$\pm$0.010 & 0.491$\pm$0.042 & 0.678$\pm$0.068 & 0.723$\pm$0.023 \\
CAE & 0.147$\pm$0.014 & 0.840$\pm$0.060 & 0.654$\pm$0.012 & 0.913$\pm$0.027 & 0.610$\pm$0.023 & 0.291$\pm$0.026 & 0.719$\pm$0.036 & 0.798$\pm$0.007 \\
\bottomrule
\end{tabular}

\par\vspace{1.4em}

\begin{tabular}{lrrrrrrr}
\toprule
method & celeba64 & shapes3d & mpi3d & smallnorb & dsprites & pbmc3k & paul15 \\
\midrule
Vanilla AE & 0.732$\pm$0.019 & 0.392$\pm$0.030 & 0.581$\pm$0.047 & 0.925$\pm$0.022 & 0.555$\pm$0.038 & 0.572$\pm$0.071 & 0.175$\pm$0.082 \\
MMAE-raw & 0.818$\pm$0.022 & 0.621$\pm$0.006 & 0.722$\pm$0.013 & 0.980$\pm$0.003 & 0.660$\pm$0.018 & 0.574$\pm$0.018 & 0.575$\pm$0.015 \\
MMAE-raw-z & 0.858$\pm$0.009 & 0.514$\pm$0.019 & 0.601$\pm$0.011 & 0.983$\pm$0.001 & 0.522$\pm$0.028 & \textbf{0.790$\pm$0.010} & \textbf{0.781$\pm$0.013} \\
MMAE-pca & 0.846$\pm$0.007 & \textbf{0.658$\pm$0.006} & \textbf{0.760$\pm$0.007} & \textbf{0.985$\pm$0.001} & \textbf{0.816$\pm$0.004} & 0.681$\pm$0.013 & 0.572$\pm$0.042 \\
MMAE-pca-z & 0.854$\pm$0.003 & 0.549$\pm$0.017 & 0.677$\pm$0.009 & 0.982$\pm$0.001 & 0.816$\pm$0.005 & 0.589$\pm$0.017 & 0.487$\pm$0.051 \\
MMAE-Isomap & 0.829$\pm$0.012 & 0.606$\pm$0.005 & 0.585$\pm$0.011 & 0.936$\pm$0.005 & 0.671$\pm$0.008 & 0.401$\pm$0.043 & 0.414$\pm$0.028 \\
MMAE-Isomap-z & 0.832$\pm$0.007 & 0.531$\pm$0.091 & 0.629$\pm$0.012 & 0.965$\pm$0.002 & 0.607$\pm$0.023 & 0.571$\pm$0.025 & 0.156$\pm$0.015 \\
SPAE & 0.722$\pm$0.016 & 0.500$\pm$0.030 & 0.621$\pm$0.022 & 0.955$\pm$0.003 & 0.657$\pm$0.022 & 0.519$\pm$0.059 & 0.543$\pm$0.037 \\
GRAE-pca & 0.829$\pm$0.002 & 0.653$\pm$0.004 & 0.678$\pm$0.008 & 0.980$\pm$0.001 & 0.811$\pm$0.004 & 0.611$\pm$0.011 & 0.225$\pm$0.003 \\
GRAE-Isomap & 0.821$\pm$0.005 & 0.637$\pm$0.003 & 0.629$\pm$0.009 & 0.952$\pm$0.005 & 0.643$\pm$0.008 & 0.604$\pm$0.010 & 0.146$\pm$0.004 \\
GRAE-umap & 0.811$\pm$0.004 & 0.444$\pm$0.009 & 0.594$\pm$0.011 & 0.515$\pm$0.024 & 0.560$\pm$0.009 & 0.141$\pm$0.012 & 0.091$\pm$0.006 \\
TopoAE & 0.850$\pm$0.008 & 0.544$\pm$0.014 & 0.675$\pm$0.014 & 0.953$\pm$0.007 & 0.599$\pm$0.030 & 0.655$\pm$0.015 & 0.263$\pm$0.030 \\
RTD-AE & \textbf{0.868$\pm$0.004} & 0.529$\pm$0.018 & 0.693$\pm$0.013 & 0.951$\pm$0.012 & 0.672$\pm$0.009 & 0.478$\pm$0.079 & 0.614$\pm$0.039 \\
GeomAE & 0.774$\pm$0.008 & 0.472$\pm$0.034 & 0.585$\pm$0.046 & 0.762$\pm$0.045 & 0.587$\pm$0.028 & 0.682$\pm$0.067 & 0.184$\pm$0.073 \\
GGAE & -- & -- & -- & -- & -- & 0.380$\pm$0.026 & 0.259$\pm$0.020 \\
IsoAE & 0.650$\pm$0.076 & 0.413$\pm$0.054 & 0.684$\pm$0.021 & 0.956$\pm$0.011 & 0.554$\pm$0.042 & 0.672$\pm$0.033 & 0.337$\pm$0.087 \\
CAE & 0.781$\pm$0.011 & 0.464$\pm$0.034 & 0.601$\pm$0.024 & 0.926$\pm$0.018 & 0.587$\pm$0.011 & 0.718$\pm$0.019 & 0.506$\pm$0.060 \\
\bottomrule
\end{tabular}
\end{sidewaystable}
% TA @ bottleneck=2  (higher better: True)
\begin{sidewaystable}[p]
\centering
\footnotesize
\caption{\textbf{Triplet accuracy (TA, $\uparrow$)} per dataset at the 2D bottleneck; mean$\pm$std over 5 seeds, \textbf{best} per column in bold; ``--'' marks GGAE where its $O(n^2)$ kernel exceeds memory. Datasets are split across two blocks for legibility.}
\label{tab:app-TA}
\begin{tabular}{lrrrrrrrr}
\toprule
method & nested\_spheres & linked\_tori & swiss\_roll & mammoth & earth & mnist & fashion\_mnist & cifar10 \\
\midrule
Vanilla AE & 0.434$\pm$0.007 & 0.793$\pm$0.009 & 0.727$\pm$0.011 & 0.872$\pm$0.010 & 0.779$\pm$0.006 & 0.656$\pm$0.014 & 0.782$\pm$0.028 & 0.757$\pm$0.012 \\
MMAE-raw & \textbf{0.641$\pm$0.008} & 0.875$\pm$0.007 & 0.779$\pm$0.034 & 0.916$\pm$0.004 & 0.803$\pm$0.001 & 0.702$\pm$0.003 & 0.866$\pm$0.004 & \textbf{0.833$\pm$0.007} \\
MMAE-raw-z & 0.609$\pm$0.009 & 0.847$\pm$0.005 & 0.801$\pm$0.004 & 0.929$\pm$0.005 & 0.808$\pm$0.005 & 0.683$\pm$0.006 & 0.851$\pm$0.002 & 0.792$\pm$0.009 \\
MMAE-pca & 0.487$\pm$0.007 & \textbf{0.876$\pm$0.007} & 0.815$\pm$0.003 & 0.922$\pm$0.005 & 0.801$\pm$0.002 & 0.716$\pm$0.007 & \textbf{0.870$\pm$0.004} & 0.832$\pm$0.007 \\
MMAE-pca-z & 0.530$\pm$0.011 & 0.843$\pm$0.004 & \textbf{0.815$\pm$0.003} & 0.930$\pm$0.004 & 0.802$\pm$0.002 & \textbf{0.717$\pm$0.006} & 0.856$\pm$0.003 & 0.801$\pm$0.006 \\
MMAE-Isomap & 0.602$\pm$0.005 & 0.790$\pm$0.008 & 0.663$\pm$0.005 & 0.890$\pm$0.003 & 0.805$\pm$0.003 & 0.706$\pm$0.001 & 0.823$\pm$0.004 & 0.824$\pm$0.006 \\
MMAE-Isomap-z & 0.585$\pm$0.005 & 0.812$\pm$0.002 & 0.654$\pm$0.005 & 0.893$\pm$0.003 & 0.806$\pm$0.003 & 0.709$\pm$0.002 & 0.818$\pm$0.004 & 0.827$\pm$0.004 \\
SPAE & 0.589$\pm$0.010 & 0.852$\pm$0.005 & 0.779$\pm$0.027 & 0.859$\pm$0.008 & \textbf{0.814$\pm$0.002} & 0.694$\pm$0.005 & 0.820$\pm$0.004 & 0.757$\pm$0.011 \\
GRAE-pca & 0.481$\pm$0.004 & 0.863$\pm$0.004 & 0.814$\pm$0.004 & \textbf{0.938$\pm$0.004} & 0.801$\pm$0.001 & 0.694$\pm$0.008 & 0.869$\pm$0.004 & 0.807$\pm$0.007 \\
GRAE-Isomap & 0.573$\pm$0.002 & 0.761$\pm$0.008 & 0.690$\pm$0.004 & 0.894$\pm$0.003 & 0.805$\pm$0.003 & 0.695$\pm$0.006 & 0.812$\pm$0.007 & 0.798$\pm$0.007 \\
GRAE-umap & 0.483$\pm$0.004 & 0.637$\pm$0.005 & 0.659$\pm$0.006 & 0.859$\pm$0.003 & 0.796$\pm$0.003 & 0.600$\pm$0.001 & 0.744$\pm$0.003 & 0.782$\pm$0.008 \\
TopoAE & 0.622$\pm$0.011 & 0.836$\pm$0.002 & 0.794$\pm$0.027 & 0.877$\pm$0.010 & 0.786$\pm$0.008 & 0.666$\pm$0.009 & 0.830$\pm$0.005 & 0.810$\pm$0.006 \\
RTD-AE & 0.564$\pm$0.014 & 0.865$\pm$0.005 & 0.813$\pm$0.001 & 0.864$\pm$0.016 & 0.775$\pm$0.005 & 0.696$\pm$0.003 & 0.823$\pm$0.008 & 0.806$\pm$0.005 \\
GeomAE & 0.433$\pm$0.012 & 0.823$\pm$0.009 & 0.721$\pm$0.022 & 0.883$\pm$0.010 & 0.773$\pm$0.015 & 0.640$\pm$0.007 & 0.728$\pm$0.023 & 0.761$\pm$0.010 \\
GGAE & 0.512$\pm$0.005 & 0.821$\pm$0.008 & 0.785$\pm$0.009 & 0.845$\pm$0.006 & 0.796$\pm$0.002 & -- & -- & -- \\
IsoAE & 0.492$\pm$0.013 & 0.793$\pm$0.018 & 0.761$\pm$0.027 & 0.864$\pm$0.005 & 0.789$\pm$0.009 & 0.641$\pm$0.013 & 0.757$\pm$0.029 & 0.752$\pm$0.011 \\
CAE & 0.502$\pm$0.005 & 0.820$\pm$0.027 & 0.747$\pm$0.006 & 0.878$\pm$0.014 & 0.766$\pm$0.006 & 0.618$\pm$0.011 & 0.803$\pm$0.011 & 0.797$\pm$0.006 \\
\bottomrule
\end{tabular}

\par\vspace{1.4em}

\begin{tabular}{lrrrrrrr}
\toprule
method & celeba64 & shapes3d & mpi3d & smallnorb & dsprites & pbmc3k & paul15 \\
\midrule
Vanilla AE & 0.762$\pm$0.006 & 0.668$\pm$0.008 & 0.730$\pm$0.018 & 0.868$\pm$0.011 & 0.702$\pm$0.019 & 0.635$\pm$0.032 & 0.539$\pm$0.017 \\
MMAE-raw & 0.813$\pm$0.008 & 0.730$\pm$0.002 & 0.798$\pm$0.002 & 0.921$\pm$0.006 & 0.785$\pm$0.007 & 0.645$\pm$0.009 & 0.671$\pm$0.003 \\
MMAE-raw-z & 0.829$\pm$0.007 & 0.714$\pm$0.009 & 0.805$\pm$0.006 & 0.924$\pm$0.004 & 0.768$\pm$0.015 & \textbf{0.701$\pm$0.008} & \textbf{0.736$\pm$0.004} \\
MMAE-pca & 0.823$\pm$0.003 & 0.736$\pm$0.006 & \textbf{0.807$\pm$0.004} & \textbf{0.932$\pm$0.006} & 0.808$\pm$0.006 & 0.669$\pm$0.005 & 0.613$\pm$0.021 \\
MMAE-pca-z & 0.825$\pm$0.005 & 0.705$\pm$0.007 & 0.780$\pm$0.004 & 0.921$\pm$0.003 & \textbf{0.808$\pm$0.005} & 0.587$\pm$0.004 & 0.583$\pm$0.010 \\
MMAE-Isomap & 0.817$\pm$0.007 & 0.722$\pm$0.002 & 0.719$\pm$0.004 & 0.873$\pm$0.005 & 0.752$\pm$0.006 & 0.548$\pm$0.011 & 0.614$\pm$0.004 \\
MMAE-Isomap-z & 0.813$\pm$0.005 & 0.711$\pm$0.005 & 0.748$\pm$0.004 & 0.904$\pm$0.004 & 0.753$\pm$0.005 & 0.549$\pm$0.007 & 0.582$\pm$0.003 \\
SPAE & 0.763$\pm$0.010 & 0.708$\pm$0.009 & 0.747$\pm$0.010 & 0.892$\pm$0.005 & 0.758$\pm$0.008 & 0.612$\pm$0.032 & 0.651$\pm$0.006 \\
GRAE-pca & 0.816$\pm$0.002 & \textbf{0.736$\pm$0.003} & 0.787$\pm$0.004 & 0.919$\pm$0.006 & 0.805$\pm$0.006 & 0.579$\pm$0.004 & 0.525$\pm$0.002 \\
GRAE-Isomap & 0.813$\pm$0.003 & 0.726$\pm$0.003 & 0.746$\pm$0.002 & 0.875$\pm$0.005 & 0.740$\pm$0.006 & 0.586$\pm$0.006 & 0.578$\pm$0.001 \\
GRAE-umap & 0.810$\pm$0.003 & 0.651$\pm$0.005 & 0.701$\pm$0.004 & 0.707$\pm$0.007 & 0.711$\pm$0.004 & 0.558$\pm$0.005 & 0.547$\pm$0.003 \\
TopoAE & 0.828$\pm$0.005 & 0.698$\pm$0.006 & 0.777$\pm$0.005 & 0.904$\pm$0.003 & 0.767$\pm$0.005 & 0.670$\pm$0.004 & 0.597$\pm$0.006 \\
RTD-AE & \textbf{0.837$\pm$0.003} & 0.701$\pm$0.009 & 0.774$\pm$0.006 & 0.906$\pm$0.007 & 0.752$\pm$0.005 & 0.622$\pm$0.012 & 0.656$\pm$0.016 \\
GeomAE & 0.773$\pm$0.003 & 0.696$\pm$0.016 & 0.745$\pm$0.028 & 0.795$\pm$0.015 & 0.720$\pm$0.011 & 0.643$\pm$0.028 & 0.532$\pm$0.014 \\
GGAE & -- & -- & -- & -- & -- & 0.572$\pm$0.004 & 0.564$\pm$0.007 \\
IsoAE & 0.751$\pm$0.012 & 0.650$\pm$0.017 & 0.772$\pm$0.014 & 0.897$\pm$0.007 & 0.695$\pm$0.021 & 0.587$\pm$0.012 & 0.558$\pm$0.013 \\
CAE & 0.794$\pm$0.007 & 0.695$\pm$0.005 & 0.719$\pm$0.014 & 0.854$\pm$0.014 & 0.720$\pm$0.005 & 0.673$\pm$0.007 & 0.636$\pm$0.015 \\
\bottomrule
\end{tabular}
\end{sidewaystable}
% TxC @ bottleneck=2  (higher better: True)
\begin{sidewaystable}[p]
\centering
\footnotesize
\caption{\textbf{Trustworthiness $\times$ continuity (T$\times$C, $\uparrow$)} per dataset at the 2D bottleneck; mean$\pm$std over 5 seeds, \textbf{best} per column in bold; ``--'' marks GGAE where its $O(n^2)$ kernel exceeds memory. Datasets are split across two blocks for legibility.}
\label{tab:app-TxC}
\begin{tabular}{lrrrrrrrr}
\toprule
method & nested\_spheres & linked\_tori & swiss\_roll & mammoth & earth & mnist & fashion\_mnist & cifar10 \\
\midrule
Vanilla AE & 0.394$\pm$0.014 & 0.910$\pm$0.006 & 0.886$\pm$0.033 & 0.948$\pm$0.004 & \textbf{0.955$\pm$0.005} & 0.689$\pm$0.019 & 0.888$\pm$0.018 & 0.744$\pm$0.011 \\
MMAE-raw & 0.542$\pm$0.007 & 0.925$\pm$0.002 & 0.888$\pm$0.020 & 0.942$\pm$0.001 & 0.827$\pm$0.001 & 0.676$\pm$0.003 & 0.895$\pm$0.003 & 0.779$\pm$0.006 \\
MMAE-raw-z & 0.508$\pm$0.018 & 0.936$\pm$0.002 & 0.913$\pm$0.017 & 0.946$\pm$0.002 & 0.925$\pm$0.007 & 0.680$\pm$0.007 & 0.902$\pm$0.002 & 0.766$\pm$0.011 \\
MMAE-pca & 0.458$\pm$0.013 & 0.925$\pm$0.002 & 0.832$\pm$0.001 & 0.946$\pm$0.002 & 0.829$\pm$0.001 & 0.666$\pm$0.009 & 0.891$\pm$0.002 & 0.779$\pm$0.007 \\
MMAE-pca-z & 0.496$\pm$0.021 & 0.936$\pm$0.001 & 0.836$\pm$0.001 & \textbf{0.958$\pm$0.002} & 0.829$\pm$0.001 & 0.707$\pm$0.007 & 0.915$\pm$0.003 & 0.773$\pm$0.005 \\
MMAE-Isomap & 0.560$\pm$0.005 & 0.904$\pm$0.005 & 0.849$\pm$0.007 & 0.953$\pm$0.006 & 0.833$\pm$0.001 & 0.701$\pm$0.001 & 0.908$\pm$0.002 & 0.775$\pm$0.005 \\
MMAE-Isomap-z & 0.555$\pm$0.013 & 0.926$\pm$0.002 & 0.842$\pm$0.008 & 0.945$\pm$0.001 & 0.911$\pm$0.004 & \textbf{0.709$\pm$0.008} & \textbf{0.920$\pm$0.002} & 0.780$\pm$0.005 \\
SPAE & 0.501$\pm$0.016 & \textbf{0.937$\pm$0.003} & 0.904$\pm$0.027 & 0.928$\pm$0.007 & 0.945$\pm$0.003 & 0.676$\pm$0.013 & 0.886$\pm$0.004 & 0.742$\pm$0.008 \\
GRAE-pca & 0.440$\pm$0.008 & 0.902$\pm$0.002 & 0.832$\pm$0.001 & 0.952$\pm$0.001 & 0.830$\pm$0.001 & 0.622$\pm$0.008 & 0.891$\pm$0.002 & 0.769$\pm$0.007 \\
GRAE-Isomap & 0.546$\pm$0.005 & 0.827$\pm$0.013 & 0.886$\pm$0.001 & 0.943$\pm$0.001 & 0.834$\pm$0.001 & 0.678$\pm$0.003 & 0.897$\pm$0.002 & 0.760$\pm$0.006 \\
GRAE-umap & 0.472$\pm$0.007 & 0.837$\pm$0.007 & 0.870$\pm$0.008 & 0.950$\pm$0.001 & 0.840$\pm$0.003 & 0.667$\pm$0.006 & 0.876$\pm$0.002 & 0.753$\pm$0.007 \\
TopoAE & \textbf{0.571$\pm$0.007} & 0.935$\pm$0.001 & \textbf{0.935$\pm$0.033} & 0.952$\pm$0.007 & 0.921$\pm$0.009 & 0.701$\pm$0.005 & 0.916$\pm$0.002 & \textbf{0.781$\pm$0.006} \\
RTD-AE & 0.513$\pm$0.009 & 0.936$\pm$0.002 & 0.929$\pm$0.005 & 0.936$\pm$0.008 & 0.847$\pm$0.018 & 0.694$\pm$0.003 & 0.910$\pm$0.003 & 0.780$\pm$0.005 \\
GeomAE & 0.398$\pm$0.010 & 0.928$\pm$0.007 & 0.865$\pm$0.029 & 0.955$\pm$0.003 & 0.939$\pm$0.010 & 0.667$\pm$0.014 & 0.865$\pm$0.025 & 0.743$\pm$0.008 \\
GGAE & 0.480$\pm$0.016 & 0.924$\pm$0.002 & 0.902$\pm$0.016 & 0.927$\pm$0.004 & 0.861$\pm$0.007 & -- & -- & -- \\
IsoAE & 0.450$\pm$0.014 & 0.882$\pm$0.015 & 0.896$\pm$0.033 & 0.915$\pm$0.008 & 0.921$\pm$0.008 & 0.591$\pm$0.011 & 0.801$\pm$0.045 & 0.707$\pm$0.016 \\
CAE & 0.471$\pm$0.004 & 0.928$\pm$0.014 & 0.891$\pm$0.019 & 0.955$\pm$0.009 & 0.924$\pm$0.008 & 0.654$\pm$0.010 & 0.909$\pm$0.004 & 0.764$\pm$0.006 \\
\bottomrule
\end{tabular}

\par\vspace{1.4em}

\begin{tabular}{lrrrrrrr}
\toprule
method & celeba64 & shapes3d & mpi3d & smallnorb & dsprites & pbmc3k & paul15 \\
\midrule
Vanilla AE & 0.752$\pm$0.006 & 0.844$\pm$0.008 & 0.861$\pm$0.002 & 0.898$\pm$0.009 & 0.859$\pm$0.014 & 0.475$\pm$0.030 & 0.446$\pm$0.010 \\
MMAE-raw & 0.777$\pm$0.005 & 0.756$\pm$0.008 & 0.814$\pm$0.001 & 0.916$\pm$0.003 & 0.870$\pm$0.003 & 0.480$\pm$0.006 & 0.518$\pm$0.006 \\
MMAE-raw-z & 0.788$\pm$0.004 & 0.836$\pm$0.012 & 0.825$\pm$0.003 & \textbf{0.921$\pm$0.001} & 0.846$\pm$0.025 & \textbf{0.536$\pm$0.006} & \textbf{0.574$\pm$0.008} \\
MMAE-pca & 0.780$\pm$0.003 & 0.756$\pm$0.007 & \textbf{0.873$\pm$0.002} & 0.919$\pm$0.003 & 0.881$\pm$0.003 & 0.508$\pm$0.005 & 0.424$\pm$0.047 \\
MMAE-pca-z & 0.787$\pm$0.004 & \textbf{0.874$\pm$0.011} & 0.867$\pm$0.001 & 0.918$\pm$0.002 & 0.883$\pm$0.003 & 0.418$\pm$0.011 & 0.457$\pm$0.005 \\
MMAE-Isomap & 0.782$\pm$0.004 & 0.782$\pm$0.017 & 0.840$\pm$0.003 & 0.898$\pm$0.005 & 0.901$\pm$0.002 & 0.353$\pm$0.016 & 0.479$\pm$0.005 \\
MMAE-Isomap-z & 0.784$\pm$0.003 & 0.873$\pm$0.012 & 0.847$\pm$0.003 & 0.918$\pm$0.001 & 0.892$\pm$0.003 & 0.351$\pm$0.014 & 0.495$\pm$0.004 \\
SPAE & 0.755$\pm$0.008 & 0.794$\pm$0.023 & 0.770$\pm$0.014 & 0.900$\pm$0.003 & 0.871$\pm$0.019 & 0.422$\pm$0.048 & 0.491$\pm$0.015 \\
GRAE-pca & 0.768$\pm$0.002 & 0.752$\pm$0.003 & 0.869$\pm$0.001 & 0.907$\pm$0.003 & 0.884$\pm$0.003 & 0.424$\pm$0.005 & 0.416$\pm$0.002 \\
GRAE-Isomap & 0.769$\pm$0.002 & 0.771$\pm$0.003 & 0.839$\pm$0.002 & 0.880$\pm$0.005 & 0.906$\pm$0.002 & 0.404$\pm$0.008 & 0.492$\pm$0.002 \\
GRAE-umap & 0.779$\pm$0.002 & 0.797$\pm$0.003 & 0.851$\pm$0.002 & 0.816$\pm$0.008 & 0.906$\pm$0.002 & 0.372$\pm$0.005 & 0.452$\pm$0.004 \\
TopoAE & 0.786$\pm$0.003 & 0.810$\pm$0.009 & 0.829$\pm$0.003 & 0.919$\pm$0.001 & 0.881$\pm$0.003 & 0.509$\pm$0.005 & 0.491$\pm$0.010 \\
RTD-AE & \textbf{0.793$\pm$0.002} & 0.783$\pm$0.021 & 0.800$\pm$0.011 & 0.917$\pm$0.004 & 0.895$\pm$0.002 & 0.438$\pm$0.020 & 0.508$\pm$0.010 \\
GeomAE & 0.760$\pm$0.006 & 0.865$\pm$0.018 & 0.860$\pm$0.007 & 0.863$\pm$0.010 & 0.834$\pm$0.012 & 0.481$\pm$0.021 & 0.435$\pm$0.020 \\
GGAE & -- & -- & -- & -- & -- & 0.405$\pm$0.008 & 0.418$\pm$0.011 \\
IsoAE & 0.679$\pm$0.011 & 0.674$\pm$0.034 & 0.847$\pm$0.007 & 0.858$\pm$0.010 & 0.696$\pm$0.041 & 0.390$\pm$0.009 & 0.412$\pm$0.027 \\
CAE & 0.767$\pm$0.003 & 0.865$\pm$0.005 & 0.840$\pm$0.007 & 0.891$\pm$0.008 & 0.903$\pm$0.002 & 0.514$\pm$0.004 & 0.559$\pm$0.018 \\
\bottomrule
\end{tabular}
\end{sidewaystable}
% W0 @ bottleneck=2  (higher better: False)
\begin{sidewaystable}[p]
\centering
\footnotesize
\caption{\textbf{Persistence-Wasserstein $W_0$ ($\downarrow$)} per dataset at the 2D bottleneck; mean$\pm$std over 5 seeds, \textbf{best} per column in bold; ``--'' marks GGAE where its $O(n^2)$ kernel exceeds memory. Datasets are split across two blocks for legibility.}
\label{tab:app-W0}
\begin{tabular}{lrrrrrrrr}
\toprule
method & nested\_spheres & linked\_tori & swiss\_roll & mammoth & earth & mnist & fashion\_mnist & cifar10 \\
\midrule
Vanilla AE & 178.475$\pm$2.531 & 18.792$\pm$0.903 & 9.380$\pm$1.551 & 93.040$\pm$0.154 & \textbf{51.825$\pm$0.390} & 182.698$\pm$0.842 & 128.072$\pm$0.960 & 164.384$\pm$1.453 \\
MMAE-raw & 179.054$\pm$2.487 & 16.183$\pm$0.065 & 10.254$\pm$0.592 & 94.005$\pm$0.146 & 56.992$\pm$0.171 & 183.709$\pm$1.100 & 127.836$\pm$0.870 & 163.313$\pm$1.781 \\
MMAE-raw-z & 176.844$\pm$2.488 & 16.572$\pm$0.180 & 10.092$\pm$0.613 & 93.773$\pm$0.264 & 56.633$\pm$0.228 & \textbf{180.287$\pm$0.818} & 126.786$\pm$0.950 & 164.457$\pm$1.841 \\
MMAE-pca & 178.911$\pm$2.435 & 16.200$\pm$0.089 & 9.709$\pm$0.063 & 94.068$\pm$0.278 & 56.451$\pm$0.116 & 185.489$\pm$1.256 & 129.205$\pm$0.786 & \textbf{163.163$\pm$1.814} \\
MMAE-pca-z & 178.082$\pm$2.441 & 16.625$\pm$0.149 & 9.549$\pm$0.043 & 93.508$\pm$0.411 & 56.529$\pm$0.133 & 183.744$\pm$1.401 & 129.119$\pm$0.733 & 163.875$\pm$1.829 \\
MMAE-Isomap & 176.136$\pm$2.569 & 19.000$\pm$0.193 & 15.484$\pm$0.551 & 93.363$\pm$0.329 & 56.775$\pm$0.099 & 185.432$\pm$1.229 & 128.810$\pm$0.795 & 165.034$\pm$1.603 \\
MMAE-Isomap-z & \textbf{173.582$\pm$2.249} & 18.572$\pm$0.336 & 16.255$\pm$0.278 & 92.600$\pm$0.226 & 57.220$\pm$0.095 & 182.682$\pm$0.960 & 128.773$\pm$0.693 & 163.867$\pm$1.590 \\
SPAE & 178.457$\pm$2.462 & 16.444$\pm$0.172 & 8.937$\pm$0.480 & 93.800$\pm$0.466 & 55.282$\pm$0.347 & 183.676$\pm$1.070 & \textbf{125.905$\pm$1.059} & 165.610$\pm$1.674 \\
GRAE-pca & 178.968$\pm$2.474 & 16.198$\pm$0.219 & 9.401$\pm$0.133 & 93.500$\pm$0.220 & 56.699$\pm$0.149 & 185.018$\pm$1.494 & 129.191$\pm$0.763 & 164.051$\pm$1.618 \\
GRAE-Isomap & 176.150$\pm$2.426 & 21.025$\pm$0.212 & 14.267$\pm$0.133 & 92.476$\pm$0.333 & 56.819$\pm$0.156 & 184.696$\pm$1.246 & 128.927$\pm$0.717 & 164.450$\pm$1.572 \\
GRAE-umap & 179.382$\pm$2.448 & 21.936$\pm$0.143 & 14.550$\pm$0.198 & \textbf{92.261$\pm$0.286} & 55.993$\pm$0.062 & 186.183$\pm$1.214 & 129.494$\pm$0.701 & 165.636$\pm$1.549 \\
TopoAE & 174.942$\pm$2.675 & 16.276$\pm$0.358 & \textbf{8.494$\pm$0.398} & 94.470$\pm$0.337 & 55.141$\pm$0.338 & 182.763$\pm$1.275 & 129.145$\pm$0.701 & 165.135$\pm$1.709 \\
RTD-AE & 176.620$\pm$2.338 & \textbf{16.082$\pm$0.165} & 9.058$\pm$0.218 & 94.177$\pm$0.267 & 55.479$\pm$0.235 & 184.224$\pm$1.408 & 128.667$\pm$0.905 & 164.400$\pm$1.725 \\
GeomAE & 177.965$\pm$2.430 & 18.324$\pm$0.806 & 13.019$\pm$1.686 & 93.562$\pm$0.286 & 55.621$\pm$0.460 & 182.499$\pm$0.903 & 127.763$\pm$0.940 & 164.171$\pm$1.735 \\
GGAE & 178.099$\pm$2.353 & 17.182$\pm$0.229 & 11.950$\pm$0.585 & 93.312$\pm$0.248 & 58.158$\pm$0.126 & -- & -- & -- \\
IsoAE & 180.474$\pm$2.404 & 17.652$\pm$0.717 & 10.238$\pm$0.680 & 92.472$\pm$0.710 & 55.504$\pm$0.377 & 185.767$\pm$1.212 & 128.294$\pm$0.659 & 163.266$\pm$2.328 \\
CAE & 178.912$\pm$2.491 & 16.605$\pm$0.323 & 9.175$\pm$0.682 & 93.575$\pm$0.231 & 55.197$\pm$0.434 & 184.830$\pm$1.042 & 128.487$\pm$0.881 & 164.318$\pm$1.799 \\
\bottomrule
\end{tabular}

\par\vspace{1.4em}

\begin{tabular}{lrrrrrrr}
\toprule
method & celeba64 & shapes3d & mpi3d & smallnorb & dsprites & pbmc3k & paul15 \\
\midrule
Vanilla AE & 161.451$\pm$1.134 & 110.579$\pm$0.838 & 128.891$\pm$0.951 & 61.813$\pm$0.905 & 140.045$\pm$0.589 & 150.991$\pm$0.143 & 127.178$\pm$0.347 \\
MMAE-raw & 161.728$\pm$0.873 & 110.450$\pm$0.664 & 130.551$\pm$0.575 & 62.519$\pm$0.944 & 140.678$\pm$0.280 & 154.734$\pm$0.077 & 129.374$\pm$0.297 \\
MMAE-raw-z & 162.296$\pm$1.020 & \textbf{107.810$\pm$0.709} & \textbf{125.704$\pm$0.956} & 62.524$\pm$0.891 & \textbf{139.466$\pm$0.628} & 153.695$\pm$0.627 & 126.283$\pm$0.333 \\
MMAE-pca & 162.998$\pm$0.967 & 113.287$\pm$0.512 & 128.312$\pm$0.819 & 62.855$\pm$0.953 & 143.782$\pm$0.297 & 154.485$\pm$0.334 & 125.340$\pm$0.619 \\
MMAE-pca-z & 162.173$\pm$0.943 & 109.570$\pm$0.382 & 130.446$\pm$0.631 & 62.660$\pm$0.919 & 143.260$\pm$0.157 & 151.798$\pm$0.214 & 126.240$\pm$0.203 \\
MMAE-Isomap & 163.145$\pm$1.141 & 110.702$\pm$0.866 & 131.188$\pm$0.455 & 61.903$\pm$0.956 & 142.473$\pm$0.243 & 153.846$\pm$0.680 & 129.059$\pm$0.252 \\
MMAE-Isomap-z & 162.573$\pm$1.082 & 110.672$\pm$0.824 & 129.035$\pm$0.579 & 62.151$\pm$1.018 & 140.045$\pm$0.445 & 153.994$\pm$0.552 & 128.343$\pm$0.041 \\
SPAE & 163.095$\pm$1.091 & 111.654$\pm$0.522 & 128.825$\pm$1.071 & 62.436$\pm$0.903 & 140.901$\pm$0.328 & 154.773$\pm$0.708 & 129.240$\pm$0.403 \\
GRAE-pca & 162.082$\pm$0.917 & 112.115$\pm$0.395 & 130.907$\pm$0.523 & 62.500$\pm$0.915 & 143.612$\pm$0.222 & \textbf{149.269$\pm$0.162} & 127.577$\pm$0.116 \\
GRAE-Isomap & 162.450$\pm$1.060 & 112.765$\pm$0.335 & 129.823$\pm$0.526 & 62.278$\pm$0.951 & 143.554$\pm$0.347 & 151.451$\pm$0.495 & 128.614$\pm$0.149 \\
GRAE-umap & 162.593$\pm$1.082 & 109.994$\pm$0.345 & 131.896$\pm$0.473 & 61.198$\pm$1.016 & 146.552$\pm$0.273 & 153.862$\pm$0.330 & 128.570$\pm$0.129 \\
TopoAE & 163.047$\pm$0.958 & 111.414$\pm$0.567 & 133.235$\pm$0.717 & 62.521$\pm$0.921 & 142.209$\pm$0.304 & 154.695$\pm$0.152 & 128.735$\pm$0.540 \\
RTD-AE & 162.141$\pm$0.936 & 110.158$\pm$0.479 & 133.666$\pm$0.915 & 61.755$\pm$0.878 & 142.949$\pm$0.400 & 154.310$\pm$0.833 & 126.903$\pm$0.521 \\
GeomAE & 161.696$\pm$1.163 & 110.752$\pm$0.884 & 126.587$\pm$1.297 & 61.401$\pm$0.947 & 140.733$\pm$0.400 & 151.054$\pm$0.542 & \textbf{124.961$\pm$0.158} \\
GGAE & -- & -- & -- & -- & -- & 151.186$\pm$0.275 & 127.257$\pm$0.104 \\
IsoAE & \textbf{156.981$\pm$1.650} & 110.561$\pm$0.779 & 128.588$\pm$1.007 & \textbf{61.065$\pm$0.974} & 141.334$\pm$1.318 & 153.768$\pm$0.463 & 127.221$\pm$0.531 \\
CAE & 160.716$\pm$1.074 & 111.617$\pm$0.836 & 132.969$\pm$0.713 & 62.035$\pm$1.073 & 144.603$\pm$0.523 & 152.802$\pm$0.855 & 126.021$\pm$0.541 \\
\bottomrule
\end{tabular}
\end{sidewaystable}
% W1 @ bottleneck=2  (higher better: False)
\begin{sidewaystable}[p]
\centering
\footnotesize
\caption{\textbf{Persistence-Wasserstein $W_1$ ($\downarrow$)} per dataset at the 2D bottleneck; mean$\pm$std over 5 seeds, \textbf{best} per column in bold; ``--'' marks GGAE where its $O(n^2)$ kernel exceeds memory. Datasets are split across two blocks for legibility.}
\label{tab:app-W1}
\begin{tabular}{lrrrrrrrr}
\toprule
method & nested\_spheres & linked\_tori & swiss\_roll & mammoth & earth & mnist & fashion\_mnist & cifar10 \\
\midrule
Vanilla AE & 2.592$\pm$0.095 & 3.444$\pm$0.083 & 1.977$\pm$0.166 & 2.582$\pm$0.073 & 3.141$\pm$0.065 & 9.230$\pm$0.222 & 3.596$\pm$0.106 & 2.243$\pm$0.126 \\
MMAE-raw & 3.024$\pm$0.123 & 3.222$\pm$0.096 & \textbf{1.797$\pm$0.080} & 2.726$\pm$0.071 & 3.213$\pm$0.067 & 9.521$\pm$0.286 & 3.871$\pm$0.111 & 2.349$\pm$0.080 \\
MMAE-raw-z & 2.890$\pm$0.115 & 3.183$\pm$0.148 & 1.980$\pm$0.121 & 2.610$\pm$0.050 & 3.361$\pm$0.078 & 9.229$\pm$0.250 & 3.611$\pm$0.118 & 2.267$\pm$0.078 \\
MMAE-pca & 2.836$\pm$0.086 & 3.182$\pm$0.076 & 2.108$\pm$0.063 & 2.659$\pm$0.046 & 3.465$\pm$0.054 & 9.564$\pm$0.293 & 3.794$\pm$0.082 & 2.353$\pm$0.058 \\
MMAE-pca-z & 2.776$\pm$0.110 & 3.142$\pm$0.123 & 2.128$\pm$0.026 & 2.600$\pm$0.056 & 3.461$\pm$0.046 & 9.345$\pm$0.271 & 3.771$\pm$0.097 & 2.212$\pm$0.024 \\
MMAE-Isomap & 3.276$\pm$0.109 & 3.286$\pm$0.047 & 2.314$\pm$0.077 & 2.507$\pm$0.024 & 3.429$\pm$0.053 & 9.654$\pm$0.245 & 3.654$\pm$0.101 & 2.342$\pm$0.058 \\
MMAE-Isomap-z & 3.241$\pm$0.046 & 3.242$\pm$0.080 & 2.298$\pm$0.068 & 2.349$\pm$0.050 & 3.057$\pm$0.035 & 9.269$\pm$0.212 & 3.538$\pm$0.105 & 2.314$\pm$0.068 \\
SPAE & 3.233$\pm$0.099 & 3.152$\pm$0.033 & 2.011$\pm$0.078 & 3.047$\pm$0.127 & 3.622$\pm$0.052 & 9.563$\pm$0.282 & 3.683$\pm$0.055 & 2.416$\pm$0.027 \\
GRAE-pca & 2.626$\pm$0.066 & 3.237$\pm$0.052 & 2.106$\pm$0.062 & 2.549$\pm$0.036 & 3.428$\pm$0.017 & 9.539$\pm$0.277 & 3.781$\pm$0.076 & 2.278$\pm$0.068 \\
GRAE-Isomap & 3.125$\pm$0.067 & 3.091$\pm$0.068 & 2.537$\pm$0.044 & \textbf{2.312$\pm$0.045} & 3.416$\pm$0.028 & 9.376$\pm$0.264 & 3.596$\pm$0.080 & 2.178$\pm$0.091 \\
GRAE-umap & 2.858$\pm$0.087 & 3.332$\pm$0.033 & 2.446$\pm$0.024 & 2.332$\pm$0.049 & 3.603$\pm$0.043 & 9.059$\pm$0.276 & 3.391$\pm$0.104 & 2.247$\pm$0.028 \\
TopoAE & 3.229$\pm$0.076 & 3.230$\pm$0.088 & 1.935$\pm$0.130 & 2.917$\pm$0.079 & 3.627$\pm$0.090 & 9.326$\pm$0.306 & 3.899$\pm$0.136 & 2.360$\pm$0.069 \\
RTD-AE & 3.185$\pm$0.098 & 3.164$\pm$0.042 & 1.807$\pm$0.061 & 3.087$\pm$0.027 & 3.618$\pm$0.068 & 9.555$\pm$0.273 & 3.939$\pm$0.081 & 2.388$\pm$0.057 \\
GeomAE & \textbf{2.544$\pm$0.110} & 3.371$\pm$0.091 & 2.274$\pm$0.223 & 2.613$\pm$0.073 & 3.386$\pm$0.149 & 9.171$\pm$0.310 & \textbf{3.374$\pm$0.126} & 2.101$\pm$0.172 \\
GGAE & 2.792$\pm$0.104 & \textbf{3.034$\pm$0.047} & 2.248$\pm$0.250 & 2.581$\pm$0.036 & \textbf{2.727$\pm$0.078} & -- & -- & -- \\
IsoAE & 2.825$\pm$0.176 & 3.151$\pm$0.088 & 2.045$\pm$0.157 & 2.413$\pm$0.084 & 3.528$\pm$0.149 & 9.232$\pm$0.147 & 3.449$\pm$0.116 & \textbf{1.948$\pm$0.229} \\
CAE & 2.612$\pm$0.079 & 3.136$\pm$0.062 & 1.886$\pm$0.108 & 2.595$\pm$0.043 & 3.343$\pm$0.203 & 9.499$\pm$0.237 & 3.571$\pm$0.164 & 2.293$\pm$0.046 \\
\bottomrule
\end{tabular}

\par\vspace{1.4em}

\begin{tabular}{lrrrrrrr}
\toprule
method & celeba64 & shapes3d & mpi3d & smallnorb & dsprites & pbmc3k & paul15 \\
\midrule
Vanilla AE & 3.132$\pm$0.162 & 11.271$\pm$0.096 & 2.522$\pm$0.080 & 1.947$\pm$0.040 & 4.968$\pm$0.136 & 0.262$\pm$0.030 & 0.385$\pm$0.078 \\
MMAE-raw & 3.383$\pm$0.118 & 11.197$\pm$0.202 & 3.303$\pm$0.108 & 1.932$\pm$0.050 & 5.152$\pm$0.085 & 0.798$\pm$0.055 & 0.858$\pm$0.054 \\
MMAE-raw-z & 3.343$\pm$0.178 & 10.845$\pm$0.148 & 2.995$\pm$0.120 & 1.824$\pm$0.022 & 4.869$\pm$0.132 & 0.597$\pm$0.101 & 0.666$\pm$0.093 \\
MMAE-pca & 3.419$\pm$0.148 & 11.264$\pm$0.197 & 2.586$\pm$0.113 & 1.922$\pm$0.040 & 5.250$\pm$0.115 & 0.769$\pm$0.093 & \textbf{0.116$\pm$0.074} \\
MMAE-pca-z & 3.336$\pm$0.129 & 11.019$\pm$0.172 & 2.463$\pm$0.092 & 1.835$\pm$0.037 & 5.254$\pm$0.139 & 0.364$\pm$0.049 & 0.578$\pm$0.145 \\
MMAE-Isomap & 3.412$\pm$0.159 & 11.022$\pm$0.276 & 2.589$\pm$0.086 & 1.941$\pm$0.047 & 5.296$\pm$0.111 & 0.720$\pm$0.048 & 0.837$\pm$0.093 \\
MMAE-Isomap-z & 3.250$\pm$0.158 & 11.068$\pm$0.150 & 2.476$\pm$0.100 & 1.888$\pm$0.047 & 5.120$\pm$0.109 & 0.735$\pm$0.095 & 0.476$\pm$0.026 \\
SPAE & 3.451$\pm$0.141 & 11.570$\pm$0.146 & 3.251$\pm$0.112 & 2.202$\pm$0.079 & 5.222$\pm$0.106 & 0.854$\pm$0.100 & 0.876$\pm$0.048 \\
GRAE-pca & 3.270$\pm$0.172 & 10.954$\pm$0.117 & 2.498$\pm$0.108 & 1.810$\pm$0.048 & 5.235$\pm$0.122 & \textbf{0.129$\pm$0.033} & 0.390$\pm$0.030 \\
GRAE-Isomap & 3.208$\pm$0.185 & 11.076$\pm$0.112 & 2.507$\pm$0.090 & 1.792$\pm$0.024 & 5.259$\pm$0.112 & 0.365$\pm$0.031 & 0.485$\pm$0.037 \\
GRAE-umap & 3.138$\pm$0.189 & 10.737$\pm$0.121 & 2.618$\pm$0.116 & 2.065$\pm$0.049 & 5.282$\pm$0.131 & 0.600$\pm$0.035 & 0.483$\pm$0.056 \\
TopoAE & 3.342$\pm$0.132 & 11.522$\pm$0.143 & 3.394$\pm$0.102 & 2.082$\pm$0.040 & 5.186$\pm$0.150 & 0.760$\pm$0.086 & 0.671$\pm$0.057 \\
RTD-AE & 3.276$\pm$0.161 & 11.406$\pm$0.140 & 3.503$\pm$0.143 & 2.119$\pm$0.072 & 5.318$\pm$0.118 & 0.839$\pm$0.070 & 0.705$\pm$0.036 \\
GeomAE & 3.095$\pm$0.117 & 10.842$\pm$0.142 & \textbf{2.463$\pm$0.103} & 1.997$\pm$0.062 & 4.914$\pm$0.226 & 0.270$\pm$0.035 & 0.161$\pm$0.062 \\
GGAE & -- & -- & -- & -- & -- & 0.424$\pm$0.037 & 0.429$\pm$0.055 \\
IsoAE & \textbf{2.287$\pm$0.324} & 10.776$\pm$0.291 & 2.554$\pm$0.111 & \textbf{1.540$\pm$0.031} & \textbf{4.327$\pm$0.196} & 0.755$\pm$0.134 & 0.532$\pm$0.061 \\
CAE & 3.175$\pm$0.128 & 11.130$\pm$0.219 & 3.167$\pm$0.107 & 2.206$\pm$0.037 & 5.311$\pm$0.124 & 0.456$\pm$0.044 & 0.554$\pm$0.093 \\
\bottomrule
\end{tabular}
\end{sidewaystable}
% KL0.1 @ bottleneck=2  (higher better: False)
\begin{sidewaystable}[p]
\centering
\footnotesize
\caption{\textbf{Density KL at $\tau{=}0.1$ ($\downarrow$)} per dataset at the 2D bottleneck; mean$\pm$std over 5 seeds, \textbf{best} per column in bold; ``--'' marks GGAE where its $O(n^2)$ kernel exceeds memory. Datasets are split across two blocks for legibility.}
\label{tab:app-KL0.1}
\begin{tabular}{lrrrrrrrr}
\toprule
method & nested\_spheres & linked\_tori & swiss\_roll & mammoth & earth & mnist & fashion\_mnist & cifar10 \\
\midrule
Vanilla AE & 0.552$\pm$0.066 & 0.037$\pm$0.010 & 0.267$\pm$0.152 & 0.050$\pm$0.007 & 0.287$\pm$0.019 & 0.227$\pm$0.030 & 0.189$\pm$0.037 & 0.146$\pm$0.020 \\
MMAE-raw & \textbf{0.028$\pm$0.006} & \textbf{0.026$\pm$0.002} & 0.060$\pm$0.009 & 0.028$\pm$0.002 & 0.055$\pm$0.002 & 0.144$\pm$0.018 & 0.104$\pm$0.006 & 0.184$\pm$0.007 \\
MMAE-raw-z & 0.045$\pm$0.021 & 0.034$\pm$0.007 & 0.060$\pm$0.017 & 0.028$\pm$0.002 & 0.033$\pm$0.003 & 0.158$\pm$0.020 & 0.126$\pm$0.012 & 0.141$\pm$0.010 \\
MMAE-pca & 0.076$\pm$0.012 & 0.027$\pm$0.002 & 0.042$\pm$0.002 & \textbf{0.024$\pm$0.002} & 0.024$\pm$0.001 & 0.237$\pm$0.010 & 0.086$\pm$0.004 & 0.192$\pm$0.007 \\
MMAE-pca-z & 0.061$\pm$0.009 & 0.032$\pm$0.006 & 0.042$\pm$0.001 & 0.032$\pm$0.001 & 0.025$\pm$0.001 & 0.321$\pm$0.021 & 0.084$\pm$0.007 & 0.167$\pm$0.005 \\
MMAE-Isomap & 0.077$\pm$0.006 & 0.043$\pm$0.004 & 0.056$\pm$0.012 & 0.031$\pm$0.005 & 0.024$\pm$0.001 & 0.173$\pm$0.011 & 0.091$\pm$0.005 & 0.123$\pm$0.004 \\
MMAE-Isomap-z & 0.109$\pm$0.011 & 0.037$\pm$0.003 & 0.048$\pm$0.024 & 0.028$\pm$0.000 & 0.049$\pm$0.005 & 0.230$\pm$0.029 & 0.098$\pm$0.008 & 0.178$\pm$0.005 \\
SPAE & 0.070$\pm$0.010 & 0.027$\pm$0.004 & 0.065$\pm$0.027 & 0.042$\pm$0.006 & 0.044$\pm$0.005 & 0.163$\pm$0.012 & 0.187$\pm$0.015 & 0.109$\pm$0.005 \\
GRAE-pca & 0.453$\pm$0.009 & 0.050$\pm$0.005 & 0.045$\pm$0.003 & 0.035$\pm$0.001 & 0.030$\pm$0.002 & 0.222$\pm$0.016 & 0.086$\pm$0.005 & 0.172$\pm$0.003 \\
GRAE-Isomap & 0.166$\pm$0.007 & 0.144$\pm$0.009 & \textbf{0.024$\pm$0.004} & 0.026$\pm$0.001 & 0.025$\pm$0.001 & 0.234$\pm$0.012 & 0.076$\pm$0.006 & 0.145$\pm$0.004 \\
GRAE-umap & 0.214$\pm$0.019 & 0.054$\pm$0.006 & 0.069$\pm$0.017 & 0.026$\pm$0.001 & 0.021$\pm$0.002 & 0.098$\pm$0.009 & 0.049$\pm$0.002 & 0.086$\pm$0.002 \\
TopoAE & 0.129$\pm$0.019 & 0.033$\pm$0.004 & 0.079$\pm$0.019 & 0.031$\pm$0.007 & 0.069$\pm$0.008 & 0.230$\pm$0.012 & 0.107$\pm$0.008 & 0.138$\pm$0.006 \\
RTD-AE & 0.149$\pm$0.022 & 0.046$\pm$0.003 & 0.043$\pm$0.004 & 0.039$\pm$0.005 & 0.042$\pm$0.021 & 0.178$\pm$0.013 & 0.127$\pm$0.012 & 0.159$\pm$0.009 \\
GeomAE & 0.654$\pm$0.039 & 0.030$\pm$0.002 & 0.064$\pm$0.010 & 0.025$\pm$0.002 & 0.078$\pm$0.028 & 0.205$\pm$0.031 & 0.102$\pm$0.013 & 0.120$\pm$0.014 \\
GGAE & 0.060$\pm$0.043 & 0.087$\pm$0.005 & 0.074$\pm$0.019 & 0.067$\pm$0.004 & 0.039$\pm$0.009 & -- & -- & -- \\
IsoAE & 0.183$\pm$0.071 & 0.110$\pm$0.032 & 0.092$\pm$0.043 & 0.075$\pm$0.025 & 0.088$\pm$0.020 & 0.108$\pm$0.023 & 0.095$\pm$0.021 & 0.159$\pm$0.039 \\
CAE & 0.258$\pm$0.024 & 0.037$\pm$0.006 & 0.077$\pm$0.030 & 0.031$\pm$0.003 & 0.197$\pm$0.032 & 0.202$\pm$0.017 & 0.107$\pm$0.012 & 0.163$\pm$0.009 \\
\bottomrule
\end{tabular}

\par\vspace{1.4em}

\begin{tabular}{lrrrrrrr}
\toprule
method & celeba64 & shapes3d & mpi3d & smallnorb & dsprites & pbmc3k & paul15 \\
\midrule
Vanilla AE & 0.140$\pm$0.013 & 0.219$\pm$0.037 & 0.131$\pm$0.009 & 0.387$\pm$0.058 & 0.225$\pm$0.040 & 0.210$\pm$0.019 & 0.137$\pm$0.020 \\
MMAE-raw & 0.155$\pm$0.016 & 0.082$\pm$0.008 & 0.123$\pm$0.006 & 0.251$\pm$0.027 & 0.174$\pm$0.011 & 0.096$\pm$0.006 & 0.168$\pm$0.013 \\
MMAE-raw-z & 0.140$\pm$0.007 & 0.098$\pm$0.003 & 0.149$\pm$0.009 & 0.350$\pm$0.008 & 0.151$\pm$0.018 & 0.092$\pm$0.011 & 0.231$\pm$0.033 \\
MMAE-pca & 0.098$\pm$0.013 & 0.053$\pm$0.011 & 0.114$\pm$0.011 & 0.285$\pm$0.015 & 0.216$\pm$0.008 & 0.119$\pm$0.008 & 0.131$\pm$0.022 \\
MMAE-pca-z & 0.137$\pm$0.012 & \textbf{0.037$\pm$0.005} & 0.103$\pm$0.008 & 0.385$\pm$0.011 & 0.222$\pm$0.005 & 0.113$\pm$0.012 & 0.156$\pm$0.019 \\
MMAE-Isomap & 0.078$\pm$0.010 & 0.070$\pm$0.004 & 0.172$\pm$0.015 & \textbf{0.172$\pm$0.007} & 0.174$\pm$0.006 & 0.143$\pm$0.016 & 0.165$\pm$0.028 \\
MMAE-Isomap-z & 0.090$\pm$0.005 & 0.038$\pm$0.008 & 0.173$\pm$0.021 & 0.225$\pm$0.027 & 0.184$\pm$0.019 & 0.095$\pm$0.022 & 0.178$\pm$0.017 \\
SPAE & 0.110$\pm$0.005 & 0.141$\pm$0.017 & 0.168$\pm$0.015 & 0.304$\pm$0.049 & 0.204$\pm$0.011 & 0.190$\pm$0.028 & 0.182$\pm$0.025 \\
GRAE-pca & 0.132$\pm$0.005 & 0.048$\pm$0.002 & \textbf{0.102$\pm$0.008} & 0.380$\pm$0.030 & 0.202$\pm$0.006 & 0.076$\pm$0.011 & 0.147$\pm$0.004 \\
GRAE-Isomap & 0.085$\pm$0.006 & 0.078$\pm$0.005 & 0.164$\pm$0.015 & 0.305$\pm$0.029 & 0.141$\pm$0.006 & 0.082$\pm$0.013 & 0.180$\pm$0.008 \\
GRAE-umap & 0.047$\pm$0.004 & 0.052$\pm$0.005 & 0.127$\pm$0.009 & 0.236$\pm$0.017 & 0.042$\pm$0.004 & 0.189$\pm$0.015 & 0.113$\pm$0.010 \\
TopoAE & 0.095$\pm$0.009 & 0.135$\pm$0.007 & 0.112$\pm$0.008 & 0.320$\pm$0.025 & 0.136$\pm$0.012 & 0.107$\pm$0.013 & 0.231$\pm$0.029 \\
RTD-AE & 0.131$\pm$0.008 & 0.116$\pm$0.013 & 0.104$\pm$0.008 & 0.175$\pm$0.012 & 0.187$\pm$0.008 & 0.248$\pm$0.019 & 0.218$\pm$0.042 \\
GeomAE & 0.106$\pm$0.014 & 0.040$\pm$0.018 & 0.148$\pm$0.029 & 0.196$\pm$0.018 & 0.216$\pm$0.030 & 0.142$\pm$0.031 & 0.116$\pm$0.026 \\
GGAE & -- & -- & -- & -- & -- & \textbf{0.058$\pm$0.011} & \textbf{0.066$\pm$0.009} \\
IsoAE & 0.255$\pm$0.094 & 0.139$\pm$0.059 & 0.119$\pm$0.009 & 0.505$\pm$0.090 & 0.196$\pm$0.024 & 0.082$\pm$0.029 & 0.159$\pm$0.032 \\
CAE & 0.198$\pm$0.021 & 0.146$\pm$0.083 & 0.111$\pm$0.011 & 0.361$\pm$0.033 & 0.132$\pm$0.015 & 0.153$\pm$0.063 & 0.214$\pm$0.074 \\
\bottomrule
\end{tabular}
\end{sidewaystable}

\end{document}